\documentclass[preprint,onefignum,onetabnum]{siamart190516}

\usepackage{adjustbox, algpseudocode, amsmath, amssymb, blindtext, bm, bbm, dblfloatfix, esint, fancyhdr, float, graphicx, letltxmacro, marginnote, mathtools, subcaption, xcolor, titlesec, esint}
\usepackage[shortlabels]{enumitem}
\usepackage[linesnumbered,algoruled,algo2e]{algorithm2e}
\usepackage{lipsum}
\usepackage{amsfonts}
\usepackage{epstopdf}

\usepackage[bottom]{footmisc}

\usepackage{listings}
\usepackage{color}

\usepackage{cite}

\allowdisplaybreaks

\newsiamremark{remark}{Remark}

\headers{Galerkin Neural Networks}{M. Ainsworth and J. Dong}

\DeclareMathOperator*{\argmax}{arg\,max}

\title{Galerkin Neural Networks: A Framework for Approximating Variational Equations with Error Control}
\author{Mark Ainsworth\thanks{Division of Applied Mathematics, Brown University
  (\email{mark\_ainsworth@brown.edu}).}
\and Justin Dong\thanks{Division of Applied Mathematics, Brown University
  (\email{justin\_dong@brown.edu}).}}

\begin{document}

\maketitle
	
	
	\begin{abstract}
		We present a new approach to using neural networks to approximate the solutions of variational equations, based on the adaptive construction of a sequence of finite-dimensional subspaces whose basis functions are realizations of a sequence of neural networks. The finite-dimensional subspaces are then used to define a standard Galerkin approximation of the variational equation. This approach enjoys a number of advantages, including: the sequential nature of the algorithm offers a systematic approach to enhancing the accuracy of a given approximation; the sequential enhancements provide a useful indicator for the error that can be used as a criterion for terminating the sequential updates; the basic approach is largely oblivious to the nature of the partial differential equation under consideration; and, some basic theoretical results are presented regarding the convergence (or otherwise) of the method which are used to formulate basic guidelines for applying the method. 
	\end{abstract}
	
	\vspace{5mm}
	


	\section{Introduction}
	
	The current surge in interest in neural network approximation has witnessed the application of neural networks to a range of traditional problems in scientific computing and, in particular, the numerical solution of differential equations. The first problem encountered in applying neural networks to approximate the solution of a partial differential equation (PDE) is to decide on the basic formulation and how to choose the loss function to be used to train the network to approximate the true solution of the equation in hand. One popular formulation is to take the loss to be the square of the $L^{2}$-norm of the residual in the equation augmented by additional terms to take account of the boundary conditions \cite{dissanayake,lagaris,lagaris2,berg,raissi}. Similar approaches also form the basis for least squares type finite element methods, which are more mainstream approaches based on formulating the PDE and boundary conditions as a variational equation. 
	
	Variational approaches for constructing neural network approximations of PDEs have been explored in \cite{weinan,he2,zang,kharazmi}. While such approaches have enjoyed success in particular cases, in practice, the achievable relative error tends to stagnate around $0.1$-$1.0$\% irrespective of the number of neurons or layers used to define the underlying network architecture. In addition, despite recent progress \cite{mishra,han,sirignano}, the current level of theoretical support and understanding of the behavior and convergence of neural network schemes pales in comparison to the vast literature on more traditional approaches. It seems that, at the time of writing, neural network methods are unable to offer a competitive alternative to more traditional approaches such as finite elements or spectral methods.

	The current work follows a quite different approach to using neural networks to approximate variational equations, based on the adaptive construction of a sequence of finite-dimensional subspaces whose basis functions are realizations of a sequence of neural networks. The finite-dimensional subspaces can be used to define a standard Galerkin approximation of the variational equation. This approach enjoys a number of advantages: the need to train a single, large network is replaced by the solution of a sequence of smaller problems combined with a Galerkin approach to select the best linear combination of the resulting corrections; the sequential nature of the algorithm offers a systematic approach to enhancing the accuracy of a given approximation (without having to resort to training a new, larger network); the sequential enhancements can be shown to provide a useful indicator for the error in the current neural network approximation that can be used as a criterion for terminating the sequential updates; and the basic approach is to some extent oblivious to the nature of the PDE under consideration beyond assuming that a stable variational formulation is available. A notable feature of the approach is that it may be applied seamlessly to a broad class of problems including PDEs of various orders and spatial dimensions without major modifications. One may contrast this with more traditional approaches, such as finite element methods, which generally require significant structural modification if the spatial dimension or order of the PDE is changed. Finally, some basic theoretical results are presented regarding the convergence (or otherwise) of the method which are used to formulate some basic guidelines for applying the method. 
%
%
%
%
	 
	The remainder of this paper is organized as follows. Section \ref{sec:galerkin} summarizes some basic properties of Galerkin approximations that are needed in Section \ref{sec:dual representation} where we describe a new approach for approximating the solutions of variational equations using neural networks. The main idea, developed in Section \ref{sec:iterative}, is to adaptively construct a sequence of subspaces consisting of neural network functions from which a Galerkin approximation is obtained.  Implementation details are provided in
Section \ref{sec:nn}.  The flexibility of the approach is illustrated in Section \ref{sec:results} by applying the method to approximate solutions of
both second order and fourth order differential equations, including cases where the data and solutions are singular.

	\section{Basic Framework \& Algorithm} 
	A feed-forward neural network with a single hidden layer of $n \in \mathbb{N}$ neurons defines a function $u_{NN} : \mathbb{R}^{d} \to \mathbb{R}$ as follows:
	\begin{align} \label{eq:NN output}
		u_{NN}(x;\theta) = \sum_{j=1}^{n} c_{j}\sigma(x \cdot W_{j} + b_{j}), \;\;\;x \in \Omega \subset \mathbb{R}^{d},
	\end{align}
	
	\noindent where $d \in \mathbb{N}$ is fixed; $\Omega$ is a compact subset of $\mathbb{R}^{d}$; $W_{j} \in \mathbb{R}^{d}$, $b_{j} \in \mathbb{R}$ are nonlinear parameters; $c_{j} \in \mathbb{R}$ are linear parameters; and $\sigma : \mathbb{R} \to \mathbb{R}$ is a smooth, bounded function. On occasion, we will use the shorthand notation $W = [W_{1}, \dots, W_{n}] \in \mathbb{R}^{d\times n}$, $b=[b_{1},\dots,b_{n}] \in \mathbb{R}^{1\times n}$, and $c=[c_{1},\dots,c_{n}]^{T} \in \mathbb{R}^{n}$. In the standard nomenclature, $n$ is referred to as the \emph{width} of the network, $W$ as the \emph{weights}, $b$ as the \emph{biases}, and $\sigma$ as the \emph{activation function} of the hidden layer. The set of nonlinear and linear parameters is collectively denoted by $\theta = \{W,b,c\}$. The set of all functions of the form \eqref{eq:NN output} is denoted by
	\begin{align} \label{eq:NN}
		V^{\sigma}_{n} := \{v  : v(x) = \sum_{i=1}^{n} c_{i}\sigma(x \cdot W_{i}+ b_{i}), \;b_{i}, c_{i} \in \mathbb{R}, \;W_{i} \in \mathbb{R}^{d}, \;x \in \Omega \}.
	\end{align}
	
	\noindent We shall also define the subset of $V^{\sigma}_{n}$ consisting of functions with bounded parameters. Given $C > 0$, we define
	\begin{align} \label{eq:NN bounded}
		V^{\sigma}_{n,C} := \{v \in V^{\sigma}_{n} : ||\theta|| \leqslant C \},
	\end{align}
	
	\noindent where $||\theta|| := \max_{ij} |W_{ij}| + \max_{j} |b_{j}| + \max_{j} |c_{j}|$.
	
	Neural networks are known to be \emph{universal approximators} \cite{cybenko, leshno} in the sense that the following result holds:
	\begin{theorem}
		\textbf{Universal Approximation} \cite{hornik}. Suppose $1 \leqslant p < \infty$, $0 \leqslant s < \infty$, and $\Omega \subset \mathbb{R}^{d}$ is compact. If $\sigma \in C^{s}(\Omega)$ is nonconstant and bounded, then $V^{\sigma}_{n}$ is dense in the Sobolev space
		\begin{align*}
			W^{s,p}(\Omega) := \{v \in L^{p}(\Omega) : D^{\alpha}v \in L^{p}(\mu) \;\forall |\alpha| \leqslant s\}.
		\end{align*}
	\end{theorem} 
	
	\noindent In particular, the universal approximation theorem means that given a function $f \in W^{s,p}(\Omega)$ and $\tau > 0$, there exists $n(\tau,f) \in \mathbb{N}$ and $\tilde{f} \in V^{\sigma}_{n(\tau,f)}$ such that $||f - \tilde{f}||_{W^{s,p}(\Omega)} < \tau$.

	However, neural networks exhibit several undesirable properties as well. In particular, $V^{\sigma}_{n}$ is not a vector space since it is not closed under addition. For instance, consider the case with $n=1$ and $\sigma(t) = \max\{0,t\}$. Then $\sigma(t) \in V^{\sigma}_{1}$ and $\sigma(t-1) \in V^{\sigma}_{1}$, but $\sigma(t) + \sigma(t-1) \notin V^{\sigma}_{1}$. The fact that $V^{\sigma}_{n}$ is not a vector space means we cannot draw upon many of the powerful results from functional analysis which are typically employed in analyzing the well-posedness of variational problems soon to be discussed. Moreover, $V^{\sigma}_{n}$ is neither closed nor compact in $C(\Omega)$ or $L^{p}(\Omega)$. As demonstrated in \cite{petersen2018topological} however, by restricting the network parameters as in $V^{\sigma}_{n,C}$, we can recover both closedness and compactness in $C(\Omega)$ and $L^{p}(\Omega)$. These topological properties of $V^{\sigma}_{n,C}$ become important in ensuring existence of a solution to the optimization problem we shall introduce in Section \ref{sec:dual representation} (see Remark \ref{remark:compactness}).

	\subsection{Galerkin Approximation of Variational Equations} \label{sec:galerkin}
	Let $s \in \mathbb{R}_{\geqslant 0}$ be fixed, let $X \subset H^{s}(\Omega) =W^{s,2}(\Omega)$ be a subspace with $V^{\sigma}_{n} \subset X$ a subset, and consider the variational equation: find
	\begin{align} \label{eq:variational}
		u \in X : \;\;\;a(u,v) = L(v), \;\;\;\forall v \in X,
	\end{align}
	
	\noindent where $L : X \to \mathbb{R}$ is a bounded linear operator and $a : X \times X \to \mathbb{R}$ is a bounded, symmetric, bilinear form on $X$ satisfying the following condition:
	\begin{align} \label{eq:condition}
		a(v,v) \geqslant 0 \;\text{with equality}  \iff v=0.
	\end{align}
	
	\noindent This condition implies that $a(\cdot,\cdot)$ defines an inner product on $X$ and an associated norm given by $|||v||| := \sqrt{a(v,v)}$. In particular, the Cauchy-Schwarz inequality holds with respect to $|||\cdot|||$. 

	A standard application of the Riesz Representation Theorem \cite{folland} implies the existence of a unique $u \in X$ satisfying \eqref{eq:variational} with $|||u||| = ||L||_{*}$, where $||\cdot||_{*}$ denotes the dual, or operator, norm of $L$ given by
	\begin{align} \label{eq:op norm}
		||L||_{*} := \sup_{v \in X\backslash \{0\}} \frac{|L(v)|}{|||v|||} = \sup_{v \in B} |L(v)|,
	\end{align}
	
	\noindent with $B = \{v \in X : |||v|||=1\}$ being the unit sphere in $X$. As $X \subset H^{s}(\Omega)$, we observe that the universal approximation property holds for the norm $|||\cdot |||$ whenever $s \in \mathbb{N}$ \cite{hornik}. That is, given $f \in X$ and $\tau > 0$, there exists $n(\tau,f) \in \mathbb{N}$ and $\tilde{f} \in V^{\sigma}_{n(\tau,f)}$ such that $|||f-\tilde{f}||| < \tau$. The result is extended to $s \in \mathbb{R}_{\geqslant 0}$ by the embedding $H^{s}(\Omega) \subset H^{\lfloor s \rfloor}(\Omega)$.
	\begin{remark} \label{remark:approx width}
		The width $n$ of the network needed to ensure $|||f-\tilde{f}||| < \tau$ depends on both $\tau$ and $f$.  For instance, suppose $f \in H^{2}(0,1)$ and $\sigma$ is chosen as the ReLU function $t \mapsto \max\{0,t\}$. Standard results for piecewise linear approximation show that there exists a constant $\mathcal{C}>0$ such that for all $m \in \mathbb{N}$, there exists $f_{m} \in V^{\sigma}_{m}$ satisfying $||f-f_{m}||_{L^{2}(0,1)} \leqslant \mathcal{C}m^{-2}||f''||_{L^{2}(0,1)}$. Hence, in order to achieve $||f-f_{m}||_{L^{2}(0,1)} < \tau$, it suffices to choose $m > \sqrt{\mathcal{C}||f''||_{L^{2}}/\tau}$. Observe that as the required tolerance decreases or the structure of $f$ grows more complex, the width of the network needed to meet the tolerance will increase.
	\end{remark}
	
	In general, the solution of the variational problem cannot be obtained in closed form and instead must be approximated. Given a finite-dimensional subspace $S \subset X$, the Galerkin approximation $u_{S} \in S$ of problem \eqref{eq:variational} consists of seeking
	\begin{align} \label{eq:galerkin}
		u_{S} \in S : \;\;\;a(u_{S},v_{S}) = L(v_{S}) \;\;\;\forall v_{S} \in S.
	\end{align}
	
%
%
	
	\noindent Equally well, we have $u_{S} = P_{S}u$, where $P_{S} : X \to S$ is the orthogonal projection given by the rule $a(P_{S}z,v_{S}) = a(z,v_{S}), \;\forall v_{S} \in S$ for each $z \in X$. Galerkin approximation schemes enjoy widespread usage due to the fact that they are not only stable since $|||P_{S}u||| \leqslant |||u|||$, but deliver the best approximation to the true solution $u$ from the subspace $S$ \cite{brennerscott,ern}, i.e.
	\begin{align} \label{eq:cea}
		|||u - u_{S}||| \leqslant |||u - v_{S}|||, \;\;\;\forall v_{S} \in S.
	\end{align}
	
	\noindent Moreover, if $\{\varphi_{j}\}_{j=1}^{m}$ is a basis for $S$, then the approximation $u_{S}$ may be computed explicitly by solving the linear system
	\begin{align} \label{eq:galerkin linear}
		\begin{dcases}
			K\hat{u} = F\\
			K_{ij} = a(\varphi_{i},\varphi_{j}), \;\;\;F_{i} = L(\varphi_{i})
		\end{dcases}
	\end{align}
	
	\noindent and forming $u_{S} = \sum_{j=1}^{m}\hat{u}_{j}\varphi_{j}$. 
	
%
%
%
	
	It is tempting to seek a neural network approximation $u_{NN} \approx u$ by employing a Galerkin scheme based on the choice $S=V^{\sigma}_{n}$. Unfortunately, the fact that $V^{\sigma}_{n}$ is not a vector space nullifies much of the above discussion and precludes us from using it to define a Galerkin scheme of the form \eqref{eq:galerkin}. Nonetheless, we observe that $V^{\sigma}_{n}$ may still be utilized to generate approximations to $u$ in two different ways. 
	
	First, by fixing the nonlinear parameters $(W,b)$ and forming the subspace $\Phi := \text{span}\{\sigma(x\cdot W_{i} + b_{i})\}_{i=1}^{n}$, a Galerkin approximation of the form \eqref{eq:NN} can be computed from $\Phi$ by choosing the linear parameters $c$ as follows:
	\begin{align} \label{eq:glsq}
		\begin{dcases}
			Kc = F\\
			K_{ij} = a(\sigma(x\cdot W_{i}+b_{i}), \sigma(x\cdot W_{j}+b_{j})), \;\;\;F_{i} = L(\sigma(w\cdot W_{i}+b_{i})).
		\end{dcases}
	\end{align}
	
	\noindent This process defines a function $c = \textsc{GalerkinLSQ}(W,b,\sigma,a,L)$ which computes the expansion coefficients of the projection of $u$ onto the subspace $\Phi = \text{span}\{\sigma(x\cdot W_{i} + b_{i}\}_{i=1}^{n}$ for fixed $W$ and $b$. In practice, however, the freedom to vary $W$ and $b$ is largely responsible for the rich approximation power of neural networks. Nevertheless, the function $\textsc{GalerkinLSQ}$ will play a role in Section \ref{sec:nn}.
	
	Second, if $\{\psi_{i}\}_{i=1}^{m}$ is a linearly independent set of functions in $V^{\sigma}_{n}$, then we may similarly define $\Phi := \text{span}\{\psi_{1},\dots,\psi_{m}\}$ and solve the linear system
	\begin{align} \label{eq:galerkin solve}
		\begin{dcases}
			K\hat{u} = F\\
			K_{ij} = a(\psi_{i},\psi_{j}), \;\;\;F_{i} = L(\psi_{i})
		\end{dcases}
	\end{align}
	
	\noindent from which the projection $P_{\Phi}u = \sum_{j=1}^{m} \hat{u}_{j}\psi_{j}$ may be obtained. This process defines a function $P_{\Phi}u = \textsc{GalerkinSolve}(\{\psi_{1}, \dots, \psi_{m}\}, a, L)$ which returns the projection of $u$ onto the subspace $\Phi := \text{span}\{\psi_{1},\dots,\psi_{m}\}$ with respect to the variational equation defined by the bilinear form $a$ and data $L$. The routine \textsc{GalerkinSolve} will play a key role in Section \ref{sec:iterative}.

	\subsection{Construction of the Galerkin Subspace} \label{sec:dual representation}
	Suppose $u_{0} \in X$ is any given approximation to the solution of \eqref{eq:variational}; even $u_{0} = 0 \in X$ is allowed. Can we improve the accuracy of the approximation $u_{0} \approx u$ by adding a suitable correction to $u_{0}$ from the set $V^{\sigma}_{n}$? 
	
	We can measure how well $u_{0}$ approximates the true solution $u$ by examining the residual $L(v) - a(u_{0},v)$, $v \in X$ in \eqref{eq:variational}, which vanishes if and only if $u_{0} = u$. In general, the residual will be nonzero and defines a functional $r(u_{0}) : X \to \mathbb{R}$ given by the rule $\langle r(u_{0}),v \rangle = L(v) - a(u_{0},v)$, where $\langle r(u_{0}),v \rangle$ denotes the duality pairing on $X$.
	
	The following result establishes the relationship between the residual $r(u_{0})$ and the error $u-u_{0}$ in the approximation $u \approx u_{0}$.
	
	\begin{proposition} \label{prop:maximizer}
		Let $u_{0} \in X$ be given and set $\varphi_{1} := (u-u_{0})/|||u-u_{0}|||$. Then $\langle r(u_{0}),\varphi_{1} \rangle = \max_{v \in B} \langle r(u_{0}),v \rangle$, where $B$ is the closed unit ball in $X$.
	\end{proposition}
	
	\begin{proof}
		Let $u_{0} \in X$ be given. We use \eqref{eq:variational} to obtain
		\begin{align*}
			\langle r(u_{0}),v \rangle &= L(v) - a(u_{0},v) = a(u,v) - a(u_{0},v) = a(u-u_{0},v), \;\;\;\forall v \in X.
		\end{align*}
		
		\noindent By the Cauchy-Schwarz inequality,
		\begin{align*}
			\sup_{v \in B} \langle r(u_{0}),v \rangle = \sup_{v \in B} a(u-u_{0},v) \leqslant \sup_{v \in B} |||u-u_{0}||| \cdot |||v||| = |||u-u_{0}|||.	
		\end{align*}
		
		On the other hand, choosing $v = \varphi_{1} = (u-u_{0})/|||u-u_{0}||| \in B$ yields
		\begin{align*}
			a(u_{0}-u, v) = \frac{|||u-u_{0}|||^{2}}{|||u-u_{0}|||} = |||u-u_{0}|||,
		\end{align*}
		
		\noindent and hence, $\max_{v \in B} \langle r(u_{0}),v \rangle = |||u-u_{0}||| = \langle r(u_{0}),\varphi_{1} \rangle$.
	\end{proof}
	
	\noindent The elementary proof of Proposition \ref{prop:maximizer} is illuminating in that it shows that the maximizer $\varphi_{1} \in B$ of the residual is proportional to the error $e=u-u_{0}$. The optimal correction to $u_{0}$ is the true error $e$ since $u_{0}+e = u_{0} + (u-u_{0}) = u$ yields the true solution. The fact that the maximizer $\varphi_{1}$ is proportional to the error means that we should take the correction to $u_{0}$ to be a multiple of $\varphi_{1}$.
	
	 Unfortunately, we will not be able to compute the maximizer $\varphi_{1}$ explicitly which, as mentioned above, would be equivalent to solving the variational problem \eqref{eq:variational} exactly. Instead, we compute an approximation $\varphi_{1}^{NN} \approx \varphi_{1}$ by seeking 
	\begin{align} \label{eq:NN maximizer}
		\varphi_{1}^{NN} \in V^{\sigma}_{n,C} : \langle r(u_{0}),\varphi_{1}^{NN} \rangle = \max_{v \in V^{\sigma}_{n,C} \cap B} \langle r(u_{0}),v \rangle.
	\end{align}
	
	\begin{remark} \label{remark:compactness}
		A function $\varphi_{1}^{NN}$ as defined in \eqref{eq:NN maximizer} exists. Observe that 
		\begin{align*}
			\sup_{v \in V^{\sigma}_{n,C} \cap B} \langle r(u_{0}),v \rangle \leqslant \sup_{v \in B} \langle r(u_{0}),v \rangle < \infty.
		\end{align*}
		
		\noindent Moreover, $V^{\sigma}_{n,C}$ is compact in $H^{s}(\Omega)$ by \cite[Proposition 3.5]{petersen2018topological} and thus, $V^{\sigma}_{n,C} \cap B$ is compact in $X$. The supremum is therefore achieved over $V^{\sigma}_{n,C} \cap B$ and, as a consequence, the function $\varphi_{1}^{NN}$ appearing in \eqref{eq:NN maximizer} is well-defined.
	\end{remark}
	
	\noindent We then form the subspace $S_{1} = \text{span}\{\varphi_{0}^{NN},\varphi_{1}^{NN}\}$ where $\varphi_{0}^{NN} := u_{0}/|||u_{0}|||$ and compute a Galerkin approximation to $u$ by seeking
	\begin{align} \label{eq:u1}
		u_{1} \in S_{1} \;:\; a(u_{1},v) = L(v) \;\;\;\forall v \in S_{1}.
	\end{align}
	
%
%
%
	
	The following result demonstrates that if the network width $n$ and the bound $C$ are sufficiently large, then any such function $\varphi_{1}^{NN} \in V^{\sigma}_{n,C}$ as defined in \eqref{eq:NN maximizer} is a good approximation to $\varphi_{1}$. 
	
	\begin{proposition} \label{prop:estimator}
	 Let $0 < \tau < 1$ be given. Then there exist $n(\tau,u_{0}) \in \mathbb{N}$ and $C(\tau,u_{0}) > 0$ such that if $n \geqslant n(\tau,u_{0})$ and $C \geqslant C(\tau,u_{0})$, then $\varphi_{1}^{NN} \in V^{\sigma}_{n,C} \cap B$ given by \eqref{eq:NN maximizer} satisfies $|||\varphi_{1} - \varphi_{1}^{NN}||| < 2\tau/(1-\tau)$.
	\end{proposition}
	
	\begin{proof}
		We first recall from Proposition \ref{prop:maximizer} that
		\begin{align*}
			\varphi_{1} = \argmax_{v \in B} \langle r(u_{0}),v \rangle = \argmax_{v \in B} a(u-u_{0},v).
		\end{align*}
		
		\noindent By the universal approximation property, there exists $n(\tau,u_{0}) \in \mathbb{N}$ and $\tilde{\varphi} \in V^{\sigma}_{n}$ such that $|||\varphi_{1}-\tilde{\varphi}||| < \tau$. Furthermore, by the triangle inequality, we have $1-\tau < |||\tilde{\varphi}||| < 1+\tau$ or, equally well, $|||\tilde{\varphi}|||-1 = \xi$ for some $\xi \in (-\tau,\tau)$.  Let $\hat{\varphi} := \tilde{\varphi}/|||\tilde{\varphi}||| \in V^{\sigma}_{n} \cap B$. Then,
		\begin{align*}
			|||\varphi_{1} - \hat{\varphi}||| &\leqslant |||\varphi_{1} - \tilde{\varphi}||| + |||\tilde{\varphi} - \frac{1}{|||\tilde{\varphi}|||}\tilde{\varphi}|||\\
			&< \tau + \frac{1}{|||\tilde{\varphi}|||}|||\;|||\tilde{\varphi}|||\cdot\tilde{\varphi} - \tilde{\varphi}|||\\
			&< \tau + \frac{1}{1-\tau}|||(1+\xi)\tilde{\varphi}-\tilde{\varphi}|||\\
			&= \tau + \frac{\xi}{1-\tau}|||\tilde{\varphi}|||\\
			&< \tau + \frac{\tau}{1-\tau}(1+\tau) = \frac{2\tau}{1-\tau}.
		\end{align*}
		
		\noindent Set $C(\tau,u_{0}) := ||\hat{\theta}||$, where $\hat{\theta}$ denotes the parameters corresponding to $\hat{\varphi} \in V^{\sigma}_{n}$.
		
		Now, $\langle r(u_{0}),v \rangle = a(u-u_{0},v) = \gamma a(\varphi_{1},v)$ where $\gamma = |||u-u_{0}|||$. Let $n \geqslant n(\tau,u_{0})$ and $C \geqslant C(\tau,u_{0})$. The solution of \eqref{eq:NN maximizer} satisfies 
		\begin{align*}
			\langle r(u_{0}),\varphi_{1}^{NN} \rangle = \max_{v \in V^{\sigma}_{n,C} \cap B} \langle r(u_{0}),v \rangle = \max_{v \in V^{\sigma}_{n,C} \cap B} \gamma a(\varphi_{1},v).
		\end{align*}

		\noindent Moreover, since $\gamma a(\varphi_{1},v) \leqslant \gamma a(\varphi_{1},\varphi_{1}^{NN})$ for all $v \in V^{\sigma}_{n,C} \cap B$, 
		\begin{align*}
			|||\varphi_{1}-\varphi_{1}^{NN}|||^{2} &= |||\varphi_{1}|||^{2} - 2a(\varphi_{1},\varphi_{1}^{NN}) + |||\varphi_{1}^{NN}|||^{2}\\
			&= 2 - 2\gamma^{-1}\cdot \gamma a(\varphi_{1},\varphi_{1}^{NN})\\
			&\leqslant 2 - 2\gamma^{-1}\cdot \gamma a(\varphi_{1},\hat{\varphi})\\
			&= |||\varphi_{1}-\hat{\varphi}|||^{2} < 4\tau^{2}/(1-\tau)^{2}
		\end{align*}
		
		\noindent as required. 
	\end{proof}
	
	
	Proposition \ref{prop:estimator} shows that $\varphi_{1}^{NN} \in V_{n,C}^{\sigma}$ tends to the true maximizer $\varphi_{1}$ as the network width $n$ increases. The next result shows that the corresponding approximation $u_{1}$ tends to the true solution $u$ as $n$ and $C$ increase.
	\begin{proposition} \label{prop:convergence}
		Let $0 < \tau < 1$, $n(\tau,u_{0}) \in \mathbb{N}$, and $C(\tau,u_{0})$ be as in Proposition \ref{prop:estimator}, and $u_{1}$ be defined as in \eqref{eq:u1}. Then $|||u-u_{1}||| \leqslant |||u-u_{0}|||\cdot\min\{1,2\tau/(1-\tau)\}$.
	\end{proposition}
	
	\begin{proof}
		Choosing $v_{S} = u_{0} \in S_{0}$ in  \eqref{eq:cea} yields $|||u-u_{1}||| \leqslant |||u-u_{0}|||$. Conversely, choosing $v_{S} = |||u_{0}|||\cdot\varphi_{0}^{NN} + |||u-u_{0}|||\cdot\varphi_{1}^{NN} \in S_{0}$ and recalling that $\varphi_{1} = (u-u_{0})/|||u-u_{0}|||$ yields
		\begin{align*}
			|||u-u_{1}||| &\leqslant |||(u-u_{0}) - |||u-u_{0}|||\cdot\varphi_{1}^{NN}|||\\
			&= |||\varphi_{1}\cdot |||u-u_{0}||| - |||u-u_{0}|||\cdot \varphi_{1}^{NN}|||\\
			&= |||u-u_{0}|||\cdot |||\varphi_{1} - \varphi_{1}^{NN}|||\\
			&< |||u-u_{0}|||\cdot \frac{2\tau}{1-\tau},
		\end{align*}
		
		\noindent where the final step follows thanks to Proposition \ref{prop:estimator}.
	\end{proof}
	
	Proposition \ref{prop:convergence} gives a bound on the accuracy of the approximation $u \approx u_{1}$ in terms of $\tau$ and the (unknown) error in $u \approx u_{0}$. Interestingly, Proposition \ref{prop:maximizer} shows that $|||u-u_{0}||| = \langle r(u_{0}),\varphi_{1} \rangle$ which, were $\varphi_{1}$ available, would mean the actual error could be computed explicitly. Does replacing $\varphi_{1}$ by the approximation $\varphi_{1} \approx \varphi_{1}^{NN}$ then result in a good estimator for the true error? To be precise, we define the quantity 
	 \begin{align} \label{eq:eta}
	 	\eta(u_{0}, v) := \langle r(u_{0}),v \rangle / |||v|||
	 \end{align}
	 
	 \noindent and observe that $\eta(u_{0},\varphi_{1}^{NN})$ can be evaluated explicitly once $\varphi_{1}^{NN}$ and $u_{0}$ are in hand. The following result shows that, provided $\tau < 1/3$, the numerical value of $\eta(u_{0},\varphi_{1}^{NN})$ is indeed equivalent to the true error $|||u-u_{0}|||$.

	\begin{corollary} \label{cor:estimator}
		Let $\eta(u_{0},v)$ be as defined in \eqref{eq:eta}. If $0 < \tau < 1/3$, then
		\begin{align} \label{eq:error equivalence}
			\frac{1-\tau}{1+\tau}\eta(u_{0},\varphi_{1}^{NN}) \leqslant |||u-u_{0}||| \leqslant \frac{1-\tau}{1-3\tau}\eta(u_{0},\varphi_{1}^{NN}).
		\end{align}
	\end{corollary}
	
	\begin{proof}
		By Proposition \ref{prop:maximizer}, we have $|||u-u_{0}||| = \langle r(u_{0}),\varphi_{1} \rangle$. Hence,
		\begin{align*}
			|\eta(u_{0},\varphi_{1}^{NN}) - |||u-u_{0}|||\;| &= |\langle r(u_{0}),\varphi_{1}^{NN} \rangle - \langle r(u_{0}),\varphi_{1} \rangle| = |a(u-u_{0},\varphi_{1}^{NN}-\varphi_{1})|,
		\end{align*}
		
		\noindent and applying Proposition \ref{prop:estimator} yields
		\begin{align*}
			|\eta(u_{0},\varphi_{1}^{NN}) - |||u-u_{0}|||\;| &\leqslant |||u-u_{0}||| \cdot |||\varphi_{1}^{NN}-\varphi_{1}||| < |||u-u_{0}||| \cdot \frac{2\tau}{1-\tau}.
		\end{align*}
		
		\noindent If $\tau \in (0,1/3)$ then $2\tau/(1-\tau) \in (0,1)$, and hence
		\begin{align*}
			\eta(u_{0},\varphi_{1}^{NN}) \leqslant |||u-u_{0}||| \cdot \left(\frac{2\tau}{1-\tau} + 1 \right) = |||u-u_{0}||| \cdot \frac{1+\tau}{1-\tau}
		\end{align*}
		
		\noindent and
		\begin{align*}
			\eta(u_{0},\varphi_{1}^{NN}) \geqslant |||u-u_{0}||| \cdot \left(-\frac{2\tau}{1-\tau} + 1 \right) = |||u-u_{0}|||\cdot\frac{1-3\tau}{1-\tau}.
		\end{align*}
		
		\noindent  Combining these estimates gives \eqref{eq:error equivalence}.
	\end{proof}
	
	In summary, once the solution $\varphi_{1}^{NN}$ of \eqref{eq:NN maximizer} is in hand, we can (a) obtain an improved approximation to $u \approx u_{1}$ by solving \eqref{eq:u1} and (b) obtain a numerical estimate $\eta(u_{0},\varphi_{1}^{NN}) \approx |||u-u_{0}|||$ for the error. The foregoing arguments show that if $\tau \to 0$, or, equally well, the network width $n \to \infty$ and bound $C \to \infty$, then the resulting approximation $u_{1} \to u$ (in $X$) and $\eta(u_{0},\varphi_{1}^{NN}) \to |||u-u_{0}|||$ (in $\mathbb{R}$).

	\subsection{Galerkin Subspace Augumentation} \label{sec:iterative}
	
	The procedure outlined in the preceding section gives an improved approximation $u_{1} = P_{S_{1}}u$ to $u$ from the two-dimensional subspace $S_{1} = \text{span}\{\varphi_{0}^{NN},\varphi_{1}^{NN}\}$. How should we proceed if the approximation $u \approx u_{1}$ is lacking? A natural extension of this approach would be to apply the same procedure that was used to improve the approximation $u \approx u_{0}$ to improve $u_{1}$. Namely, we compute an approximate maximizer $\varphi_{2}^{NN}$ of the new residual $\langle r(u_{1}),v \rangle$ and use it to further augment the Galerkin subspace by setting $S_{2} = \text{span}\{\varphi_{0}^{NN},\varphi_{1}^{NN},\varphi_{2}^{NN}\}$. Proceeding recursively, we construct the finite-dimensional subspace $S_{j} = \text{span}\{\varphi_{0}^{NN},\varphi_{1}^{NN}, \varphi_{2}^{NN}, \dots, \varphi_{j}^{NN}\}$ with $\varphi_{i}^{NN} \in V^{\sigma_{i}}_{n_{i},C_{i}}$.
%
	
	Here, we have allowed for the possibility to vary both the network width $n_{i}$ and activation function $\sigma_{i}$. Proceeding as in Section \ref{sec:dual representation}, we define $\varphi_{i} = (u-u_{i-1})/|||u-u_{i-1}|||$ and compute an approximation to $\varphi_{i}$ by seeking
	\begin{align} \label{eq:phi_j}
		\varphi_{i}^{NN} \in V^{\sigma_{i}}_{n_{i},C_{i}} : \langle r(u_{i-1}),\varphi_{i}^{NN} \rangle = \max_{v \in V^{\sigma_{i}}_{n_{i},C_{i}} \cap B}\langle r(u_{i-1}),v \rangle 
	\end{align}
	
	\noindent and then computing a new approximation $u \approx u_{i}$ as follows: 
	\begin{align} \label{eq:uj}
		u_{i} \in S_{i} \;:\; a(u_{i},v) = L(v) \;\;\;\forall v \in S_{i},
	\end{align}
	
	\noindent or, equally well, $u_{i} = P_{S_{i}}u$. 
	
	We have the following analogs to Propositions \ref{prop:maximizer}, \ref{prop:estimator}, and \ref{prop:convergence} and Corollary \ref{cor:estimator}.
	\begin{corollary}
		For each $i > 0$, let $u_{i} = P_{S_{i}}u$ be given and define $\varphi_{i} :=(u-u_{i-1})/|||u-u_{i-1}|||$. Then $\langle r(u_{i-1}),\varphi_{i} \rangle = \max_{v \in B} \langle r(u_{i-1}),v \rangle$.
	\end{corollary}
	
	\begin{corollary} \label{cor:estimator2}
		For each $i>0$, let $0 < \tau_{i} < 1$ be given. Then there exists $n(\tau_{i},u_{i-1}) \in \mathbb{N}$ and $C(\tau_{i},u_{i-1}) > 0$ such that if $n \geqslant n(\tau_{i},u_{u-1})$ and $C \geqslant C(\tau_{i},u_{i-1})$, then the function $\varphi_{i}^{NN} \in V^{\sigma_{i}}_{n,C}$ as defined in \eqref{eq:phi_j} satisfies $|||\varphi_{i} - \varphi_{i}^{NN}||| < 2\tau_{i}/(1-\tau_{i})$. Moreover, if $\tau_{i} < 1/3$, then 
		\begin{align*}
			\frac{1-\tau_{i}}{1+\tau_{i}}\eta(u_{i-1},\varphi_{i}^{NN}) \leqslant |||u-u_{i-1}||| \leqslant \frac{1-\tau_{i}}{1-3\tau_{i}}\eta(u_{i-1},\varphi_{i}^{NN}).
		\end{align*}
	\end{corollary}
	
	In particular, Corollary \ref{cor:estimator2} provides a computable estimate of the error $|||u-u_{i-1}|||$ that can be used as a stopping criterion for the recursive solution procedure as follows: Given a tolerance $\texttt{tol} > 0$, if $\eta(u_{i-1},\varphi_{i}^{NN}) > \texttt{tol}$, then we augment the subspace $S_{i-1}$ to $S_{i} = S_{i-1} \oplus \{\varphi_{i}^{NN}\}$ and compute an updated approximation $u_{i} \approx u$; otherwise, the approximation $u_{i-1}$ is deemed satisfactory, and the procedure terminates. Algorithm \ref{alg:adaptive} summarizes this recursive approach, where the function \textsc{AugmentBasis} is the routine which computes the maximizer $\varphi_{i}^{NN}$ in the subspace augmentation process and which will be described in Section \ref{sec:nn}.
	\begin{algorithm2e}[t!]
		\KwIn{Data $L$, bilinear operator $a$, network widths $\{n_{i}\}$, initial approximation $u_{0} \in X$, tolerance $\texttt{tol}>0$.}
		\KwOut{Stopping iteration $N \in \mathbb{N}$, numerical approximation $u_{N}$ to the variational problem: $u \in X$ such that $a(u,v) = \langle f,v \rangle$ for all $v \in X$; basis functions $\{\varphi_{i}^{NN}\}_{i=0}^{N}$.}
		\BlankLine
		\textbf{Require:}\\
		\hspace{5mm}$a$ symmetric and $a(v,v) \geqslant 0$, with $a(v,v) = 0$ if and only if $v=0$.\\
		\BlankLine
		\BlankLine
		Set $i=1$ and $\varphi_{0}^{NN} = u_{0}/|||u_{0}|||$.\\
		$(\varphi_{1}^{NN},\eta) \leftarrow \textsc{AugmentBasis}(u_{0})$.\\
			\BlankLine
			\While{$\eta > \texttt{tol}$}{
				$u_{i} \leftarrow \textsc{GalerkinSolve}(\{\varphi_{0}^{NN}, \varphi_{1}^{NN}, \dots, \varphi_{i}^{NN}\}, a, L)$\\
				$(\varphi_{i+1}^{NN}, \eta) \leftarrow \textsc{AugmentBasis}(u_{i})$.\\
				Set $i \gets i+1$.
			}
			\BlankLine
			Return $N = i-1$, $u_{N}$, and $\{\varphi_{j}^{NN}\}_{j=0}^{N}$.\\
		
		\caption{Adaptive Construction of Subspace.}
		\label{alg:adaptive}
	\end{algorithm2e}
	
	The next result addresses the convergence, or otherwise, of the recursive procedure.
	\begin{corollary} \label{cor:convergence}
		Let $j \geqslant 1$. If $\{\tau_{i}\}_{i=1}^{j}$, $\{n(\tau_{i},u_{i-1})\}_{i=1}^{j}$, and $\{C(\tau_{i},u_{i-1})\}_{i=1}^{j}$ are chosen as in Corollary \ref{cor:estimator2}, then
		\begin{align*}
			|||u-u_{j}||| \leqslant |||u-u_{0}||| \cdot \prod_{i=1}^{j} \min\{1,\frac{2\tau_{i}}{1-\tau_{i}}\}.
		\end{align*}
	\end{corollary}
	
	\begin{proof}
		Proposition \ref{prop:convergence} implies that for each $i$: $|||u-u_{i}||| \leqslant |||u-u_{i-1}|||\cdot \min\{1,2\tau_{i}/(1-\tau_{i})\}$. Bootstrapping this result gives
		\begin{align*}
			|||u-u_{j}||| &\leqslant |||u-u_{j-1}||| \cdot \min\{1,\frac{2\tau_{j}}{1-\tau_{j}}\}\\
			&\hspace{25mm}\vdots\\
			&\leqslant |||u-u_{0}||| \cdot \prod_{i=1}^{j} \min\{1,\frac{2\tau_{i}}{1-\tau_{i}}\}
		\end{align*}
		
		\noindent as claimed.
	\end{proof}


	Lastly, we discuss the conditioning of the linear system $K^{(j)}\hat{u} = F^{(j)}$ associated with the approximation $u \approx u_{j}$ in \eqref{eq:uj}. This linear system follows the framework described by \eqref{eq:galerkin solve}. The matrix $K^{(j)}$ has entries
	\begin{align*}
		K^{(j)}_{k\ell} = a(\varphi_{k-1}^{NN}, \varphi_{\ell-1}^{NN}),
	\end{align*}
	
	\noindent where the basis functions $\{\varphi_{i}^{NN}\}_{i=0}^{j}$ are obtained according to Algorithm \ref{alg:adaptive}. We observe that if $k=\ell$, then $K^{(j)}_{kk} = 1$ since $\varphi_{k}^{NN} \in B$. 
	
%
	
	\begin{proposition} \label{prop:condition number}
		
		Let $\{\varphi_{i}^{NN}\}_{i=0}^{N}$ be the set of basis functions constructed by Algorithm \ref{alg:adaptive}. Then for $i>1$, $\varphi_{i}^{NN} \notin S_{i-1}$. Moreover, if $\gamma = \sum_{k=1}^{N} 2\tau_{k}/(1-\tau_{k}) < 1$, then 
		\begin{align} \label{eq:condition number}
			\text{cond}(K^{(N)}) < \frac{1 + \gamma}{1-\gamma}.
		\end{align}
	\end{proposition}
	
	\begin{proof}
		Suppose, to the contrary, that $\varphi_{i}^{NN} \in S_{i-1}$. Then \eqref{eq:uj} implies $L(\varphi_{i}^{NN}) - a(u_{i-1},\varphi_{i}^{NN}) = \eta(u_{i-1},\varphi_{i}^{NN}) = 0$. But the stopping criterion in Algorithm \ref{alg:adaptive} (see Line 5) ensures that each $\varphi_{i}^{NN}$ satisfies $\eta(u_{i-1},\varphi_{i}^{NN}) > \texttt{tol}$, a contradiction. Thus, $\varphi_{i}^{NN} \notin S_{i-1}$ and, in particular,  the functions $\{\varphi_{i}^{NN}\}_{i=0}^{N}$ are linearly independent.

		For the second assertion, if $k > \ell$, then we have $a(\varphi_{k},\varphi_{\ell}^{NN}) \propto a(u-u_{k-1},\varphi_{\ell}^{NN}) = 0$ by \eqref{eq:uj}, and applying Corollary \ref{cor:estimator2} yields
		\begin{align*}
			|a(\varphi_{k}^{NN},\varphi_{\ell}^{NN})| &= |a(\varphi_{k}^{NN},\varphi_{\ell}^{NN}) - a(\varphi_{k},\varphi_{\ell}^{NN})| = |a(\varphi_{k}^{NN} - \varphi_{k}, \varphi_{\ell}^{NN})|\\
			&\leqslant |||\varphi_{k}^{NN} - \varphi_{k}||| \cdot |||\varphi_{\ell}^{NN}||| < \frac{2\tau_{k}}{1-\tau_{k}}.
		\end{align*}
		
		\noindent Thus, $|K^{(N)}_{\ell k}| = |K^{(N)}_{k\ell}| < 2\tau_{\ell}/(1-\tau_{\ell})$ for $k \neq \ell$ thanks to the symmetry of $a(\cdot,\cdot)$. 
		
		By Gershgorin's Theorem \cite{gershgorin}, for each eigenvalue $\lambda$ of $K^{(N)}$ there exists $\ell$ such that
		\begin{align*}
			|\lambda - K^{(N)}_{\ell\ell}| = |\lambda - 1| < \sum_{k \neq \ell} |K^{(N)}_{k\ell}| < \sum_{k \neq \ell} \frac{2\tau_{k}}{1-\tau_{k}}.
		\end{align*}
		
		\noindent Then $|\lambda - 1| < \gamma$ for each $\lambda$, and
		\begin{align*}
			\text{cond}(K^{(N)}) = \frac{|\max \lambda|}{|\min \lambda|} < \frac{1 + \gamma}{1-\gamma}
		\end{align*}
		
		\noindent as claimed. 
	\end{proof}
	
	In summary, the procedure given by Algorithm \ref{alg:adaptive} adaptively constructs a subspace using the basis functions $\varphi_{i}^{NN} \in V^{\sigma_{i}}_{n_{i},C_{i}}$, $i=1,\dots,N$. By Corollary \ref{cor:convergence}, the resulting Galerkin approximation $u_{N}$ converges to $u$ provided that the widths $n_{i}$ and bounds $C_{i}$ are sufficiently large. Moreover, the off-diagonal entries of $K^{(N)}$ are small provided that $n_{i}$ and $C_{i}$ are sufficiently large. In particular, $\text{cond}(K^{(N)}) \approx 1$.

	\subsubsection{Selection of Network Widths} \label{sec:widthselection}
	Algorithm \ref{alg:adaptive} allows for the flexibility to vary both the width $n_{i}$ of the network and the activation function $\sigma_{i}$ as $i$ is increased. Surprisingly, we can fix $n$ and set $n_{i}\equiv n$ and yet still hope to achieve convergence as $i \to \infty$:

	\begin{proposition} \label{prop:fixed width}
		Let $n \in \mathbb{N}$ be fixed and $\Omega \subset \mathbb{R}^{d}$ compact. Suppose $f \in H^{s}(\Omega)$, $s \in \mathbb{R}_{\geqslant 0}$. For $\tau>0$, there exists $r \in \mathbb{N}$ and $\{u_{j}\}_{j=1}^{r} \in V^{\sigma}_{n}$ such that $||f-\sum_{j=1}^{r}u_{j}||_{H^{s}} < \tau$. 
	\end{proposition}
	
	\begin{proof}
		By the universal approximation property, there exists $n(\tau,f) \in \mathbb{N}$ and $\hat{f} \in V^{\sigma}_{n(\tau,f)}$ such that $||f-\hat{f}||_{H^{s}} < \tau$. Choose $r$ such that $rn > n(\tau,f)$. Then there exists $\tilde{f} \in V^{\sigma}_{nr}$ such that $||f-\tilde{f}||_{H^{s}} < \tau$, where $\tilde{f}(x) = \sum_{k=1}^{nr} c_{k}\sigma(x\cdot W_{k} + b_{k}) = \sum_{j=1}^{r} u_{j}(x)$, with $u_{j}(x) = \sum_{k>(j-1)n}^{jn} c_{k}\sigma(x\cdot W_{k}+b_{k})$. 
	\end{proof}
	
	Proposition \ref{prop:fixed width} shows that any $f \in H^{s}(\Omega)$ can be approximated by a sum of functions in $V^{\sigma}_{n}$ to arbitrary accuracy. In principle, this means fixing $n_{i}\equiv n$ and carrying out sufficiently many iterations of Algorithm \ref{alg:adaptive} should give convergence. However, in practice the rate of convergence will become arbitrarily slow.
	
	For instance, let us again consider the scenario described in Remark \ref{remark:approx width}, where it was found that a ReLU network of width $n(\tau,f) = \sqrt{C||f''|||_{L^{2}}/\tau}$ was sufficient to approximate a function $f \in H^{2}(0,1)$ to a tolerance $\tau$ in $L^{2}(0,1)$. The proof of Proposition \ref{prop:fixed width} shows that a sum of $r \geqslant n^{-1}\sqrt{C||f''||_{L^{2}}/\tau}$ functions in $V^{\sigma}_{n}$ achieves a tolerance of $\tau$. In other words, $\mathcal{O}(\tau^{-1/2})$ iterations of Algorithm \ref{alg:adaptive} are needed to achieve tolerance $\tau$. This means that, on average, the error is reduced by a factor of $\tau^{\sqrt{\tau}}$ per iteration of the algorithm during this phase. However, in the limit as $\tau \to 0$, the average reduction factor tends to $1$, i.e. the algorithm converges arbitrarily slowly. 
	
	
	A similar conclusion is reached if one considers Corollary \ref{cor:convergence} with $n_{i}=n$ is fixed. Specifically, the sequence $\{\tau_{i}\}_{i=1}^{j}$ will generally be increasing due to the decreased ability of a fixed-width network $\varphi_{i}^{NN} \in V^{\sigma}_{n,C}$ to capture higher resolution features of the error as $i$ increases. In other words, there exists $i^{*} \in \mathbb{N}$ such that $\min\{1,2\tau_{i}/(1-\tau_{i})\} = 1$ for all $i \geqslant i^{*}$, meaning that the convergence stagnates eventually.

	On the other hand, Corollary \ref{cor:convergence} also suggests how we might avoid this stagnation. Observe that if $0 < \tau < 1/3$ is fixed and $\tau_{i} = \tau$ for all $i$, then it is expected that the sequence $\{n(\tau,u_{i})\}_{i=0}^{j}$ is nondecreasing. In particular, the error in $u_{i}$ is a factor of $2\tau/(1-\tau)$ smaller than the error in $u_{i-1}$. In order for $\varphi_{i}^{NN} \in V^{\sigma_{i}}_{n(\tau,u_{i-1}), C(\tau,u_{i-1})}$ to capture the higher resolution features of $(u-u_{i-1})/|||u-u_{i-1}|||$ (as compared to the relatively lower resolution features of $(u-u_{i-2})/|||u-u_{i-2}|||$), we necessarily require that $n(\tau,u_{i-1}) > n(\tau,u_{i-2})$. Hence, by increasing the width $n_{i}$ of the networks as $i$ increases, we have $|||u-u_{j}||| \leqslant |||u-u_{0}|||\cdot (2\tau/(1-\tau))^{j}$. 

	\subsection{Neural Network Approximation of Augmented Basis Function} \label{sec:nn}
	
	In order to compute the augmented basis function $\varphi_{i}^{NN} \in V^{\sigma_{i}}_{n_{i},C_{i}} \cap B$ according to \eqref{eq:phi_j}, we first note that functions $v \in V^{\sigma_{i}}_{n_{i},C_{i}}$ do not necessarily satisfy $|||v|||=1$. We thus normalize by forming $v/|||v|||$:
	\begin{align} \label{eq:init obj}
		\max_{v \in V^{\sigma_{i}}_{n_{i},C_{i}} \cap B} \langle r(u_{i-1}),v \rangle = \max_{v \in V^{\sigma_{i}}_{n_{i},C_{i}} \backslash \{0\}} \langle r(u_{i-1}), \frac{v}{|||v|||} \rangle = \max_{v \in V^{\sigma_{i}}_{n_{i},C_{i}} \backslash \{0\}} \eta(u_{i-1},v),
	\end{align}
	
	\noindent where $\eta(u_{i-1},v)$ is as defined in \eqref{eq:eta}. Next, we note that any $v \in V^{\sigma_{i}}_{n_{i},C_{i}}$ is fully characterized by the set of parameters $\theta^{(i)} = (W^{(i)},b^{(i)},c^{(i)})$. Thus, \eqref{eq:init obj} becomes 
	\begin{align}
		\max_{v \in V^{\sigma_{i}}_{n_{i},C_{i}} \backslash \{0\}} \eta(u_{i-1},v) = \max_{\substack{\theta^{(i)}\\ \theta \neq (0, 0, 0)\\ ||\theta^{(i)}||<C_{i}}} \eta(u_{i-1},v(\theta^{(i)})).
	\end{align}
	
	\noindent In practice, we do not enforce boundedness of $\theta^{(i)}$. While boundedness can be enforced by selecting a large value of $C_{i}$ and clipping the network parameters should they exceed this value during training, we find that for the examples in Section \ref{sec:results}, the network parameters are uniformly bounded without explicitly enforcing $||\theta^{(i)}|| < C_{i}$. 
	
	The standard approach to computing optima of objective functions associated with neural networks is to apply a gradient-based optimizer to all of the parameters in $\theta$, for example \cite{raissi, kharazmi, berg, zang, weinan, lagaris, dissanayake}. In order to solve \eqref{eq:phi_j}, we instead propose a training procedure in the spirit of \cite{cyr} as follows:
	\begin{enumerate}
		\setcounter{enumi}{0}		
		\item On the first training iteration, the hidden parameters $(W^{(i)},b^{(i)})$ are initialized appropriately, which we describe on a case-by-case basis in Section \ref{sec:results}. Otherwise, the hidden parameters are updated via a gradient-based optimizer. For instance, with gradient descent, the update is described by
		\begin{align} \label{eq:gd}
			W^{(i)} &\leftarrow W^{(i)} + \alpha\nabla_{W^{(i)}}\left[\eta(u_{i-1},v(\theta^{(i)}))\right]\\
			b^{(i)} &\leftarrow b^{(i)} + \alpha\nabla_{b^{(i)}}\left[\eta(u_{i-1},v(\theta^{(i)}))\right]
		\end{align}
		
		\noindent with learning rate $\alpha$.
		
		\item We form $\Phi := \text{span}\{\sigma(x \cdot W_{j} + b_{j})\}_{j=1}^{n}$ and take the projection $v(\theta^{(i)}) \leftarrow P_{\Phi}(u-u_{i-1})$ as described by $\textsc{GalerkinLSQ}(W^{(i)}, b^{(i)}, \sigma_{i}, a, L(\cdot)-a(u_{i-1},\cdot))$.
	\end{enumerate}
	
	\begin{algorithm2e}[t!]
		\KwIn{Data $L$, bilinear operator $a$, network width $n_{i}$, activation function $\sigma_{i}$, initial approximation $u_{i-1} \in X$.}
		\KwOut{Numerical approximation $\varphi_{i}^{NN}$ to the maximizer $\argmax_{v \in B} \langle r(u_{i}),v \rangle = (u-u_{i-1})/|||u-u_{i-1}|||$, and error estimator $\eta(u_{i-1},\varphi_{i}^{NN}) \approx (u-u_{i-1})/|||u-u_{i-1}|||$}
		\BlankLine
		\BlankLine
		
		\SetKwFunction{FMain}{AugmentBasis}
    		\SetKwProg{Fn}{Function}{:}{}
    		\Fn{\FMain{$u_{j}$}}{
			Initialize hidden parameters $(W^{(i)},b^{(i)}) \in \mathbb{R}^{d\times n_{i}} \times \mathbb{R}^{n_{i}}$\footnotemark.\\
			Compute corresponding activation coefficients: $c^{(i)}\leftarrow \textsc{GalerkinLSQ}(W^{(i)},b^{(i)}, \sigma_{i}, a, L(\cdot) - a(u_{i-1},\cdot))$.\\
			\BlankLine
			\For{\text{each training epoch}}{
				Compute the update rule for $(W^{(i)},b^{(i)})$:
				\begin{align*}
					W^{(i)} &\leftarrow W^{(i)}+ \alpha\nabla_{W^{(i)}}\left[ \eta(u_{i-1},v(\theta^{(i)})) \right]\\
					b^{(i)} &\leftarrow b^{(i)}+ \alpha\nabla_{b^{(i)}}\left[ \eta(u_{i-1},v(\theta^{(i)})) \right].
				\end{align*}
				Compute corresponding activation coefficients: $c^{(i)} \leftarrow \textsc{GalerkinLSQ}(W^{(i)},b^{(i)}, \sigma_{i}, a, L(\cdot)-a(u_{i-1},\cdot))$.\\
			}
			\BlankLine
			Return $\varphi_{i}^{NN} = v(\theta^{(i)})/|||v(\theta^{(i)})|||$ and $\eta(u_{i-1},\varphi_{i}^{NN})$.\\
		}
		\textbf{EndFunction}
		\caption{Basis Function Generation and Error Estimation.}
		\label{alg:oneshot}
	\end{algorithm2e}
	\footnotetext{We use a number of existing initializations for the hidden parameters in our examples, which are detailed in Section \ref{sec:results}.}
	
	\noindent The solution of the linear system \eqref{eq:glsq} with the data $L(\cdot) - a(u_{i-1},\cdot)$  provides the expansion coefficients of the projection of the error $u-u_{i-1}$ onto $\Phi$ with respect to $a$. Specifically, by \eqref{eq:cea}, $P_{\Phi}(u-u_{i-1})$ is the best approximation to $u-u_{i-1}$ from $\Phi$. 
	
%
	
	The training procedure for $\varphi_{i}^{NN}$ is summarized in Algorithm \ref{alg:oneshot}.

	\subsubsection{Implementation Details}
	Most problems in which the bilinear operator $a$ is associated with the weak formulation of a PDE require the computation of $L^{2}$ inner products in $\Omega$. For instance, consider the case with $a(u,v) := (u,v)_{L^{2}(\Omega)}$ and $L(v) := (f,v)_{L^{2}(\Omega)}$ where $f \in L^{2}(\Omega)$. In order to compute the objective function $\eta(u_{0},v(\theta)) = \langle r(u_{0}),v(\theta) \rangle / |||v(\theta)|||$, we approximate the associated integrals by employing a quadrature rule consisting of nodes $\{x_{i}\}_{i=1}^{n_{G}} \subset \Omega$ and weights $\{w_{i}\}_{i=1}^{n_{G}}$:
	\begin{align*}
		\int_{\Omega} v(x)\;dx \approx \sum_{i=1}^{n_{G}} w_{i}v(x_{i}).
	\end{align*}
	
	\noindent We refer the reader to \cite{davis} for further details on the derivation, accuracy, and convergence of such quadrature rules. 
	
	The objective function in the above example is approximated by
	\begin{align*}
		\frac{L(v(\theta)) - a(u_{0},v(\theta))}{|||v(\theta)|||} \approx \frac{\sum_{i=1}^{n_{G}} w_{i}[ f(x_{i})v(x_{i};\theta) - u_{0}(x_{i})v(x_{i};\theta) ]} {\sqrt{ \sum_{i=1}^{n_{G}} w_{i}v(x_{i};\theta)^{2}}}.
	\end{align*}
	
	\noindent In all examples in Section \ref{sec:results}, a single step of the Adam optimizer \cite{adam} with full batch is employed to update the hidden parameters $\{W,b\}$ at each training epoch following Algorithm \ref{alg:adaptive}. In other words, the training data essentially consists of the set of $n_{G}$ quadrature nodes which are used to evaluate the loss function.
	
	For most numerical examples, the activation function $\sigma_{i}$ at Galerkin iteration $i$ is adaptive in the sense that we utilize a scaling factor as a hyperparameter \cite{jagtap}. Given a fixed nonpolynomial activation function $\sigma : \mathbb{R} \to \mathbb{R}$, we define $\sigma_{i}(t) = \sigma(\beta_{i}t)$, where $\{\beta_{i}\}$ is a nondecreasing sequence in $\mathbb{R}_{+}$. As $\beta_{i}$ grows larger, the gradient of $\sigma_{i}$ generally increases, which allows for better resolution of the fine-scale error features as we shall demonstrate in Section \ref{sec:results}. Finally, in most of the proceeding examples, the network widths $\{n_{i}\}$ follow a geometric progression of the form $n_{i} = N \cdot r^{i-1}$, where $N$ is the initial width for the first Galerkin iteration $r$ is the geometric ratio. The learning rate is fixed for each Galerkin iteration $i$ and follows an exponential decay of the form $\alpha_{i} = A \cdot \rho^{-(i-1)}$, where $A$ is the initial learning rate for the first Galerkin iteration and $\rho$ is the rate of decay. Standard neural network approaches where one trains only a single fixed-architecture network commonly utliize a decreasing learning rate schedule as the training epochs progress. In our approach, we only decrease the learning rate \emph{between} each Galerkin iteration to compensate for the increasing widths $n_{i}$ as described in Section \ref{sec:widthselection}.
	\begin{figure}[t!]
		\centering
		\begin{subfigure}{.47\textwidth}
			\centering
			\includegraphics[width=\linewidth]{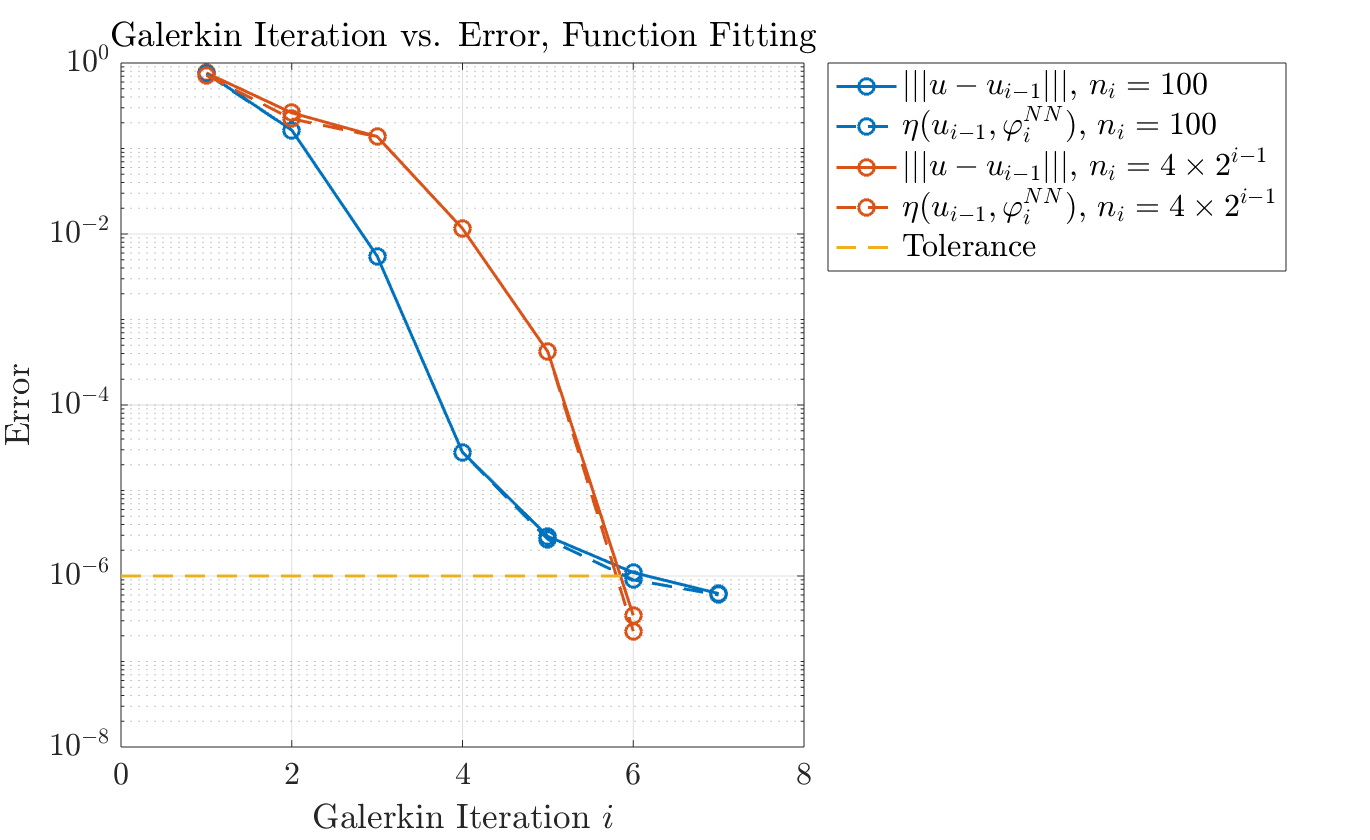}
			\caption{}
			\label{fig:function fitting1}
		\end{subfigure}
		\quad
		\begin{subfigure}{.47\textwidth}
			\centering
			\includegraphics[width=\linewidth]{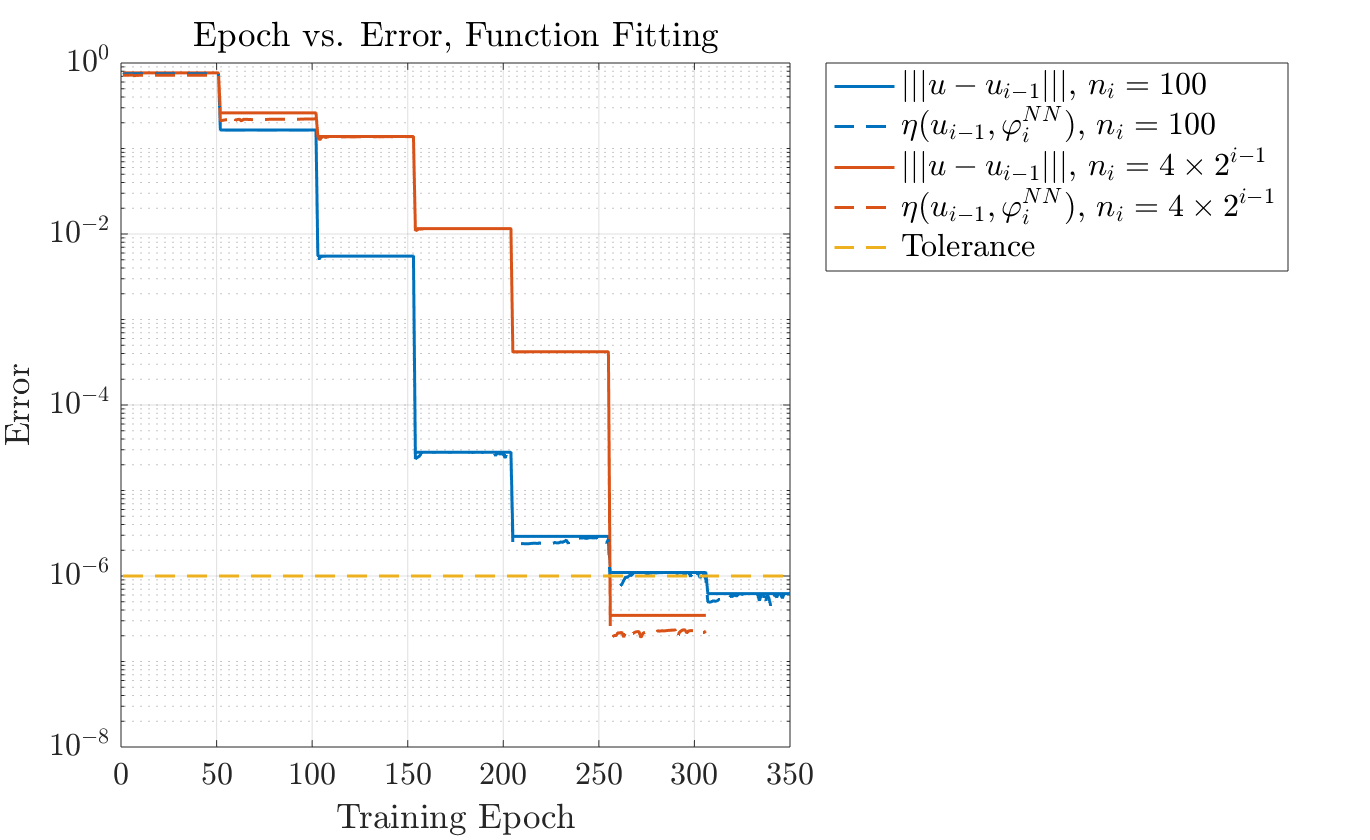}
			\caption{}
			\label{fig:function fitting2}
		\end{subfigure}
		\caption{(a) Estimated and true $L^{2}$ error at each Galerkin iteration. (b) The piecewise constant segments (solid lines) denote the true errors for each Galerkin iteration $i$, while the dashed red line (resp. blue line) denotes the progress of the loss function in approximating the true errors within each Galerkin iteration for increasing network widths (resp. constant network widths). The $x$-axis thus denotes the cumulative training epoch over all Galerkin iterations.}
		\label{fig:function fitting}
	\end{figure}

	The main computational expense in Algorithm \ref{alg:adaptive} is the computation of the activation coefficients $c_{j}$ defined by the dense linear system \eqref{eq:glsq} which costs $\mathcal{O}(n_{i}^{3})$ operations for Galerkin iteration $i$. If the network size $n_{0}$ is increased by a fixed factor $\mu > 1$ at each iteration, then the total computational cost is of order
	\begin{align*}
		\sum_{i=1}^{k} \mu^{3(i-1)}n_{0}^{3} = n_{0}^{3}\frac{\mu^{3k} - 1}{\mu^{3}-1} \approx n_{F}^{3},
	\end{align*} 
	
	\noindent where $n_{F} = n_{0}\mu^{k-1}$ is the number of neurons in the final network. In other words, the computational cost associated with obtaining $c$ for the entire sequence of networks is comparable to the cost of a single network of width $n_{F}$.

	\section{Applications} \label{sec:results}
	We apply Algorithm \ref{alg:adaptive} to a number of problems in one and two dimensions, demonstrating the robustness of our algorithm for several different classes of problems as well as the robustness of the error indicator $\eta(u_{i-1},\varphi_{i}^{NN})$. 
	
	\subsection{Function Fitting} \label{sec:function fitting}
	We consider the following illuminating example in order to investigate whether stagnation of the error $|||u-u_{i}|||$ occurs when $n_{i} \equiv n$ is fixed: given $f \in L^{2}(\Omega)$, $\Omega = (0,1)$, let $a(u,v) := (u,v)_{L^{2}(\Omega)}$ and $L(v) := (f,v)_{L^{2}(\Omega)}$ with $f$ being the first four terms of the Fourier series for the square wave:
	\begin{align*}
		f(x) = \sin(x) + \frac{1}{3}\sin(3\pi x) + \frac{1}{5}\sin(5\pi x) + \frac{1}{7}\sin(7\pi x).
	\end{align*}
	
	\noindent This is a simple function fitting problem whose exact solution is $u=f$. For this and all subsequent examples, we set $u_{0} = 0$. The loss function is approximated by
	\begin{align*}
		\frac{L(v(\theta)) - a(u_{0},v(\theta))}{|||v(\theta)|||} \approx \frac{\sum_{i=1}^{n_{G}} w_{i}[ f(x_{i})v(x_{i};\theta) - u_{0}(x_{i})v(x_{i};\theta) ]} {\sqrt{ \sum_{i=1}^{n_{G}} w_{i}v(x_{i};\theta)^{2}}}.
	\end{align*} 
		
	\begin{figure}[t!]
		\centering
		\begin{subfigure}{.47\textwidth}
			\centering
			\includegraphics[width=2.0in]{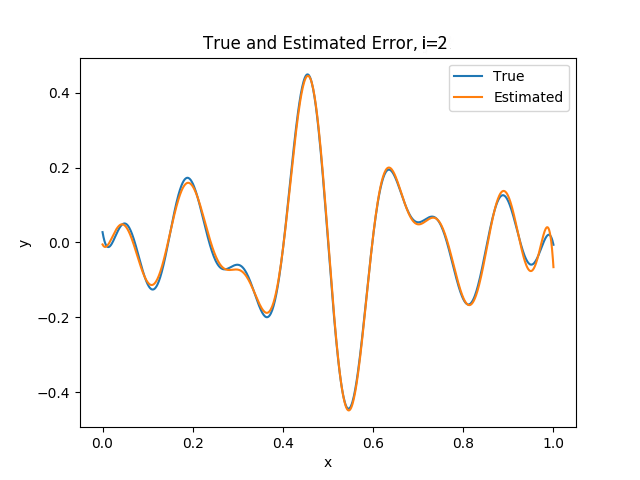}
			\caption{}
			\label{fig:function fitting error1}
		\end{subfigure}
		\quad
		\begin{subfigure}{.47\textwidth}
			\centering
			\includegraphics[width=2.0in]{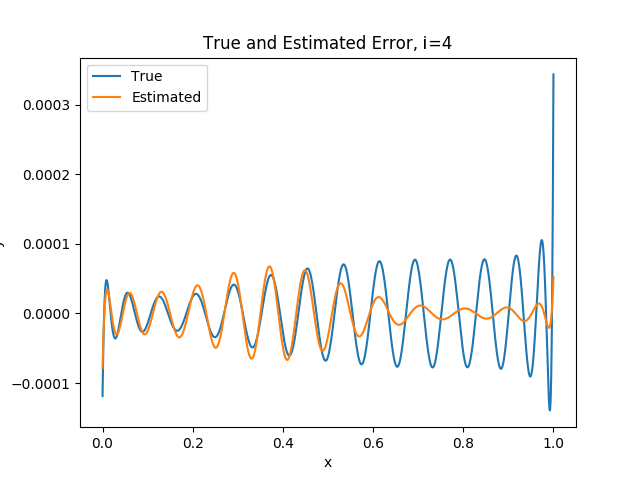}
			\caption{}
			\label{fig:function fitting error2}
		\end{subfigure}
		\caption{(a) True error $u-u_{i-1}$ and approximate error $\varphi_{j}^{NN}$ for $i=2$. (b) True error $u-u_{i-1}$ and approximate error $\varphi_{i}^{NN}$ for $i=4$ with fixed width $n_{i} = 100$.}
		\label{fig:function fitting error}
	\end{figure}
	
	\begin{figure}[b!]
		\centering
		\begin{subfigure}{.47\textwidth}
			\centering
			\includegraphics[width=2.4in]{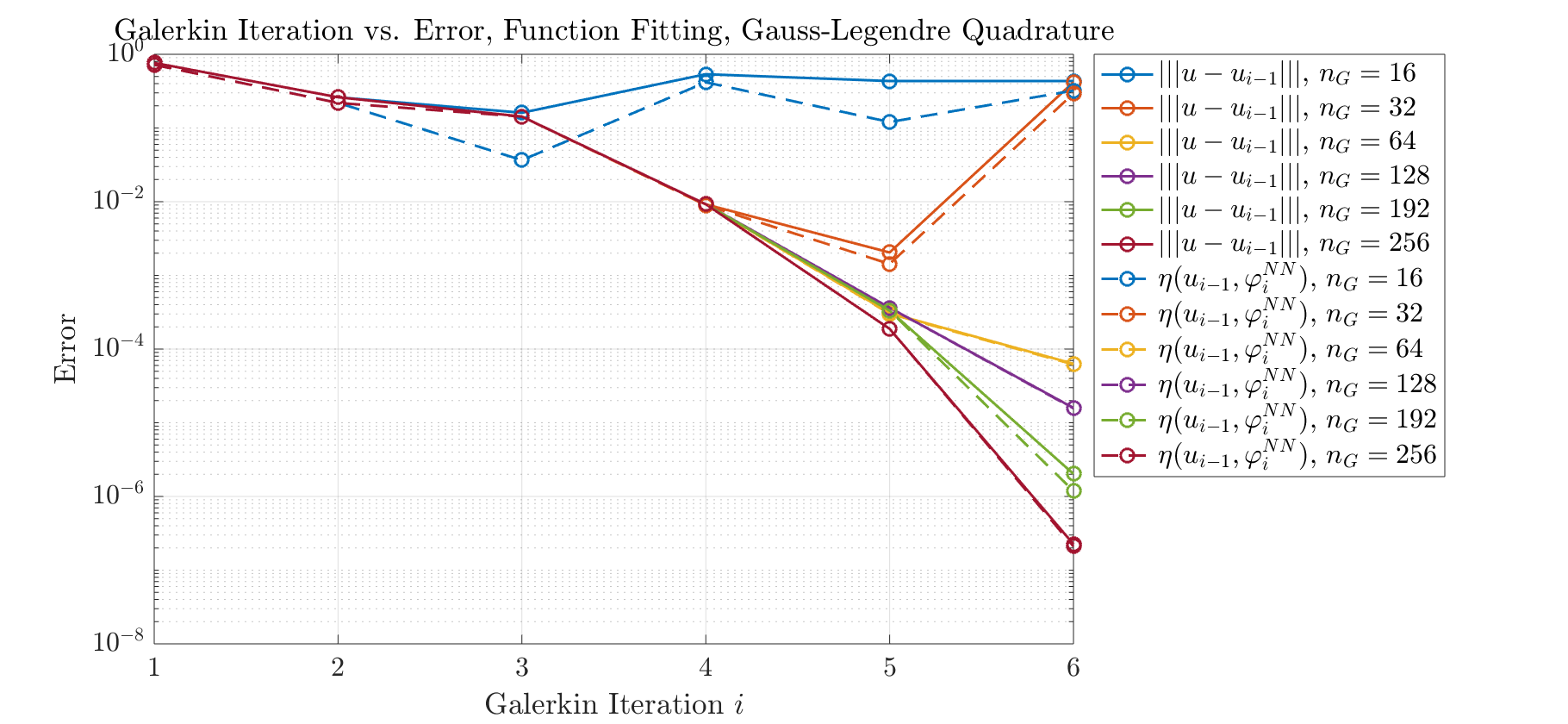}
			\caption{}
			\label{fig:function fitting gausslegendre}
		\end{subfigure}
		\quad
		\begin{subfigure}{.47\textwidth}
			\centering
			\includegraphics[width=2.4in]{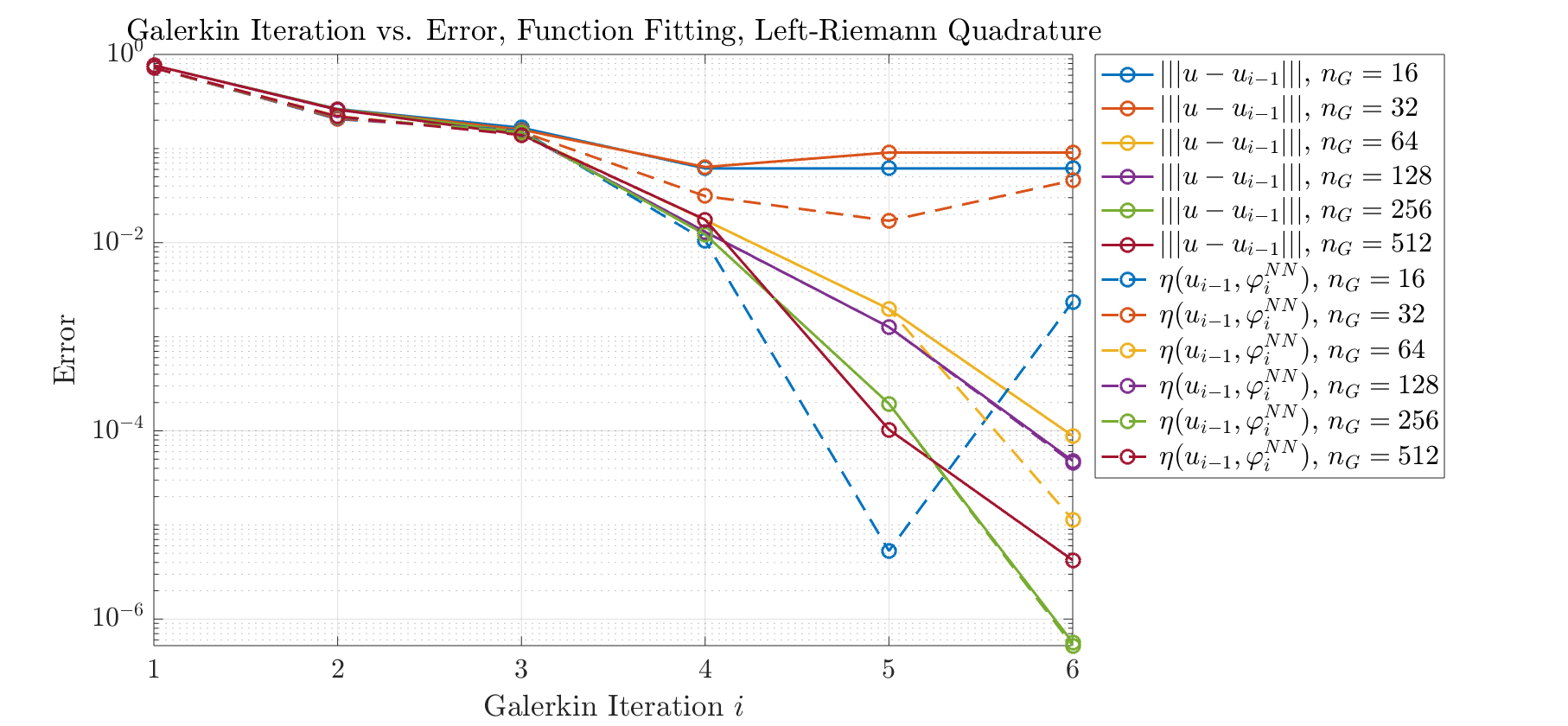}
			\caption{}
			\label{fig:function fitting riemann}
		\end{subfigure}
		\caption{(a) Error at each Galerkin iteration $i$ using a Gauss-Legendre quadrature rule with varying number of nodes (b) the analogous results using a left Riemann-sum with varying number of nodes.}
		\label{fig:function fitting quadrature}
	\end{figure}
	
	We use the hyperbolic tangent activation function $\sigma_{i}(t) = \tanh(\beta_{i}t)$, $\beta_{i} = 1+3(i-1)$ at each Galerkin iteration and consider both the case when successive networks have fixed widths $n_{i}=100$ for all $i$ and the case when successive networks have increasing widths $n_{i}=4\times 2^{i-1}$. We set $\texttt{tol} = 10^{-6}$ and adaptively construct a Galerkin subspace that achieves an approximation with $\eta(u_{i-1},\varphi_{i}^{NN}) < \texttt{tol}$. The hidden parameters are initialized such that $W_{j}^{(i)} = 1$ and $b_{j}^{(i)} = -jh$, $h=1/n_{i}$ i.e. the biases are spaced uniformly throughout $\Omega$. To evaluate the objective function and train the network, we employ a fixed Gauss-Legendre quadrature rule with 512 nodes to approximate all inner products. A separate Gauss-Legendre quadrature rule with 1000 nodes is used for validation, i.e. to compute the true errors. 
	\begin{figure}[t!]
		\centering
		\begin{subfigure}{.47\textwidth}
			\centering
			\includegraphics[width=2.0in]{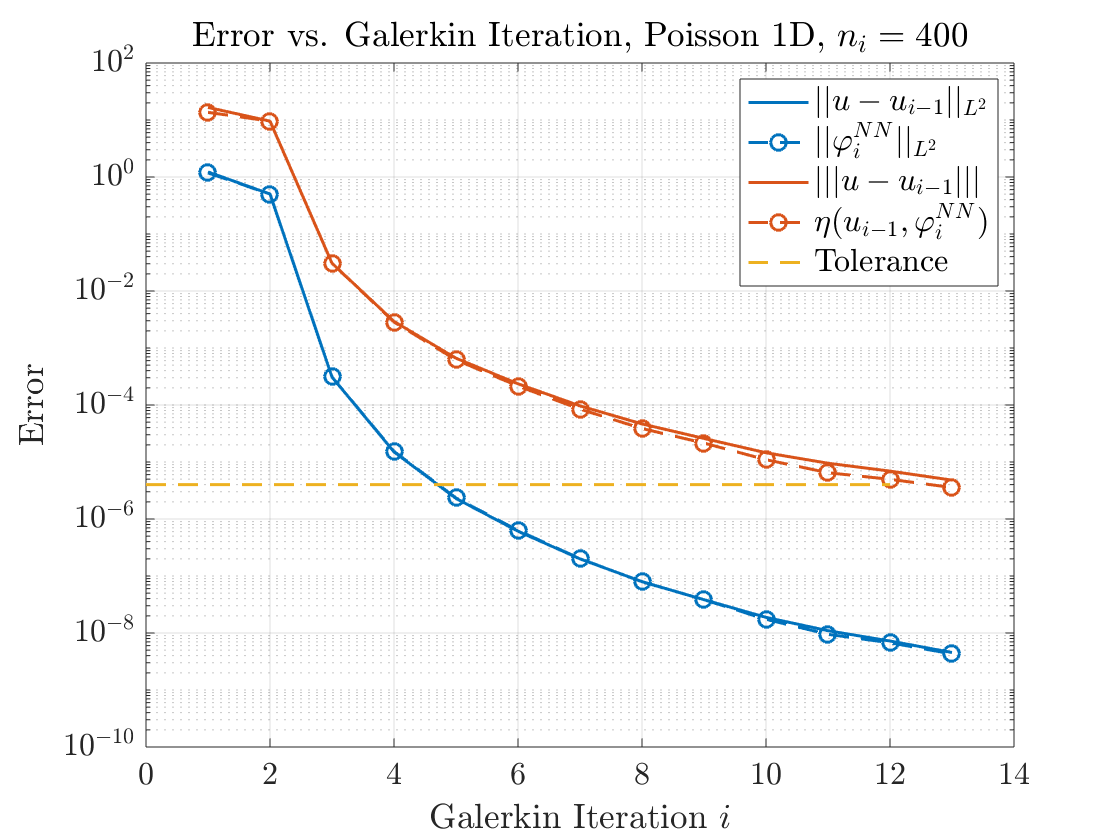}
			\caption{}
		\end{subfigure}
		\begin{subfigure}{.47\textwidth}
			\centering
			\includegraphics[width=2.0in]{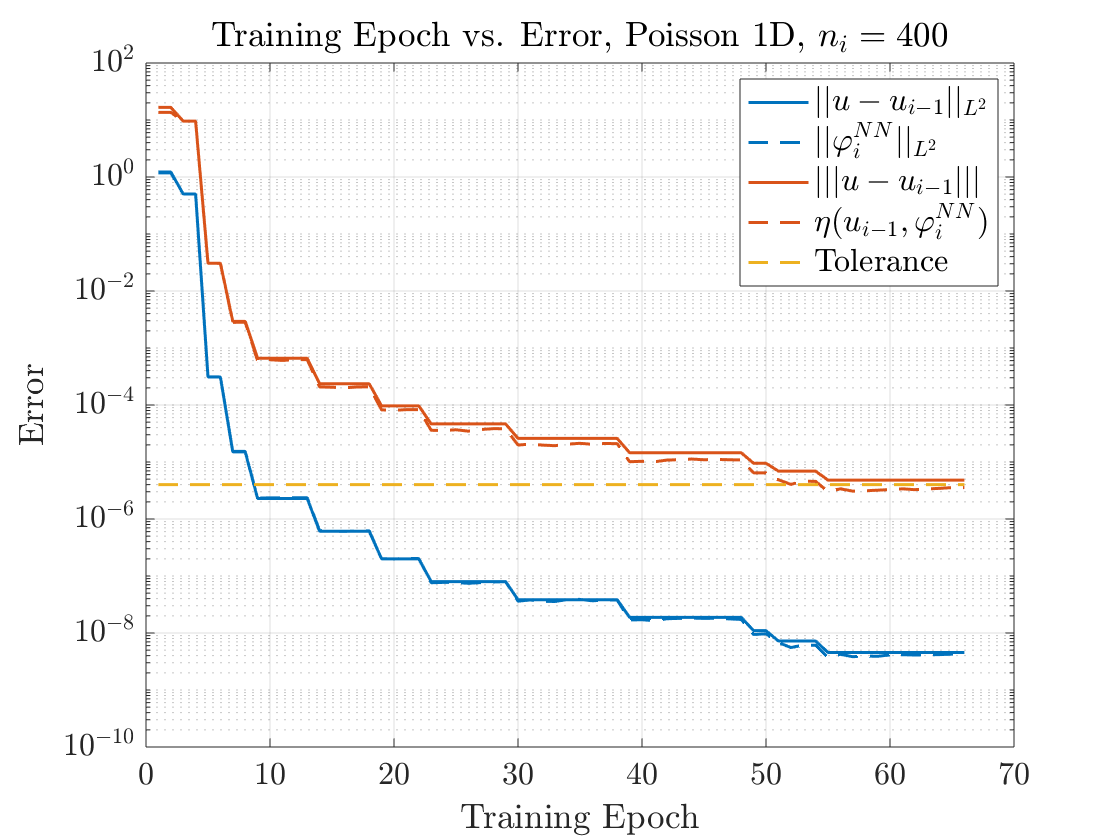}
			\caption{}
		\end{subfigure}
		\begin{subfigure}{.47\textwidth}
			\centering
			\includegraphics[width=2.0in]{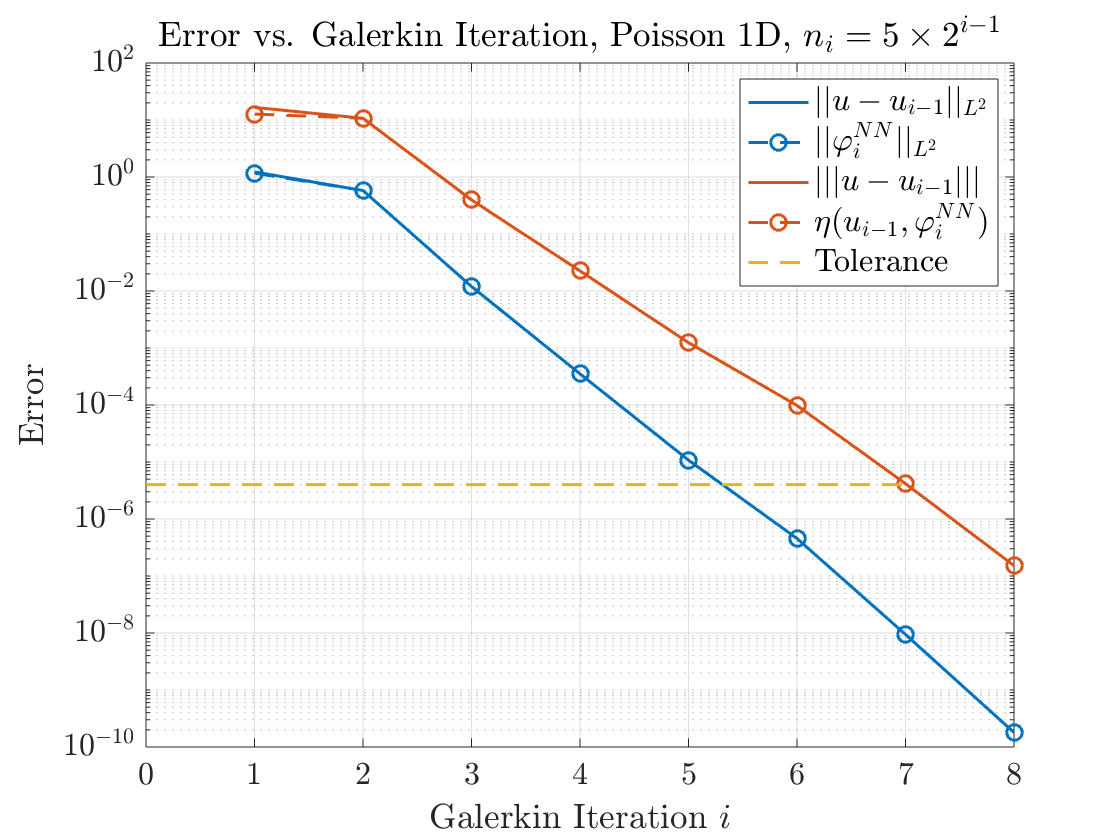}
			\caption{}
		\end{subfigure}
		\begin{subfigure}{.47\textwidth}
			\centering
			\includegraphics[width=2.0in]{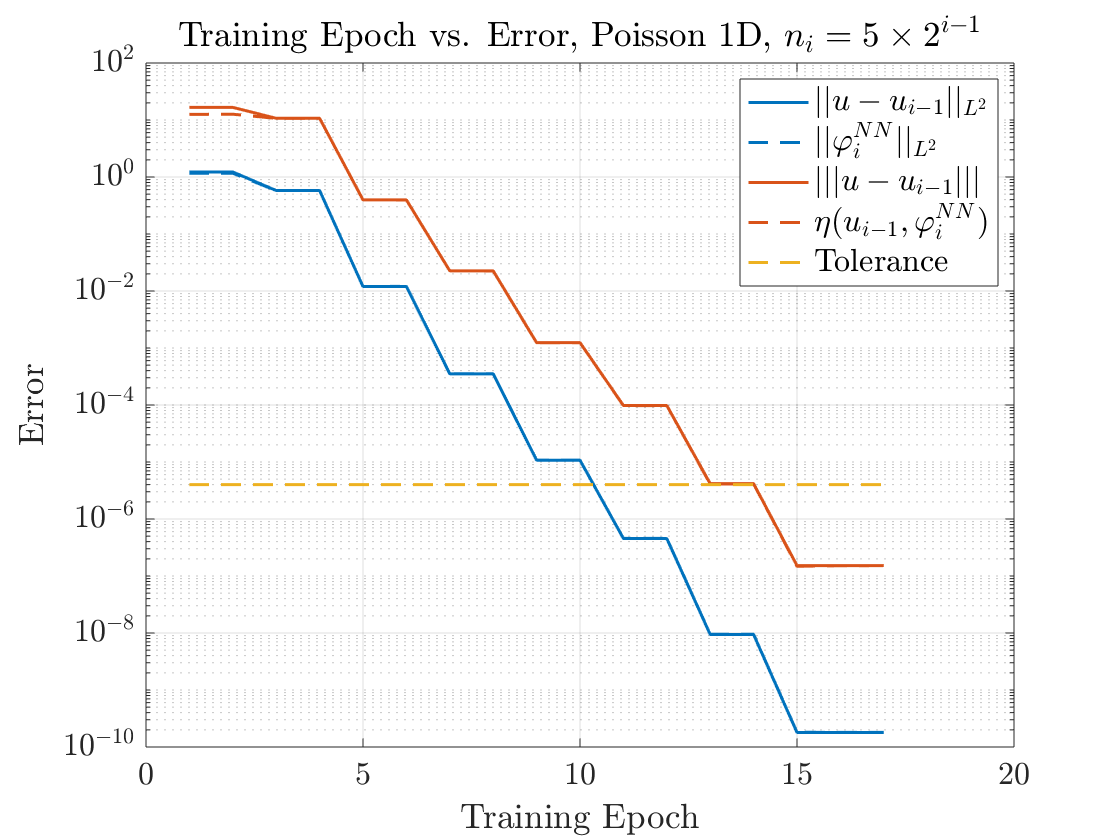}
			\caption{}
		\end{subfigure}
		\caption{Displacement of a string (Poisson equation in 1D). (a) Estimated and true error in the $L^{2}$ and energy norms at each Galerkin iteration, $n_{i}=400$ for all $i$. (b) The piecewise constant segments (solid lines) denote the true errors for each Galerkin iteration $i$, while the dashed red line (resp. blue line) denotes the progress of the loss function (resp. $L^{2}$ error estimator $||\varphi_{i}^{NN}||_{L^{2}}$) in approximating the true errors within the each Galerkin iteration with $n_{i}=400$. The $x$-axis thus denotes the cumulative training epoch over all Galerkin iterations. (c) and (d) The analogous results for $n_{i} = 5\times 2^{i-1}$.}
		\label{fig:1d error galerkin}
	\end{figure}
	
	Figure \ref{fig:function fitting1} shows the true $L^{2}$ error at each Galerkin iteration along with the estimator $\eta(u_{i-1},\varphi_{i}^{NN})$ while Figure \ref{fig:function fitting} shows the true and estimated $L^{2}$ error at each epoch. Figure \ref{fig:function fitting error} shows the true error $u-u_{i-1}$ and estimated error $\varphi_{i}^{NN}$ for $i=2$ and $i=4$ with $n_{i}=100$. It is apparent that initially, the low frequency components of the error are learned, with later iterations learning the high frequency error components. For $i=2$, the low frequency error components are sufficiently learned by the network. However, for $i=4$, the network lacks sufficient capacity to learn the high frequency error components, thus leading to inadequate training of the network and stagnation in the error. On the other hand, we observe from Figure \ref{fig:function fitting} that there is no stagnation of the error in the later Galerkin iterations if $n_{i}$ is increased with $i$. Moreover, the case when $n_{i}$ is increased at each Galerkin iteration eventually achieves the same accuracy as the case when $n_{i}$ is fixed at each Galerkin iteration while also achieving better computational efficiency due to utilizing initial networks with smaller widths.
	
	Of course, the quadrature rule used to approximate the loss function must be chosen sufficiently accurately. Figure \ref{fig:function fitting quadrature} shows the convergence results obtained as the number of Gauss-Legendre quadrature points increases as well as a comparison with a left-Riemann sum quadrature rule. In each case, we observe that if the number of nodes is too small, the error either stagnates or increases eventually due to the inability of the quadrature rule to adequately evaluate the inner products in the loss function and \eqref{eq:glsq} and \eqref{eq:galerkin solve}.

	\subsection{Second Order Problems}
	\subsubsection{Displacement of a String} 
	We now illustrate the approach for a simple linear ODE:
	\begin{align*}
		\begin{dcases}
			-u'' = f, &\text{in}\;\Omega = (0,1)\\
			u+\varepsilon\partial_{n}u = 0, &\text{on}\;\partial\Omega,
		\end{dcases}
	\end{align*}
		
	\begin{figure}[t!]
		\centering
		\begin{subfigure}{.47\textwidth}
			\includegraphics[width=2.0in]{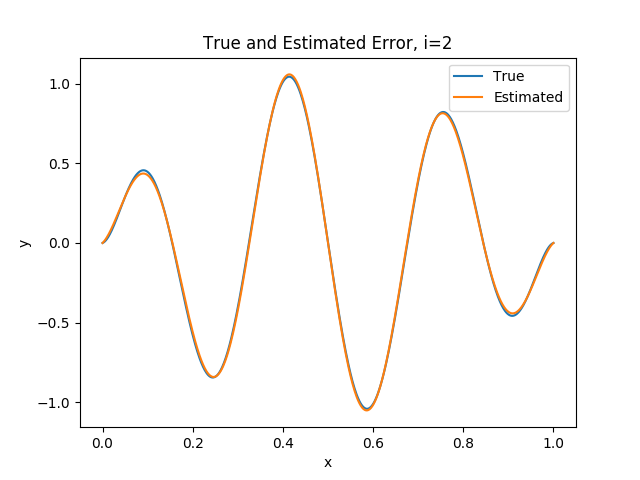}
			\caption{}
		\end{subfigure}
		\quad
		\begin{subfigure}{.47\textwidth}
			\includegraphics[width=2.0in]{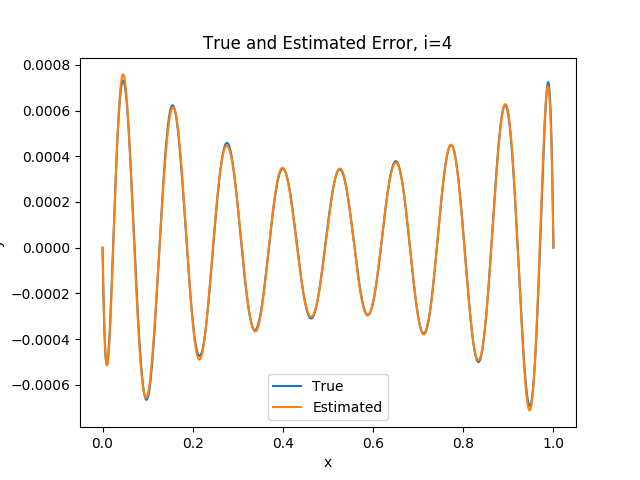}
			\caption{}
		\end{subfigure}
		\quad
		\begin{subfigure}{.47\textwidth}
			\includegraphics[width=2.0in]{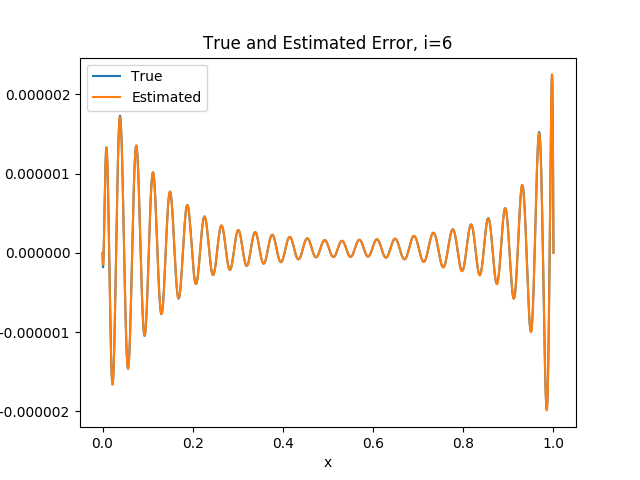}
			\caption{}
		\end{subfigure}
		\quad
		\begin{subfigure}{.47\textwidth}
			\includegraphics[width=2.0in]{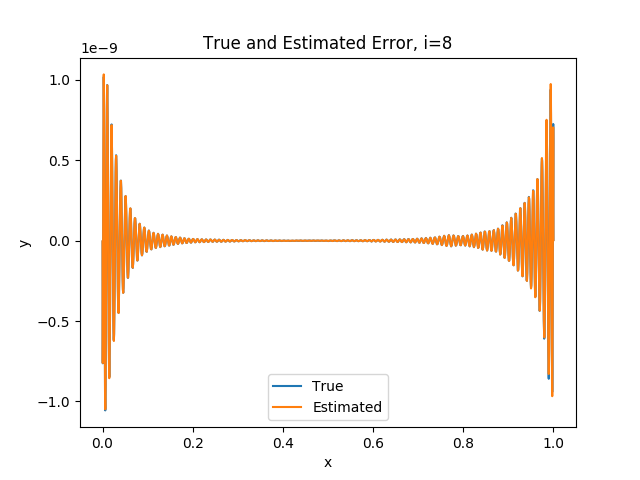}
			\caption{}
		\end{subfigure}
		\caption{Displacement of a string (Poisson equation in 1D). True error $u-u_{i-1}$ and estimated error $\varphi_{i}^{NN}$ for $i=2,4,6,8$.}
		\label{fig:poisson1d plots}
	\end{figure}
	
	\noindent where $\varepsilon > 0$ and $f(x)=(2\pi)^{2}\sin(2\pi x) + (4\pi)^{2}\sin(4\pi x) + (6\pi)^{2}\sin(6\pi x)$, with $f$ inducing a linear functional given by the rule $L(v) := (f,v)_{\Omega}$. The variational formulation is posed on the Hilbert space $X=H^{1}(\Omega)$ corresponding to the bilinear operator $a(u,v) = (u',v')_{\Omega} + \varepsilon^{-1}(u(0)v(0) + u(1)v(1))$. One can easily verify that $a$ is symmetric and satisfies the conditions in \eqref{eq:condition}. Observe that as $\varepsilon \to 0$, the derivative terms in the boundary condition vanish and we obtain $u(0) \to 0$ and $u(1) \to 0$, recovering the standard Poisson equation with homogeneous Dirichlet boundary conditions. The parameter $\varepsilon$ may thus be regarded as a penalization term which approximately enforces the homogeneous boundary condition $u(0)=u(1)=0$. Alternatively, as $\varepsilon \to \infty$ we obtain $\partial_{n}u \to 0$ on $\partial\Omega$, and $\varepsilon$ may be regarded as approximately imposing a homogeneous Neumann boundary condition. The true solution is 
	\begin{align*}
		u(x) = \sin(2\pi x) + \sin(4\pi x) + \sin(6\pi x) + \frac{1}{1+2\varepsilon}\left( -24\pi\varepsilon x + 12\pi\varepsilon \right).
	\end{align*}
	
	\noindent The corresponding loss function is approximated by
	\begin{align*}
		\frac{L(v(\theta)) - a(u_{0},v(\theta))}{|||v(\theta)|||} \approx \frac{L_{1} - L_{2} - L_{3}} {\sqrt{ \sum_{i=1}^{n_{G}} w_{i}v'(x_{i};\theta)^{2} + \varepsilon^{-1}(v(0;\theta)^{2} + v(1;\theta)^{2})}}
	\end{align*} 
	
	\noindent where $L_{1} = \sum_{i=1}^{n_{G}} w_{i} f(x_{i})v(x_{i};\theta)$, $L_{2} = \sum_{i=1}^{n_{G}} w_{i} u_{0}'(x_{i})v'(x_{i};\theta)$, and $L_{3} = \varepsilon^{-1}[u_{0}(0)\cdot v(0;\theta) + u_{0}(1)v(1;\theta)]$. 
	
	\begin{table}
		\begin{tabular}{c|ccccccccccc}
			$i$ & 1 & 2 & 3 & 4 & 5 & 6 & 7 & 8 & 9 & 10\\ \hline
			$\text{cond}(K^{(i)})$ & 1.00 & 1.00 & 1.00 & 1.00 & 1.02 & 1.03 & 1.07 & 1.08 & 1.09 & 1.18\\
		\end{tabular}
		\caption{Displacement of a string. Condition number for the Galerkin matrix at each Galerkin iteration $i$.}
		\label{tab:poisson1d cond}
	\end{table}
	
	\begin{figure}[t!]
		\centering
		\includegraphics[width=1.8in]{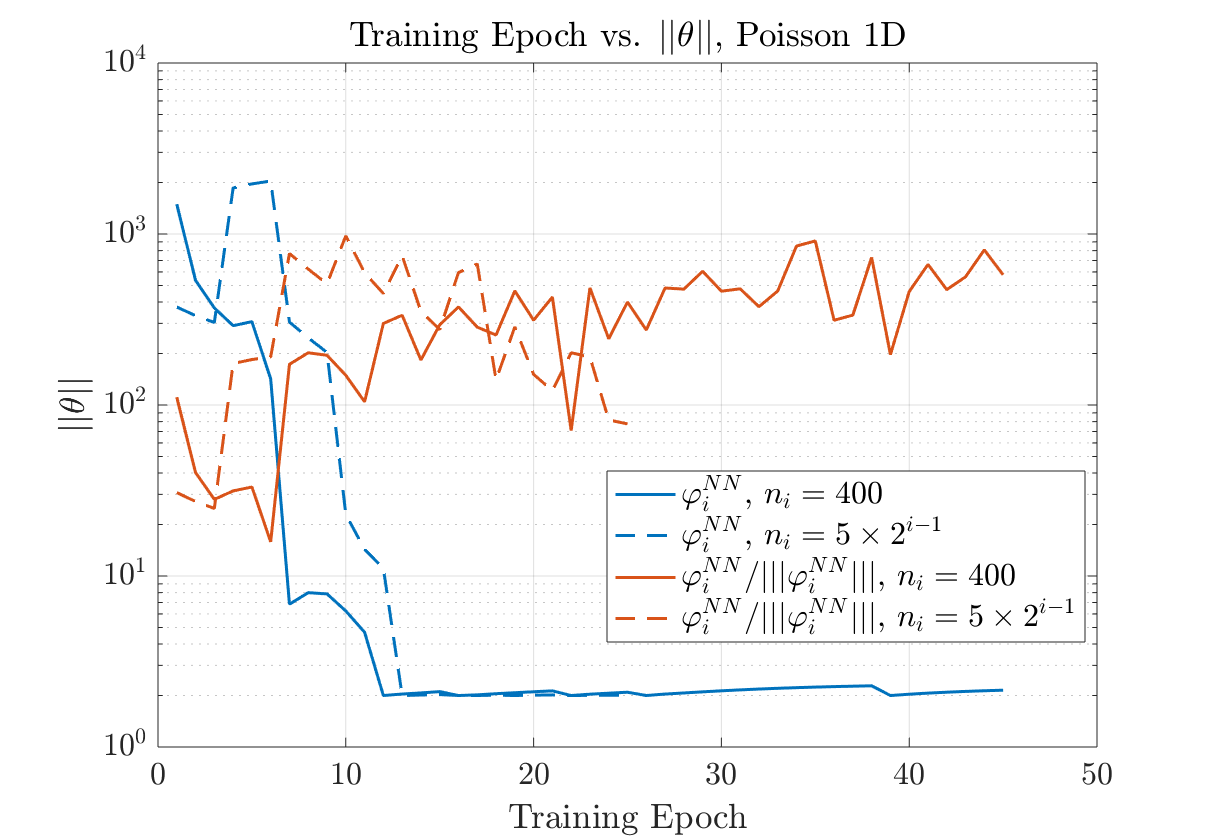}
		\caption{Displacement of a string. The total norm of the network parameters for $\varphi_{i}^{NN}$ and $\varphi_{i}^{NN}/|||\varphi_{i}^{NN}|||$.}
		\label{fig:poisson1d weights}
	\end{figure}

	\begin{figure}[t!]
		\centering
		\begin{subfigure}{.30\textwidth}
			\centering
			\includegraphics[width=1.5in]{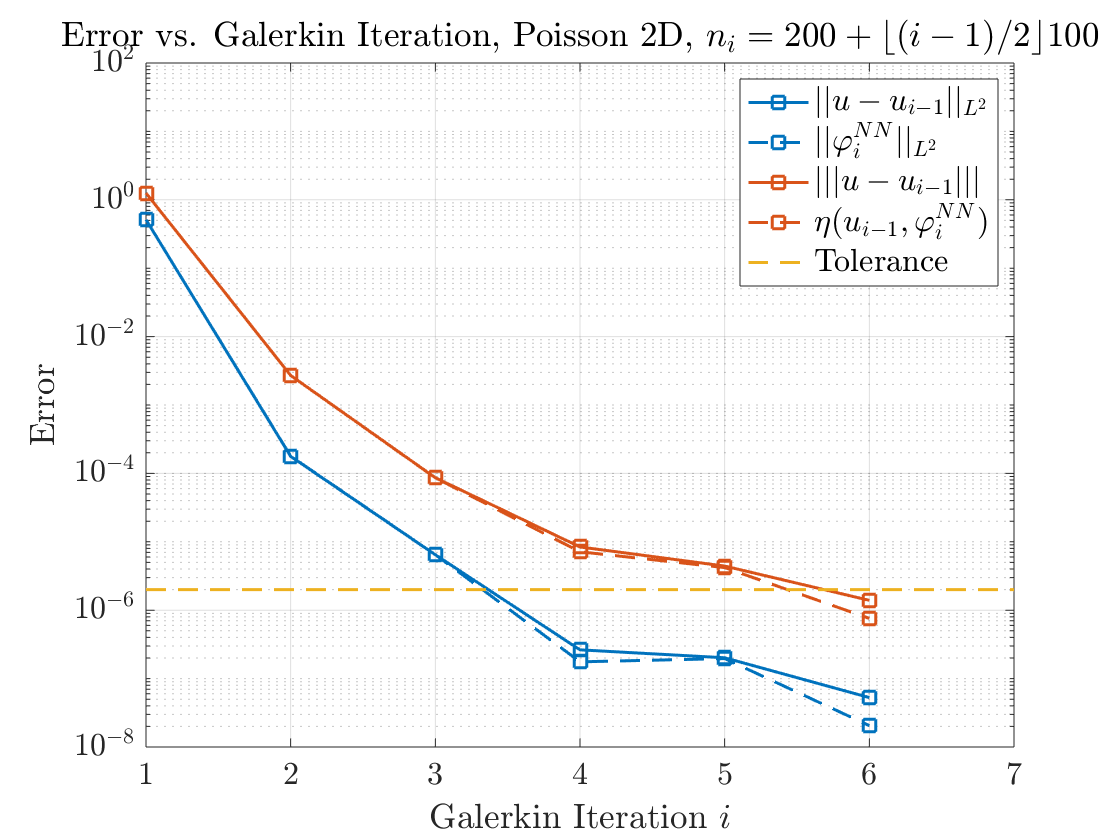}
			\caption{}
			\label{fig:2d error galerkin}
		\end{subfigure}
		\quad
		\begin{subfigure}{.30\textwidth}
			\centering
			\includegraphics[width=1.5in]{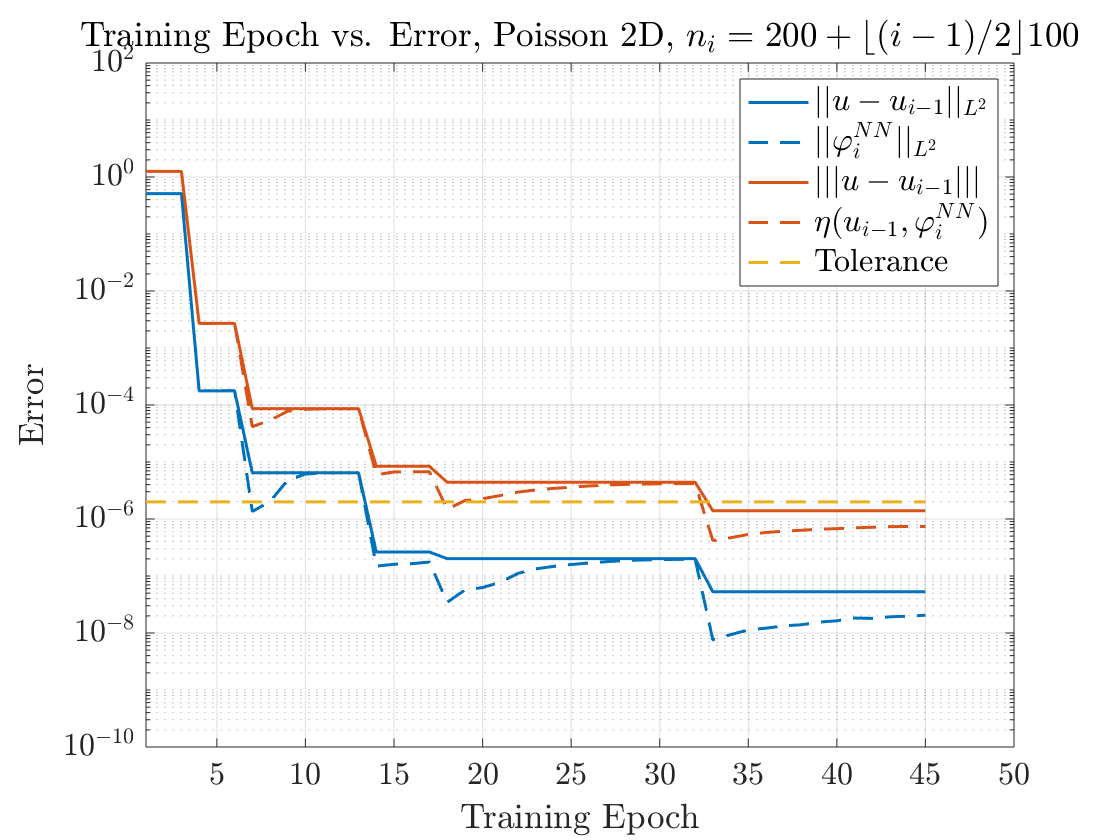}
			\caption{}
			\label{fig:2d error epoch}
		\end{subfigure}
		\quad
		\begin{subfigure}{.30\textwidth}
			\centering
			\includegraphics[width=1.5in]{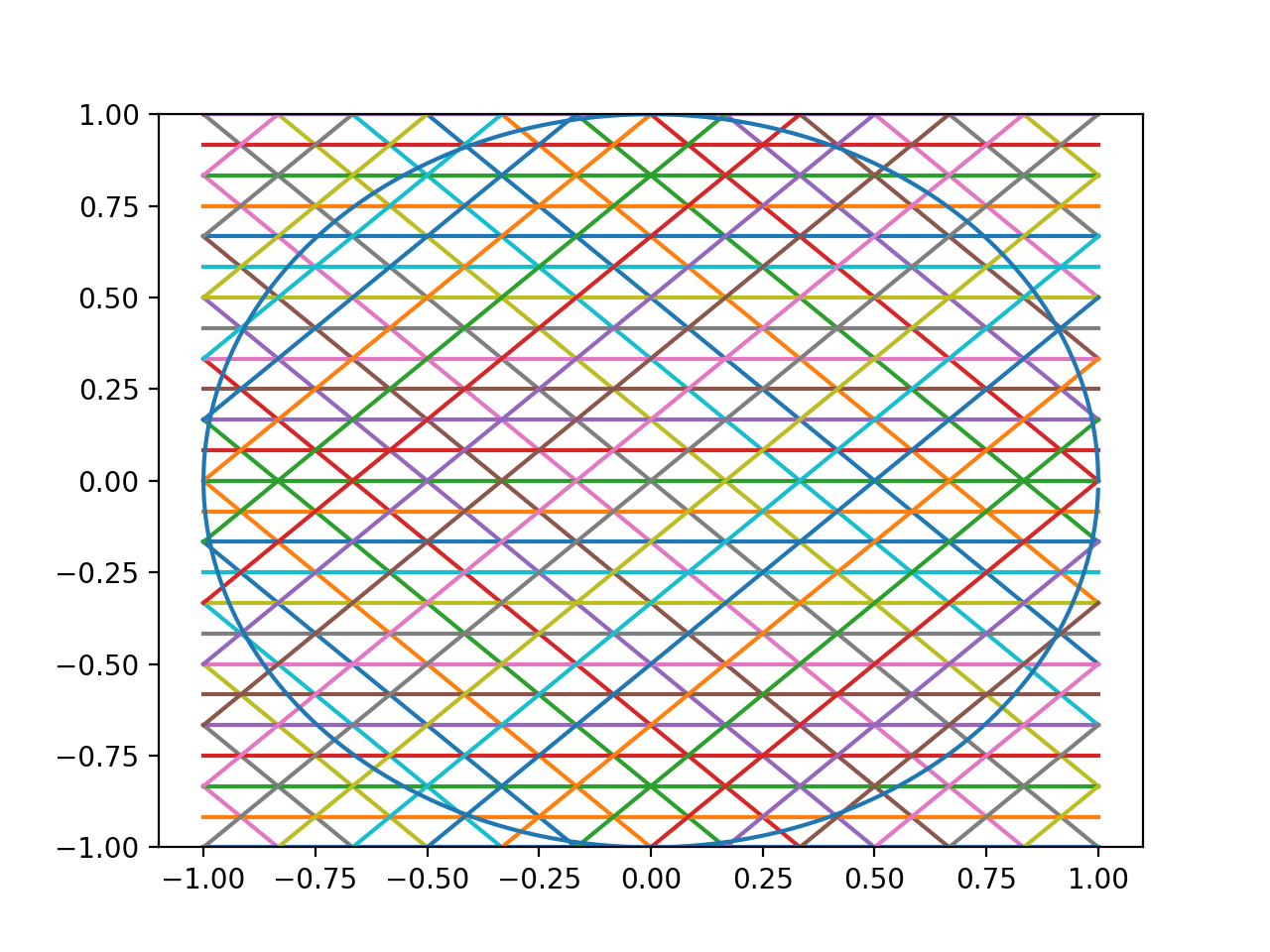}
			\caption{}
		\end{subfigure}
		\caption{Displacement of a membrane (Poisson equation in 2D). Estimated and exact error in the $L^{2}$ and energy norms at each Galerkin iteration (a) and epoch (b). Initial hyperplanes (c).}
		\label{fig:2d errors}
	\end{figure}
	
	The tolerance is set to $\texttt{tol} = 2\times 10^{-6}$, and $\varepsilon = 10^{-4}$. The network architecture for Galerkin iteration $i$ is as follows: $n_{i} = 400$, $\sigma_{i}(t) = \tanh(\beta_{i}t)$, $\beta_{i} = i$. The learning rate is $\alpha = 2 \times 10^{-2}$. The hidden parameters are initialized in the same manner as in Section \ref{sec:function fitting}. For comparison, we shall also consider the set of network architectures given by $n_{i} = 5\times 2^{i-1}$ and learning rate $\alpha = (2\times 10^{-2})/1.1^{i-1}$. To evaluate the objective function and train the network, we employ a fixed Gauss-Legendre quadrature rule with 512 nodes to approximate all inner products. A separate Gauss-Legendre quadrature rule with 1000 nodes is used for validation.
	\begin{figure}[t!]
		\centering
		\begin{subfigure}{.47\textwidth}
			\includegraphics[width=2.0in]{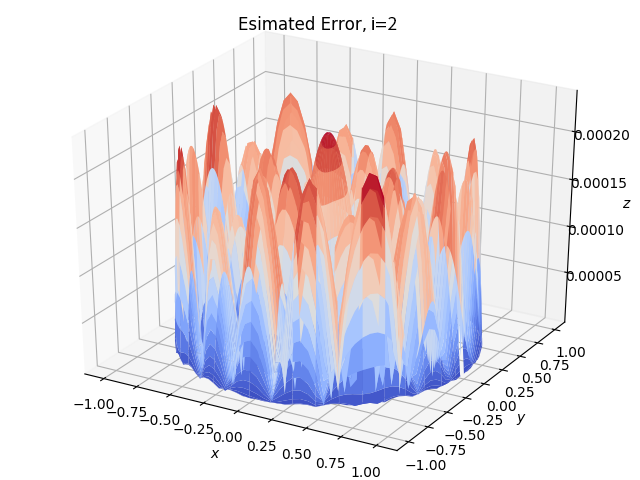}
			\caption{}
		\end{subfigure}
		\quad
		\begin{subfigure}{.47\textwidth}
			\includegraphics[width=2.0in]{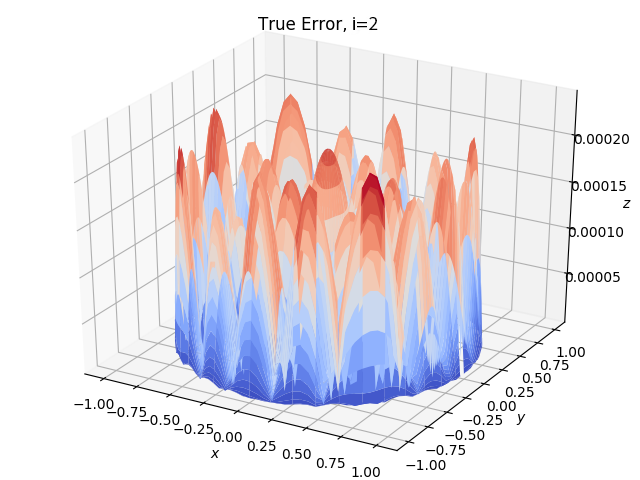}
			\caption{}
		\end{subfigure}
		\quad
		\begin{subfigure}{.47\textwidth}
			\includegraphics[width=2.0in]{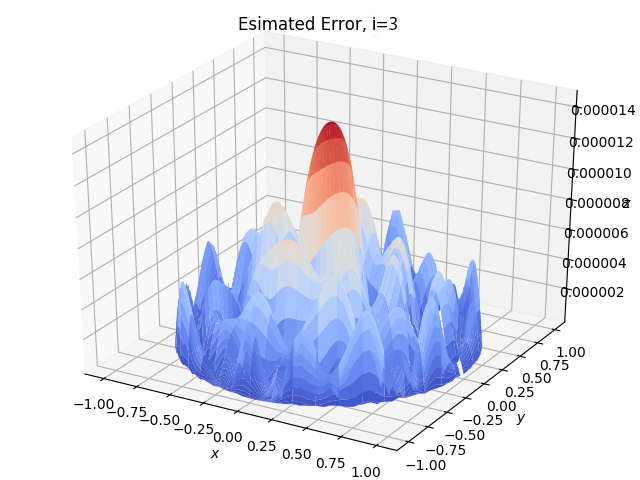}
			\caption{}
		\end{subfigure}
		\quad
		\begin{subfigure}{.47\textwidth}
			\includegraphics[width=2.0in]{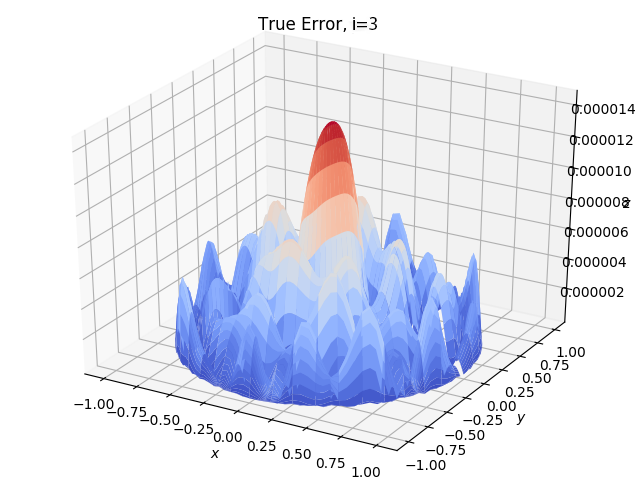}
			\caption{}
		\end{subfigure}
		\quad
		\begin{subfigure}{.47\textwidth}
			\includegraphics[width=2.0in]{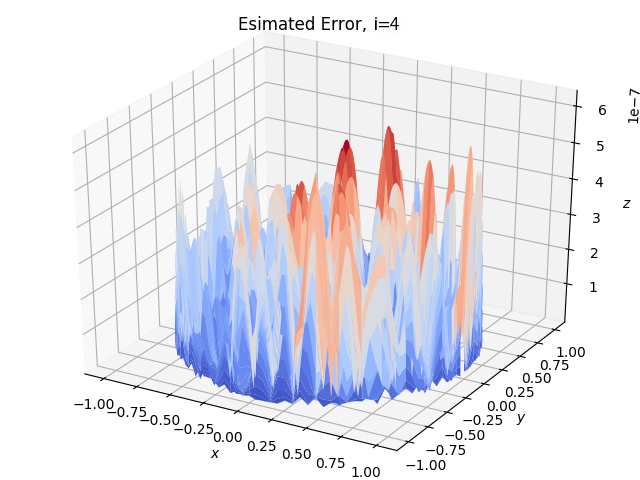}
			\caption{}
		\end{subfigure}
		\quad
		\begin{subfigure}{.47\textwidth}
			\includegraphics[width=2.0in]{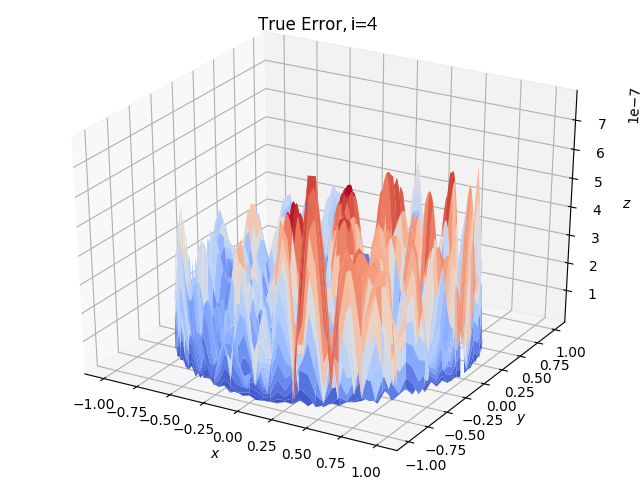}
			\caption{}
		\end{subfigure}
		\quad
		\caption{Displacement of a membrane (Poisson equation in 2D). Exact error $u-u_{i-1}$ (right column) and approximate error $\varphi_{i}^{NN}$ (left column) for $i=2,3,4$.}
		\label{fig:poisson2d plots}
	\end{figure}
	
	Figure \ref{fig:1d error galerkin} shows the true errors $||u-u_{i-1}||_{L^{2}}$ and $|||u-u_{i-1}|||$ at the end of each Galerkin iteration as well as the corresponding error estimates $||\varphi_{i}^{NN}||_{L^{2}}$ and $\eta(u_{i-1},\varphi_{i}^{NN})$. We also provide the analogous results after each training epoch. Figure \ref{fig:poisson1d plots} shows the exact error $u-u_{i-1}$ as well as the maximizer $\varphi_{i}^{NN}$ at several stages of the algorithm. We observe first that the initial Galerkin iterations reduce the error substantially, with further decreases beginning to stagnate as $i$ grows larger whenever $n_{i}$ is constant; stagnation is not an issue when $n_{i}$ is increased with $i$. Moreover, both estimators $\eta(u_{i-1},\varphi_{i}^{NN})$ and $||\varphi_{i}^{NN}||_{L^{2}}$ provide accurate estimates of the error, with negligible discrepancies beginning to show in the final Galerkin iterations. Figure \ref{fig:poisson1d weights} shows the total norm of the network parameters of $\varphi_{i}^{NN}$ and $\varphi_{i}^{NN}/|||\varphi_{i}^{NN}|||$ at each training epoch. We observe that in both the normalized and unnormalized case, the network parameters are well-bounded and do not exhibit explosive behavior. Lastly, Table \ref{tab:poisson1d cond} shows the condition number of the matrix $K^{(j)}$ at each Galerkin iteration $i$. We observe that the condition number is approximately unity as expected. 
	\begin{table}
		\centering
		\begin{tabular}{c|cccccccccccc}
			$i$ & 1 & 2 & 3 & 4 & 5 & 6 & 7\\ \hline
			$\text{cond}(K^{(i)})$ & 1.00 & 1.00 & 1.00 & 1.00 & 1.01 & 1.36 & 1.37\\
		\end{tabular}
		\caption{Displacement of a membrane. Condition number for the Galerkin matrix at each Galerkin iteration $i$. We observe that $\text{cond}(A^{(i)}) = \mathcal{O}(1)$.}
		\label{tab:poisson2d cond}
	\end{table}
	
	\begin{figure}[t!]
		\centering
		\includegraphics[width=1.8in]{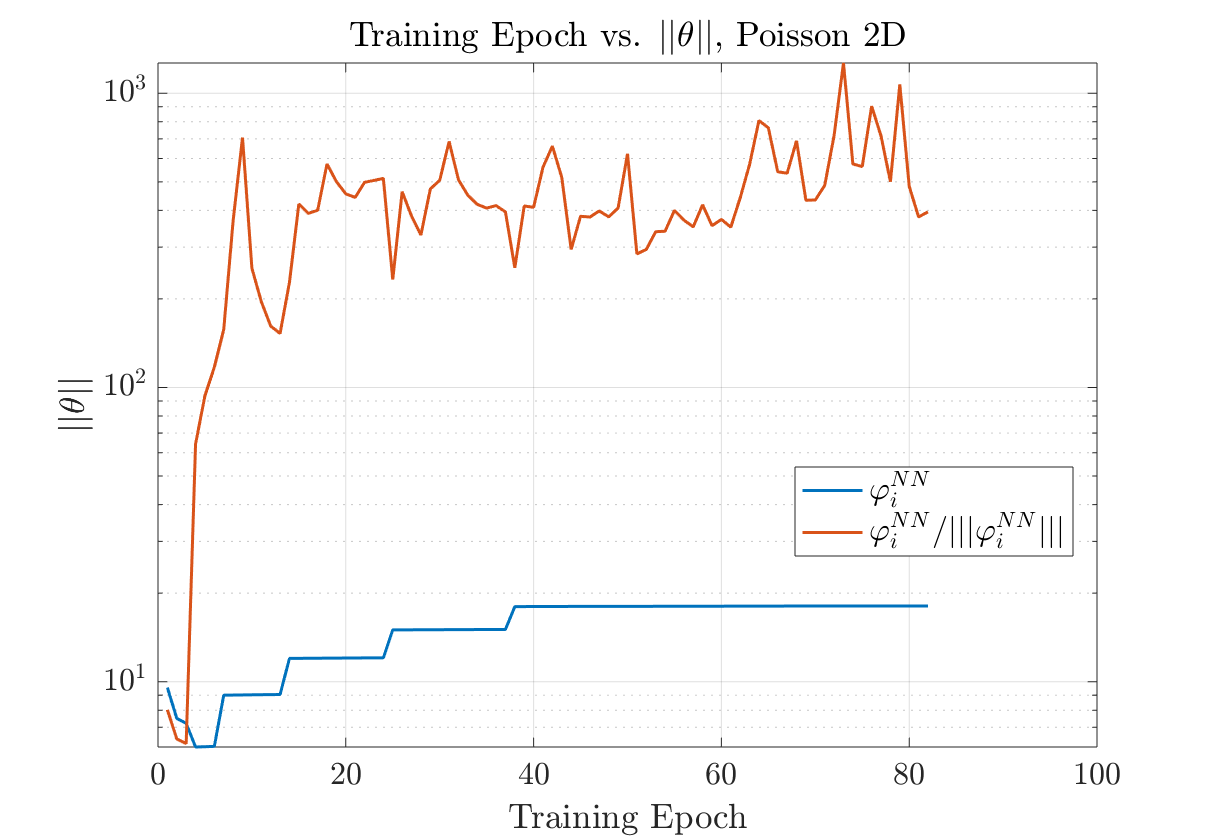}
		\caption{Displacement of a membrane. The total norm of the network parameters for $\varphi_{i}^{NN}$ and $\varphi_{i}^{NN}/|||\varphi_{i}^{NN}|||$.}
		\label{fig:poisson2d weights}
	\end{figure}
	
	\subsubsection{Displacement of a Membrane} \label{sec:membrane}
	We consider the Poisson equation in the unit disk:
	\begin{align*}
		\begin{dcases}
			-\Delta u = f, &\text{in}\;\Omega = \{(x,y) : x^{2}+y^{2}<1\}\\
			u+\varepsilon\partial_{n}u = 0, &\text{on}\;\partial\Omega,
		\end{dcases}
	\end{align*}
	
	\noindent where, $f(x,y) = 2$. The variational formulation is posed on the Hilbert space $X=H^{1}(\Omega)$ corresponding to the bilinear operator $a(u,v) = (\nabla u,\nabla v)_{\Omega} + \varepsilon^{-1}(u,v)_{\partial\Omega}$ with data $L(v) := (f,v)_{\Omega}$. One can easily verify that $a$ satisfies \eqref{eq:condition}. The true solution is 
	\begin{align*}
		u(x,y) = -\frac{x^{2}+y^{2}}{2} + \varepsilon + \frac{1}{2}.
	\end{align*}
	
	\noindent The corresponding loss function is approximated by
	\begin{align} \label{eq:poisson2d loss}
		\frac{L_{1} - L_{2} - L_{3}} {\sqrt{ \sum_{i=1}^{n_{\Omega}} w_{i}|\nabla v(x_{i}^{\Omega};\theta)|^{2} + \varepsilon^{-1}\sum_{i=1}^{n_{\partial\Omega}} w_{i}^{\partial\Omega}v(x_{i}^{\partial\Omega};\theta)^{2}}}
	\end{align} 
	
	\noindent where $L_{1} = \sum_{i=1}^{n_{\Omega}} w_{i}^{\Omega} f(x_{i})v(x_{i};\theta)$, $L_{2} = \sum_{i=1}^{n_{\Omega}} w_{i}^{\Omega} \nabla u_{0}(x_{i}^{\Omega})\cdot \nabla v(x_{i}^{\Omega};\theta)$, and $L_{3} = \varepsilon^{-1}\sum_{i=1}^{n_{\partial\Omega}} w^{\partial\Omega}_{i}u_{0}(x_{i}^{\partial\Omega})v(x_{i}^{\partial\Omega};\theta)$. Here, $\{w^{\Omega},x^{\Omega}\}$ is the quadrature rule in $\Omega$ while $\{w^{\partial\Omega}, x^{\partial\Omega}\}$ is the quadrature rule on $\partial\Omega$.
	
	The tolerance is set to $\texttt{tol} = 2 \times 10^{-6}$, and $\varepsilon = 10^{-4}$. The network architecture for Galerkin iteration $i$ is as follows: $n_{i} = 200 + \left\lfloor \frac{i-1}{2} \right\rfloor \cdot 100$, $\sigma_{i}(t) = \tanh(t)$. The learning rate is $\alpha = (1 \times 10^{-2})/1.1^{i-1}$. The hidden parameters are initialized such that the hyperplanes lie parallel to $y=0$, $x=0$, $y=x$, or $y=-x$ as visualized in Figure \ref{fig:2d errors} (right). To evaluate the loss function and train the network, we employ a fixed Gauss-Legendre quadrature rule with $128 \times 128$ nodes to approximate all inner products in the interior of the domain. A quadrature rule with $256$ equally-spaced nodes is employed to approximate inner products on the boundary of the domain.

	Figure \ref{fig:2d errors} (left) shows the exact errors $||u-u_{i-1}||_{L^{2}}$ and $|||u-u_{i-1}|||$ at the end of each epoch along with the estimated errors $||\varphi_{i}^{NN}||_{L^{2}}$ and $\eta(u_{i-1},\varphi_{i}^{NN})$ for our algorithm. Figure \ref{fig:poisson2d plots} shows the exact error $u-u_{i-1}$ (right column) as well as the computed maximizer $\varphi_{i}^{NN}$ (left column) at several stages of our algorithm. We again observe good performance of the error estimator, with minor discrepancies beginning to show in the final Galerkin iteration. Figure \ref{fig:poisson2d weights} shows the total norm of the network parameters of $\varphi_{i}^{NN}$ and $\varphi_{i}^{NN}/|||\varphi_{i}^{NN}|||$ at each training epoch, and we again observe well-boundedness of the parameters. Table \ref{tab:poisson2d cond} shows the condition number of the Galerkin matrix $K^{(i)}$ at Galerkin iteration $i$. We observe again that $\text{cond}(K^{(i)})$ is approximately unity. 
	\begin{figure}[t!]
		\centering
		\begin{subfigure}{.3\textwidth}
			\centering
			\includegraphics[width=1.4in]{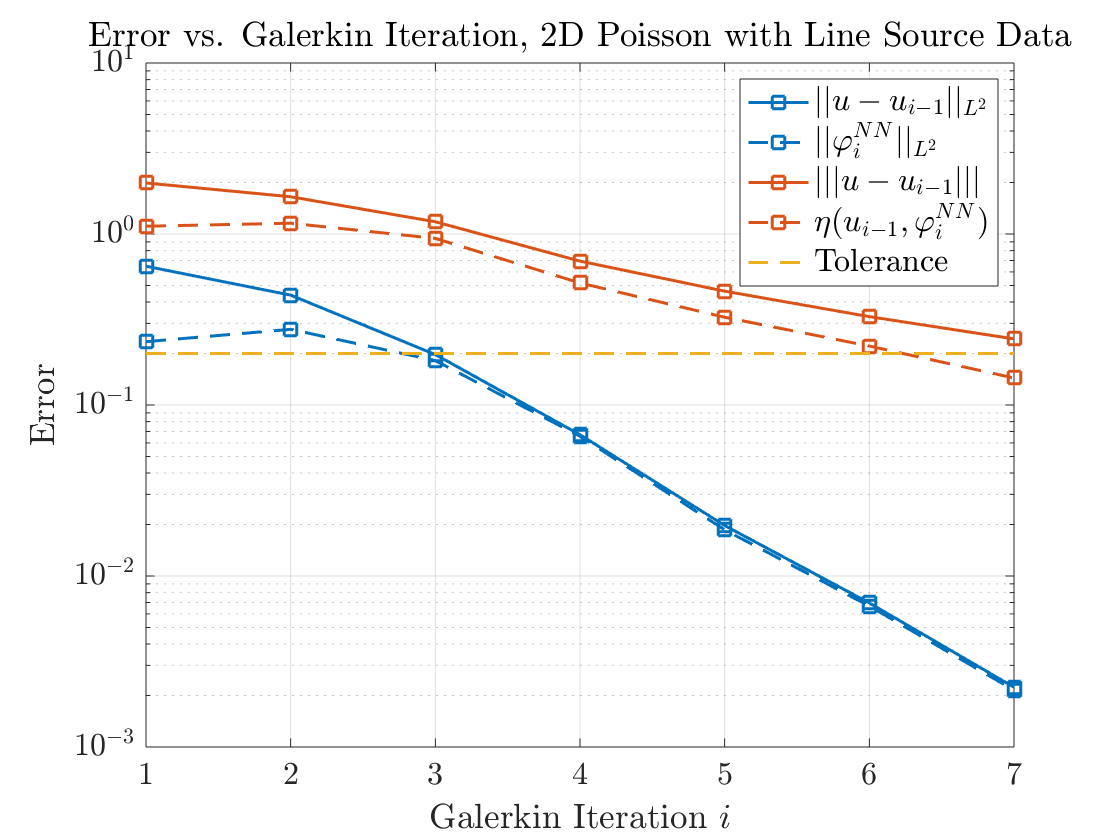}
			\caption{}
		\end{subfigure}
		\begin{subfigure}{.3\textwidth}
			\centering
			\includegraphics[width=1.6in]{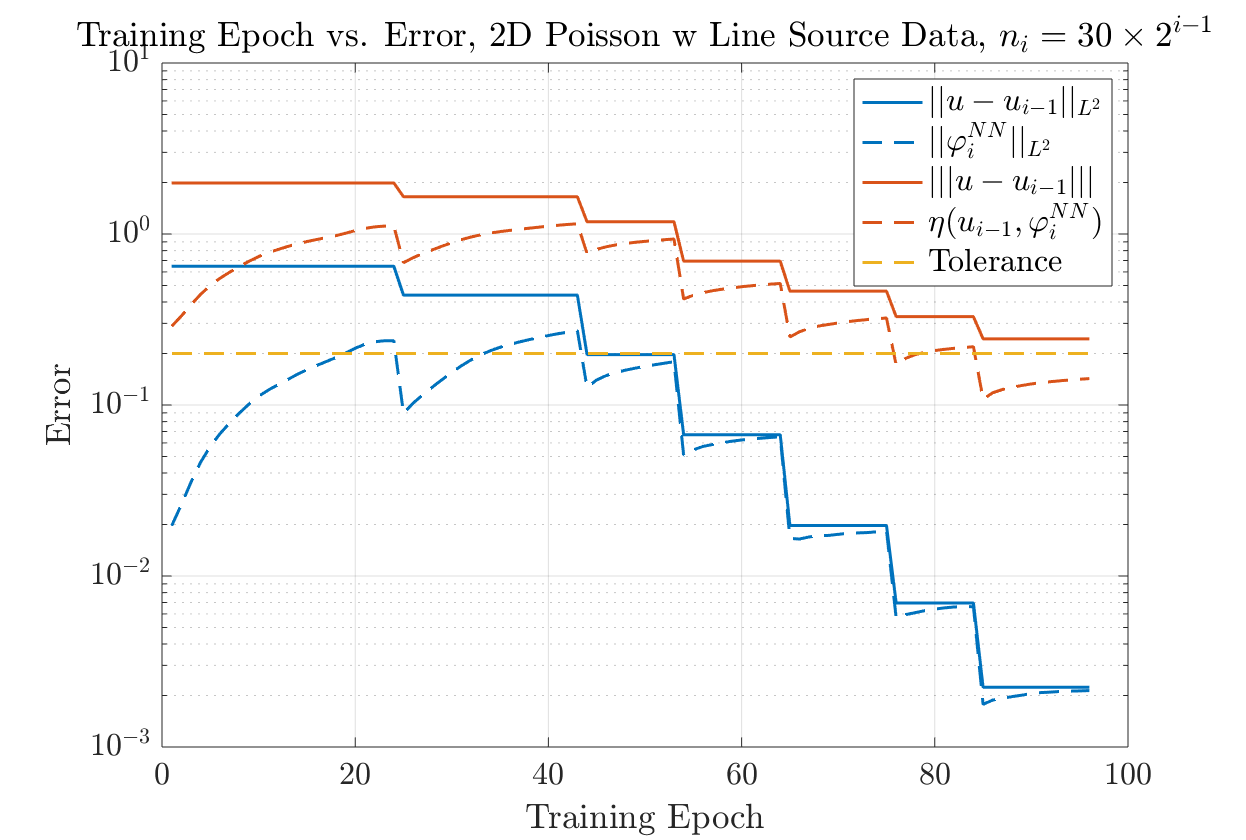}
			\caption{}
		\end{subfigure}
		\begin{subfigure}{.3\textwidth}
			\centering
			\includegraphics[width=1.5in]{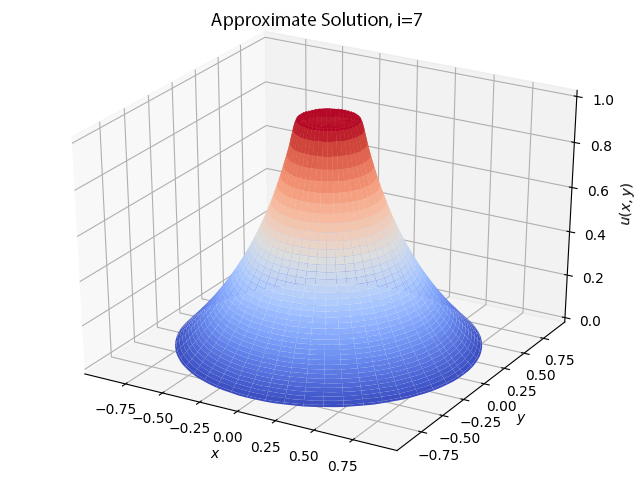}
			\caption{}
		\end{subfigure}
		\caption{Poisson equation with line source data, $R_{e} = 1-1/\pi^{2}$ and $R_{0} = 1/\sqrt{29}$. (a) true and estimated errors after each Galerkin iteration (b) true and estimated errors after each training epoch (c) approximate solution $u_{i}$ for $i=7$.}
		\label{fig:delta2d}
	\end{figure}

	\subsubsection{Poisson Equation with Line Source Data}
	
	We once more consider the Poisson equation in two dimensions, this time in the domain $\Omega = \{(x,y) : x^{2}+y^{2}<R_{e}^{2}\}$ with data $L : H^{1}(\Omega) \to \mathbb{R}$ defined by
	\begin{align*}
		L(v) := \int_{\Gamma} v\;ds, \;\;\;\Gamma = \{(x,y) : x^{2} + y^{2} = R_{0}^{2} \}, \;\;\;R_{0} < R_{e}.
	\end{align*}
	

	\begin{figure}[t!]
		\centering
		\begin{subfigure}{.3\textwidth}
			\centering
			\includegraphics[width=1.4in]{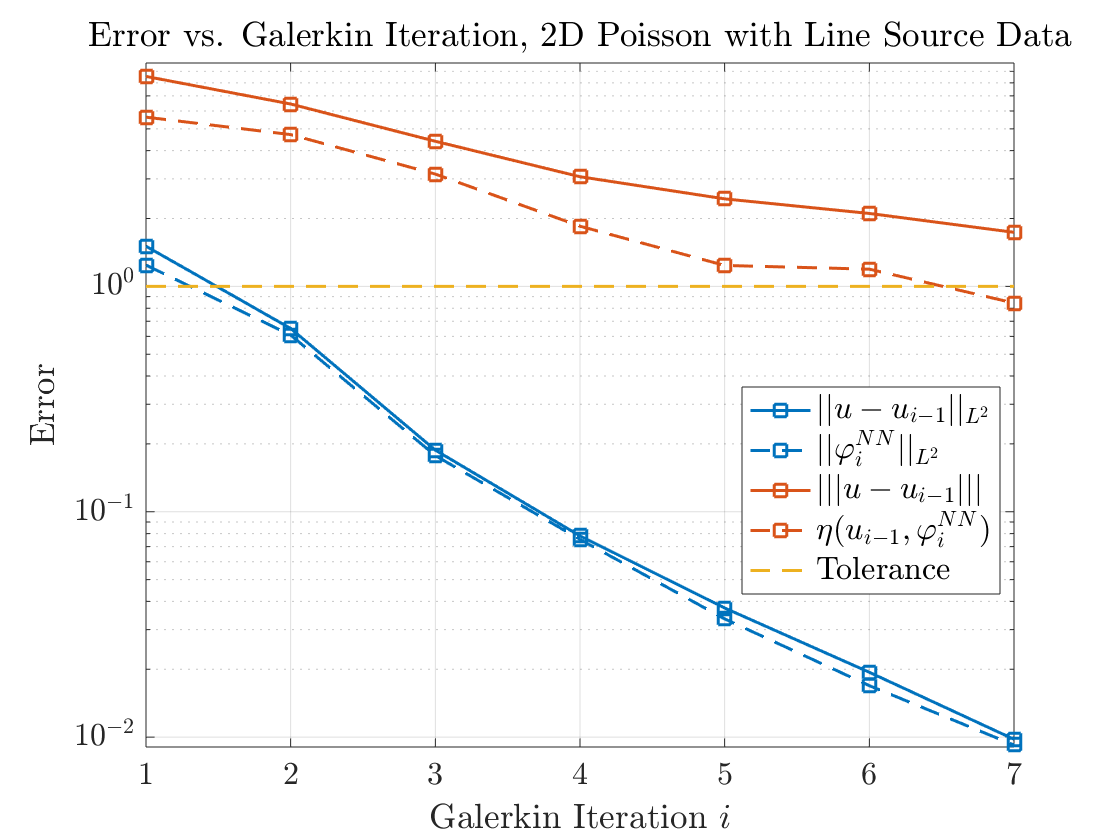}
			\caption{}
		\end{subfigure}
		\begin{subfigure}{.3\textwidth}
			\centering
			\includegraphics[width=1.6in]{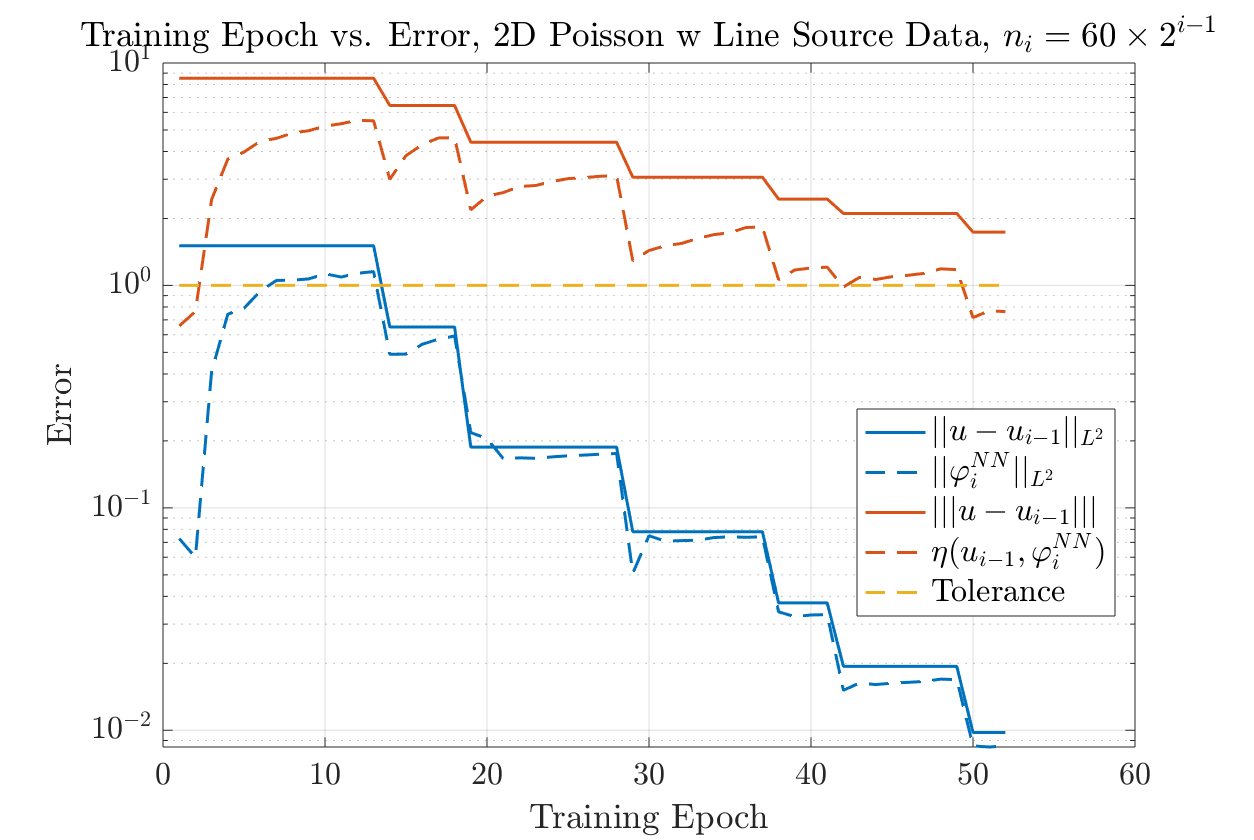}
			\caption{}
		\end{subfigure}
		\begin{subfigure}{.3\textwidth}
			\centering
			\includegraphics[width=1.5in]{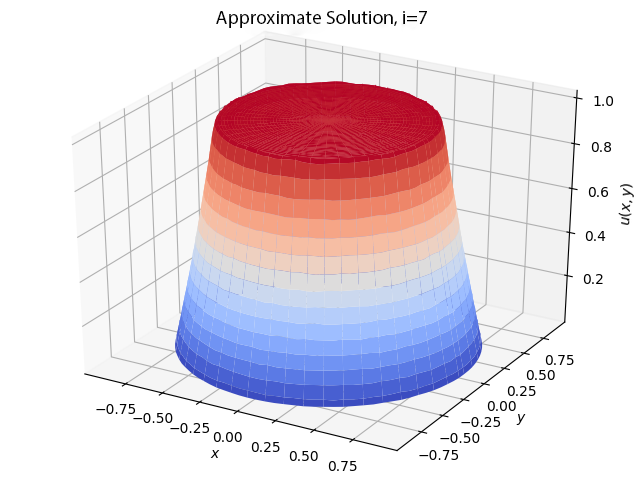}
			\caption{}
		\end{subfigure}
		\caption{Poisson equation with line source data, $R_{e} = 1-1/\pi^{2}$ and $R_{0} = 7/(6\sqrt{2})$. (a) true and estimated errors after each Galerkin iteration (b) true and estimated errors after each training epoch (c) approximate solution $u_{i}$ for $i=7$.}
		\label{fig:delta2dBL}
	\end{figure}

	\noindent The exact variational solution is constant in the inner disk $\{(x,y) : x^{2}+y^{2}\leqslant R_{0}^{2}\}$ and logarithmic in the annulus $\Omega \backslash \{(x,y) : x^{2}+y^{2}\leqslant R_{0}^{2}\}$:
	\begin{align*}
		u(r,\theta) = -\frac{\varepsilon}{R_{0}\log(R_{0}/R_{e})} + \begin{dcases}
			1, &r \leqslant R_{0}\\
			\frac{1}{\log(R_{0}/R_{e})} \log(r/R_{e}), &R_{0} < r \leqslant R_{e}.
		\end{dcases}
	\end{align*}
	
	\noindent We consider two separate cases:
	\begin{enumerate}
		\item $R_{e} = 1-1/\pi^{2}$ and $R_{0} = 1/\sqrt{29}$,
		
		\item $R_{e} = 1-1/\pi^{2}$ and $R_{0} = 7/(6\sqrt{2})$. 
	\end{enumerate}
	
	\noindent In the first case, the logarithmic behavior dominates with a sharp cusp in the solution at $r=R_{0}$. In the second case, the solution still has a sharp cusp at $r=R_{0}$, but the logarithmic behavior is limited to the small annulus $R_{0} \leqslant r \leqslant R_{e}$, which resembles a boundary layer. The corresponding loss function is approximated by
	\begin{align*}
		\frac{L_{1} - L_{2} - L_{3}} {\sqrt{ \sum_{i=1}^{n_{\Omega}} w_{i}|\nabla v(x_{i}^{\Omega};\theta)|^{2} + \varepsilon^{-1}\sum_{i=1}^{n_{\partial\Omega}} w_{i}^{\partial\Omega}v(x_{i}^{\partial\Omega};\theta)^{2}}}
	\end{align*} 
	
	\noindent where $L_{1} = \sum_{i=1}^{n_{\Gamma}} w_{i}^{\Gamma} v(x_{i}^{\Gamma};\theta)$, $L_{2} = \sum_{i=1}^{n_{\Omega}} w_{i}^{\Omega} \nabla u_{0}(x_{i}^{\Omega})\cdot \nabla v(x_{i}^{\Omega};\theta)$, and $L_{3} = \varepsilon^{-1} \cdot \sum_{i=1}^{n_{\partial\Omega}} w^{\partial\Omega}_{i} u_{0}(x_{i}^{\partial\Omega}) v(x_{i}^{\partial\Omega};\theta)$. Here, $\{w^{\Omega},x^{\Omega}\}$ is the quadrature rule in $\Omega$, $\{w^{\partial\Omega}, x^{\partial\Omega}\}$ is the quadrature rule on $\partial\Omega$, and $\{w^{\Gamma},x^{\Gamma}\}$ is the quadrature rule on $\Gamma$.
		
	The tolerance is set to $\texttt{tol} = 2 \times 10^{-1}$ in the first case and $\texttt{tol} =  1.0$ in the second case, and $\varepsilon = 10^{-3}$. The network architecture for iteration $i$ is as follows: $n_{i} =  30 \times 2^{i-1}$, $\sigma_{i}(t) = \tanh(\beta_{i}t)$, $\beta_{i} = i$. The learning rate for each basis function $\varphi_{i}^{NN}$ is $\alpha_{i} = \frac{2 \times 10^{-2}}{1.1^{i-1}}$. The hidden parameters are initialized according to the box initialization detailed by Cyr, et. al. \cite{cyr}. We employ fixed tensor-product Gauss-Legendre quadrature rule with $128 \times 128$ nodes in order to approximate inner products in the interior of the domain. We approximate the inner products on $\partial\Omega$ and $\Gamma$ using $512$ equally-spaced nodes on the circles of radii $R_{e}$ and $R_{0}$, respectively.
	
	The true and estimated errors after each refinement and epoch are provided in Figure \ref{fig:delta2d} along with a plot of the neural network approximation for case 1. The analogous results for case 2 are provided in Figure \ref{fig:delta2dBL}. Interestingly, the cusp along $r=R_{0}$ is resolved well in both cases without any encoding of the line source into the training data, albeit with a lower rate of convergence as compared with the examples with smooth solutions.
	\begin{figure}[t!]
		\centering
		\begin{subfigure}{.3\textwidth}
			\centering
			\includegraphics[width=1.5in]{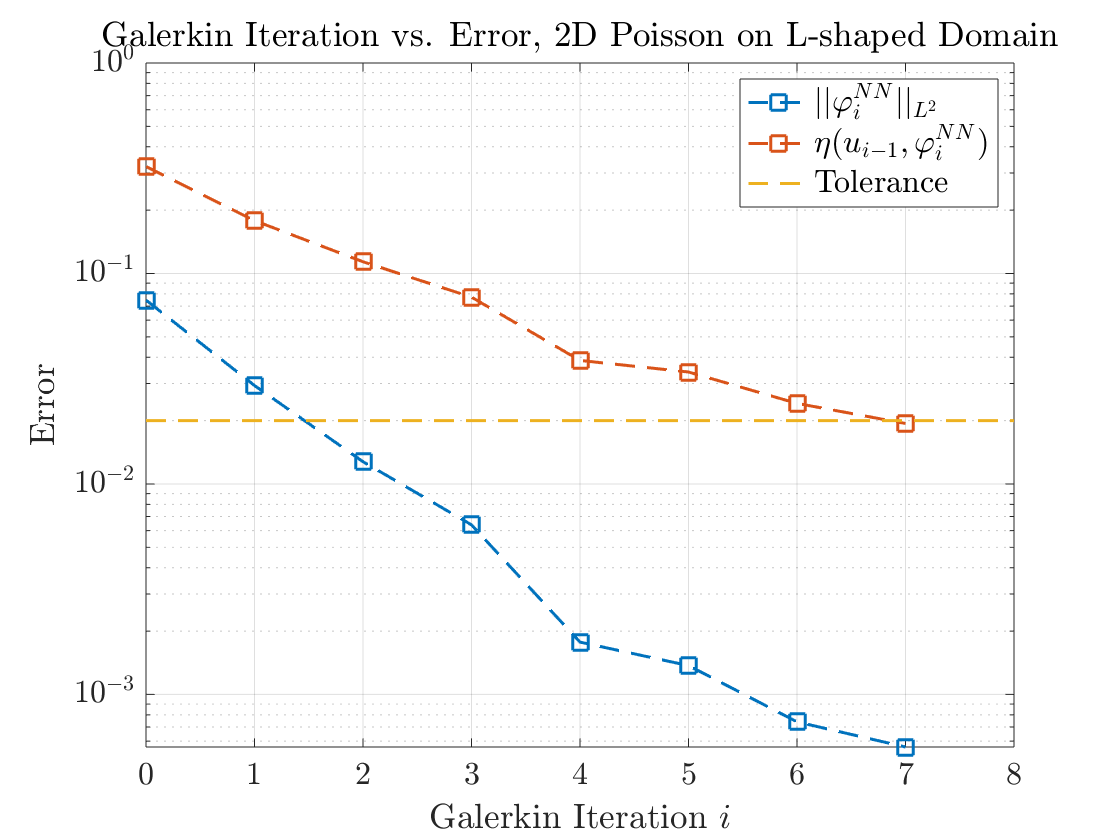}
			\caption{}
			\label{fig:Lshaped error galerkin}
		\end{subfigure}
		\quad
		\begin{subfigure}{.3\textwidth}
			\centering
			\includegraphics[width=1.5in]{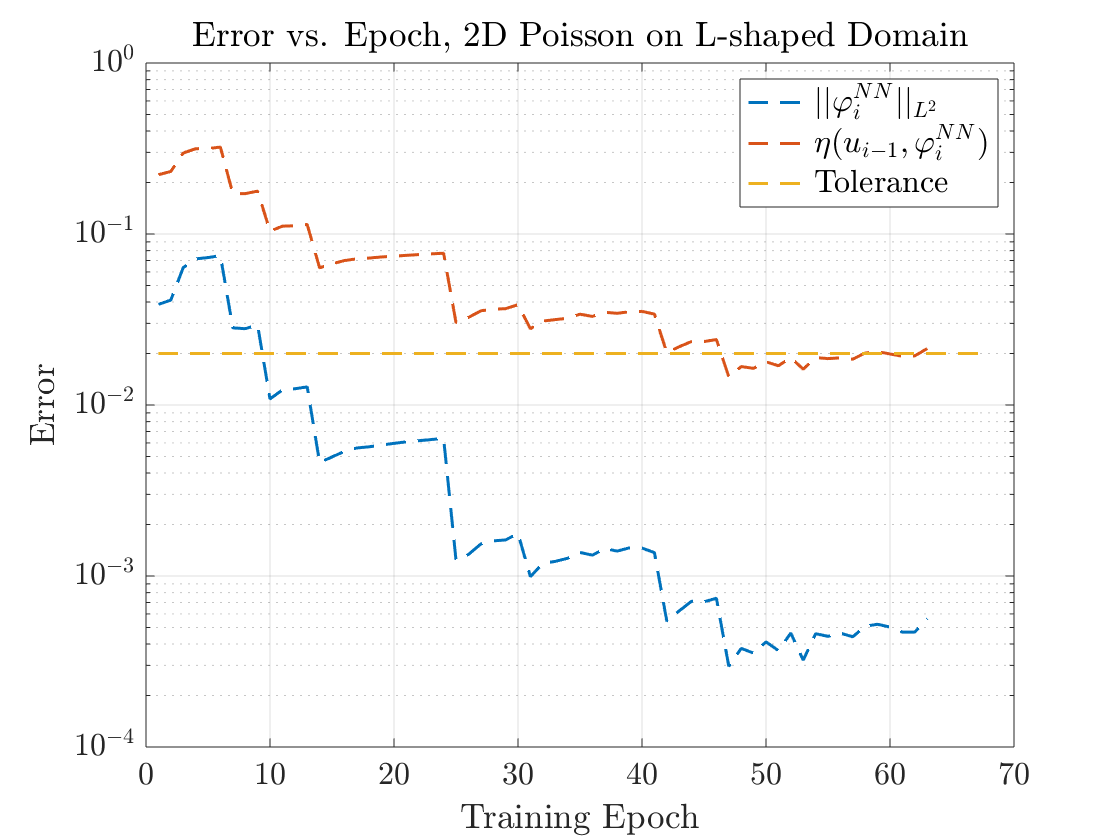}
			\caption{}
			\label{fig:Lshaped error epoch}
		\end{subfigure}
		\quad
		\begin{subfigure}{.3\textwidth}
			\centering
			\includegraphics[width=1.5in]{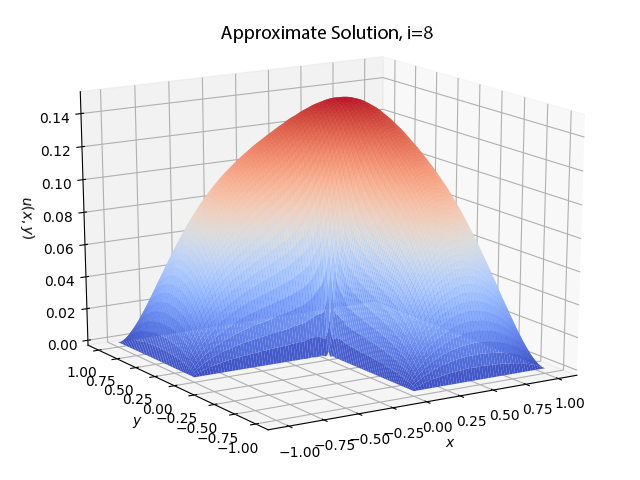}
			\caption{}
			\label{fig:Lshaped plot}
		\end{subfigure}
		\caption{Poisson equation in L-shaped domain. Estimated errors after each Galerkin iteration (a) and epoch (b) as well as approximate solution $u_{i}$ for $i=8$ (c).}
		\label{fig:Lshaped error}
	\end{figure}
	
	\subsubsection{L-shaped Domain}
	In this example, we consider the Poisson equation on an L-shaped domain in which the solution contains a corner singularity:
	\begin{align*}
		\begin{dcases}
			-\Delta u = 1, &\text{in}\;\Omega = (-1,1)^{2} \backslash (-1,0]^{2}\\
			u+\varepsilon\partial_{n}u = 0, &\text{on}\;\partial\Omega.
		\end{dcases}
	\end{align*}
	
	\noindent The loss function is given by \eqref{eq:poisson2d loss} with $f \equiv 1$ and is approximated using the same methodology as in \ref{sec:membrane}.
	
	The tolerance is set to $\texttt{tol} = 2\times 10^{-2}$, and $\varepsilon = 10^{-4}$. The network architecture for iteration $i$ is as follows: $n_{i} = 20 \times 2^{i-1}$, $\sigma_{i}(t) = \tanh(t)$. The learning rate for each basis function $\varphi_{i}^{NN}$ is $\alpha_{i} = \frac{2 \times 10^{-2}}{1.1^{i-1}}$. The hidden parameters are again initialized according to the box initialization in $(-1,1)^{2}$. We employ fixed Gauss-Legendre quadrature rules on each square subdomain of $128 \times 128$ nodes each in order to approximate inner products in the interior of the domain. A Gauss-Lobatto quadrature rule is employed on $[-1,0] \times \{0\}$ and $\{0\} \times [-1,0]$ to approximate the boundary containing the singularity, while Gauss-Legendre quadrature rules are employed on the remaining boundary edges.

	Figure \ref{fig:Lshaped error galerkin} shows the estimated errors $||\varphi_{i}^{NN}||_{L^{2}}$ and $\eta(u_{i-1},\varphi_{i}^{NN})$ after each Galerkin iteration. Figure \ref{fig:Lshaped error epoch} shows the the estimated errors at each training iteration, while Figure \ref{fig:Lshaped plot} shows the approximate solution to the variational problem after eight Galerkin iterations. As expected, without a priori knowledge of the location of the corner singularity, we observe a reduced rate of convergence as compared to prior examples with smooth solutions. Nevertheless, our algorithm is capable of resolving the corner singularity.

	\subsection{Fourth Order Problems}
	\subsubsection{Displacement of a Beam} \label{sec:biharmonic smooth}
	We consider the fourth-order ODE
	\begin{align*}
		\begin{dcases}
			u^{(4)}(x) = f(x), &\text{in}\;\Omega=(0,1)\\
			u-\varepsilon_{1} \partial_{n}(u'') = 0, &\text{on}\;\partial\Omega\\
			u'+\varepsilon_{2} \partial_{n}(u') = 0, &\text{on}\;\partial\Omega
		\end{dcases}
	\end{align*}
	
	\noindent with $f(x) = (2\pi)^{4}\sin(2\pi x)$. The variational formulation is posed on the Hilbert space $X=H^{2}(\Omega)$ with corresponding bilinear operator $a(u,v) = (u'',v'')_{\Omega} + \varepsilon_{1}^{-1}(u(0)v(0) + u(1)v(1)) + \varepsilon_{2}^{-1}(u'(0)v'(0) + u'(1)v'(1))$ and data $L(v) = (f,v)_{\Omega}$. The exact solution is given by
	\begin{align*}
		u(x) = \sin(2\pi x) - 2\pi (2x-1)\frac{4\varepsilon_{1}(\pi^{2}(6\varepsilon_{2} - 2x^{2} + 2x + 1) + 3) + x(x-1)}{24\varepsilon_{1} + 6\varepsilon_{2} + 1}.
	\end{align*}
	
	\noindent The corresponding loss function is approximated by
	\begin{align*}
		\frac{L_{1} - L_{2} - L_{3} - L_{4}} {\sqrt{ \sum_{i=1}^{n_{G}} w_{i}v''(x_{i};\theta)^{2} + \varepsilon_{1}^{-1}(v(0;\theta)^{2} + v(1;\theta)^{2}) + \varepsilon_{2}^{-1}(v'(0;\theta)^{2} + v'(1;\theta)^{2}}}
	\end{align*} 
	
	\noindent where $L_{1} = \sum_{i=1}^{n_{G}} w_{i} f(x_{i})v(x_{i};\theta)$, $L_{2} = \sum_{i=1}^{n_{G}} w_{i} u_{0}''(x_{i})v''(x_{i};\theta)$, $L_{3} = \varepsilon_{1}^{-1}[u_{0}(0)\cdot v(0;\theta) + u_{0}(1)v(1;\theta)]$, and $L_{4} = \varepsilon_{2}^{-1}[u_{0}'(0)v'(0;\theta) + u_{0}'(1)v'(1;\theta)]$.
	
	\begin{figure}[t!]
		\centering
		\begin{subfigure}{.47\textwidth}
			\centering
			\includegraphics[width=2.0in]{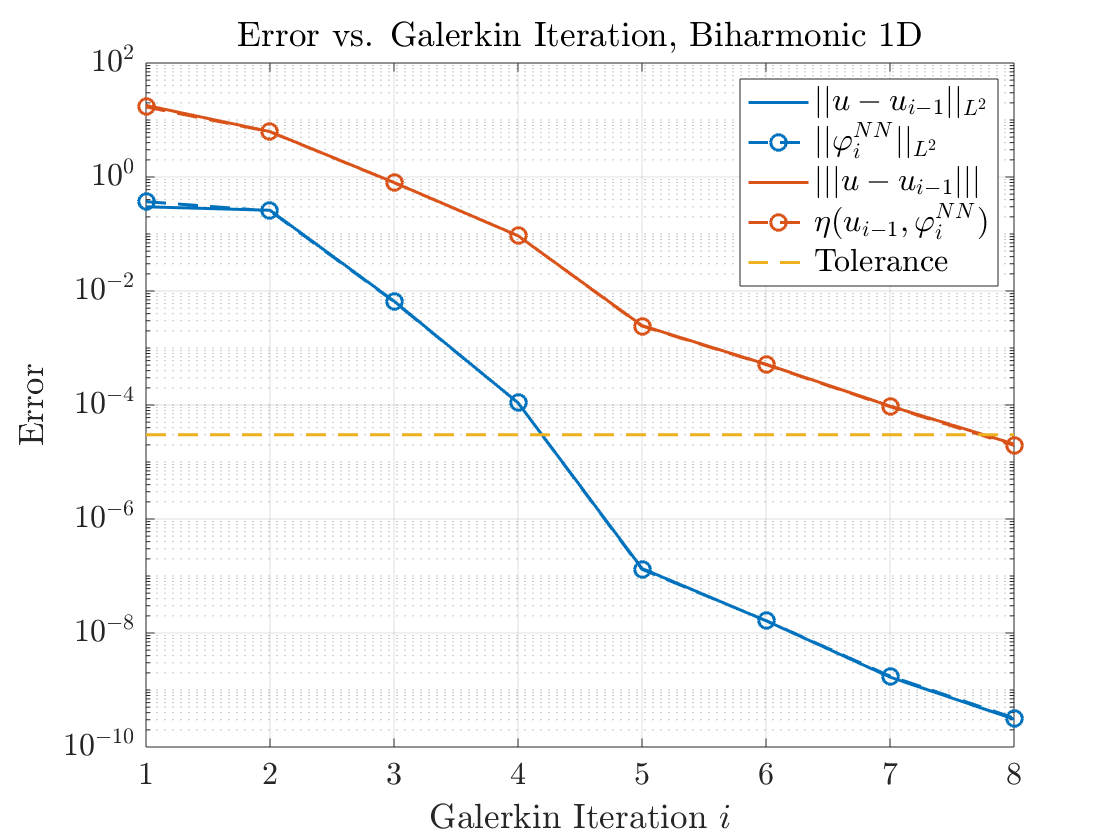}
			\caption{}
			\label{fig:biharmonic1d galerkin}
		\end{subfigure}
		\quad
		\begin{subfigure}{.47\textwidth}
			\centering
			\includegraphics[width=2.0in]{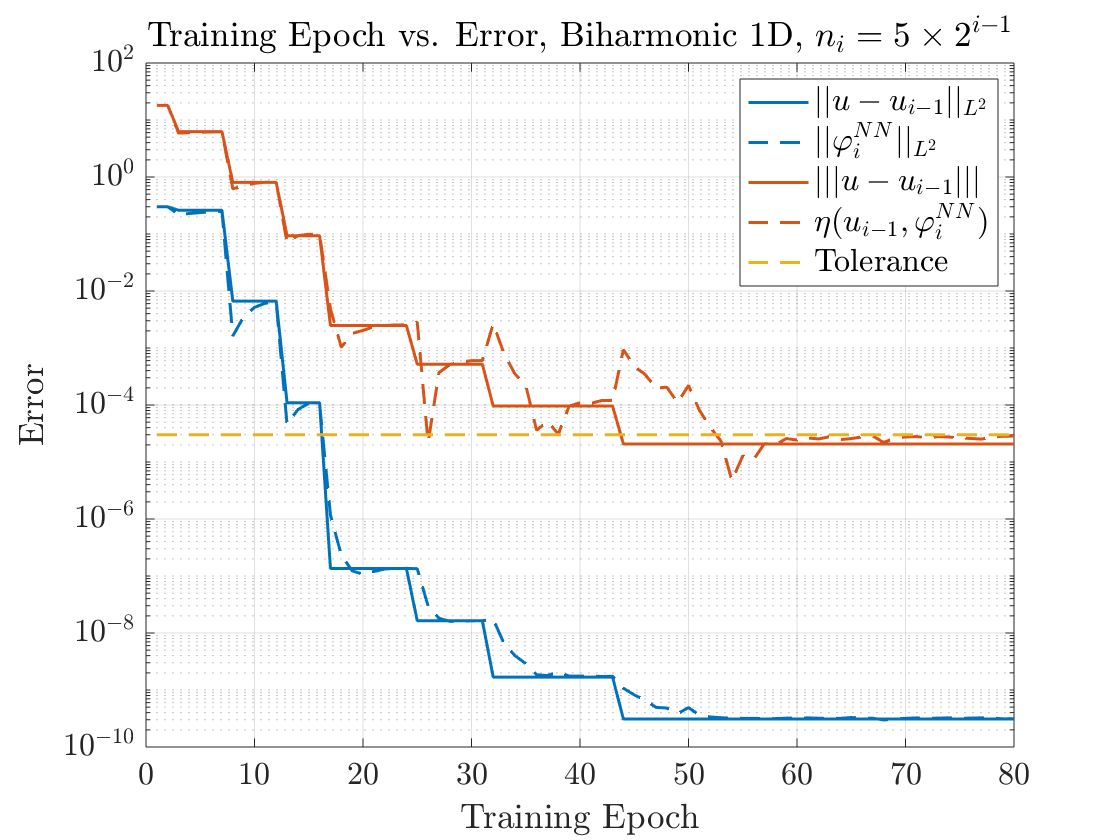}
			\caption{}
			\label{fig:biharmonic1d epoch}
		\end{subfigure}
		\caption{The biharmonic equation in 1D. True and estimated error in the $L^{2}$ and energy norm at the end of each Galerkin iteration (a) and epoch (b).}
		\label{fig:biharmonic1d errors}
	\end{figure}
	
	The tolerance is set to $\texttt{tol} = 3 \times 10^{-5}$, and $\varepsilon_{1} = \varepsilon_{2} = 10^{-4}$. The network architecture for iteration $i$ is as follows: $n_{i} = 30\times 2^{i-1}$, $\sigma_{i}(t) = \tanh(\beta_{i}t)$, $\beta_{j} = 1+3(i-1)$. The learning rate for each basis function $\varphi_{i}^{NN}$ is $\alpha_{i} = \frac{2 \times 10^{-2}}{1.1^{i-1}}$. The hidden parameters are initialized as in Section \ref{sec:function fitting}. To evaluate the loss function and train the network, we employ a fixed Gauss-Legendre quadrature rule with 512 nodes to approximate all inner products. A separate Gauss-Legendre quadrature rule with 1000 nodes is used for validation, i.e. to compute the exact errors in the $L^{2}$ and energy norms. 
	\begin{figure}[t!]
		\centering
		\begin{subfigure}{.47\textwidth}
			\centering
			\includegraphics[width=1.8in]{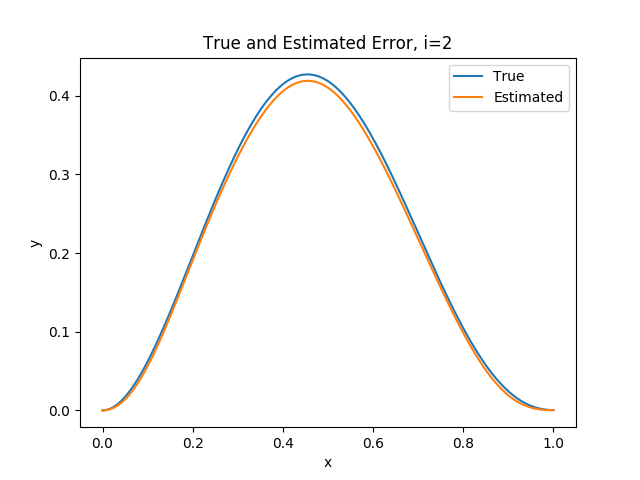}
			\caption{}
		\end{subfigure}
		\quad
		\begin{subfigure}{.47\textwidth}
			\centering
			\includegraphics[width=1.8in]{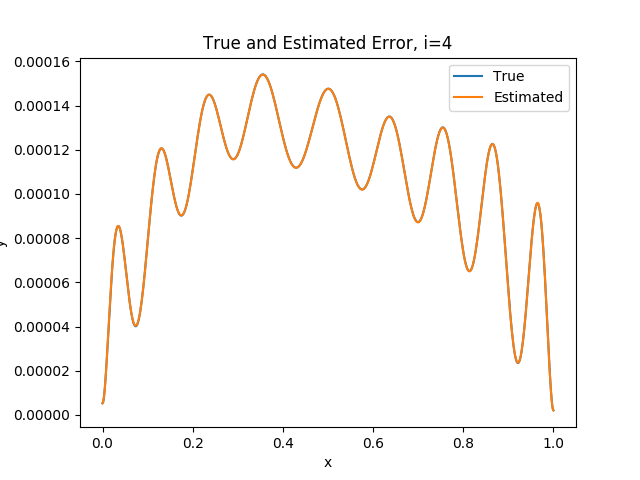}
			\caption{}
		\end{subfigure}
		\quad
		\begin{subfigure}{.47\textwidth}
			\centering
			\includegraphics[width=1.8in]{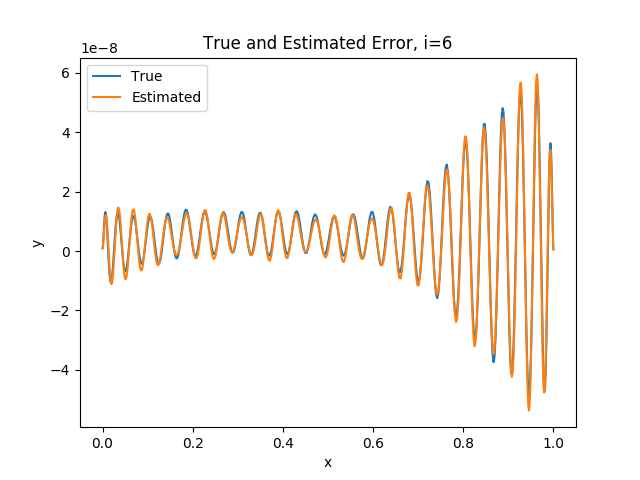}
			\caption{}
		\end{subfigure}
		\quad
		\begin{subfigure}{.47\textwidth}
			\centering
			\includegraphics[width=1.8in]{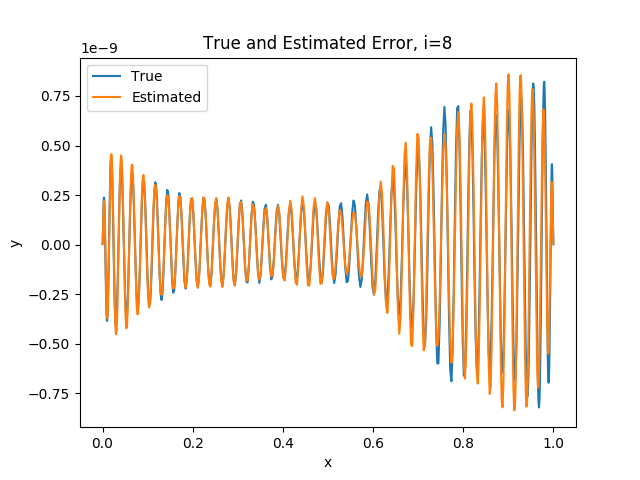}
			\caption{}
		\end{subfigure}
		\caption{Biharmonic equation. Exact error $u-u_{i-1}$ and approximate error $\varphi_{i}^{NN}$ for $i=2,4,6,8$.}
		\label{fig:biharmonic1d plots}
	\end{figure}
	
	Figure \ref{fig:biharmonic1d galerkin} shows the exact errors $||u-u_{i-1}||_{L^{2}}$ and $|||u-u_{i-1}|||$ at the end of each Galerkin iteration along with the estimated errors $||\varphi_{i}^{NN}||_{L^{2}}$ and $\eta(u_{i-1},\varphi_{i}^{NN})$ for our algorithm. Figure \ref{fig:biharmonic1d epoch} shows the true and estimated errors at each training epoch, while Figure \ref{fig:biharmonic1d plots} shows the exact error $u-u_{i-1}$ as well as the maximizer $\varphi_{i}^{NN}$ at several stages of refinement. We observe no difficulties in approximating both the true error and the true solution.

	\subsubsection{Beam with Applied Couple}
	We consider the biharmonic equation with point moment source term:
	\begin{align*}
		\begin{dcases}
			u^{(4)} = \delta'_{\{x=1/2\}}, &\text{in}\;\Omega=(0,1)\\
			u-\varepsilon_{1} \partial_{n}(u'') = 0, &\text{on}\;\partial\Omega\\
			u'+\varepsilon_{2} \partial_{n}(u') = 0, &\text{on}\;\partial\Omega.
		\end{dcases}
	\end{align*}
	
	\noindent The bilinear operator $a$ is as defined in Section \ref{sec:biharmonic smooth} and $L(v) = -v'(1/2)$. The exact variational solution corresponding to $\varepsilon_{1}=\varepsilon_{2}=0$ is 
	\begin{align*}
		u(x) = \frac{1}{4}(x-1/2)|x-1/2| - \frac{1}{4}x^{3} + \frac{3}{8}x^{2} - \frac{1}{4}x + \frac{1}{16}.
	\end{align*}
	
	\noindent The corresponding loss function is approximated by
	\begin{align*}
		\frac{L_{1} - L_{2} - L_{3} - L_{4}} {\sqrt{ \sum_{i=1}^{n_{G}} w_{i}v''(x_{i};\theta)^{2} + \varepsilon_{1}^{-1}(v(0;\theta)^{2} + v(1;\theta)^{2}) + \varepsilon_{2}^{-1}(v'(0;\theta)^{2} + v'(1;\theta)^{2}}}
	\end{align*} 
	
	\noindent where $L_{1} = -v'(1/2;\theta)$, $L_{2} = \sum_{i=1}^{n_{G}} w_{i} u_{0}''(x_{i})v''(x_{i};\theta)$, $L_{3} = \varepsilon_{1}^{-1}[u_{0}(0)v(0;\theta) + u_{0}(1)v(1;\theta)]$, and $L_{4} = \varepsilon_{2}^{-1}[u_{0}'(0)v'(0;\theta) + u_{0}'(1)v'(1;\theta)]$.
	
	\begin{figure}[t!]
		\centering
		\begin{subfigure}{.47\textwidth}
			\centering
			\includegraphics[width=1.8in]{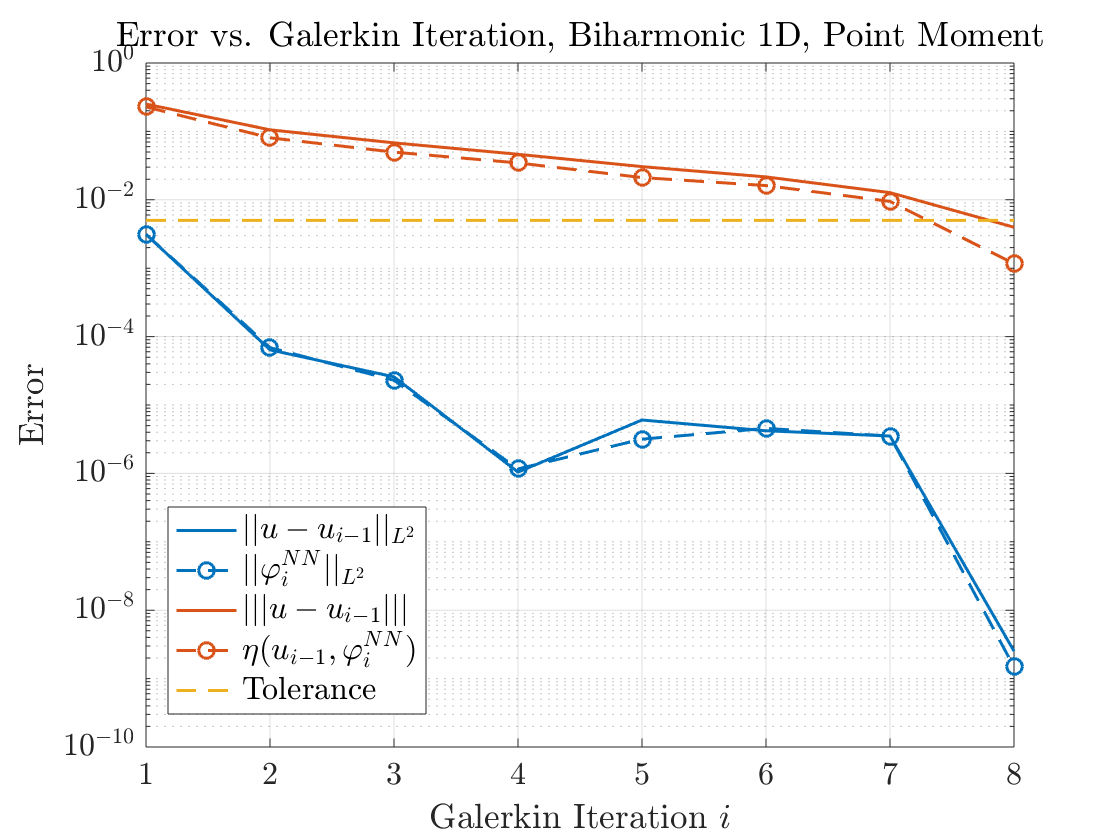}
			\caption{}
			\label{fig:biharmonic1d delta error galerkin}
		\end{subfigure}
		\begin{subfigure}{.47\textwidth}
			\centering
			\includegraphics[width=2.1in]{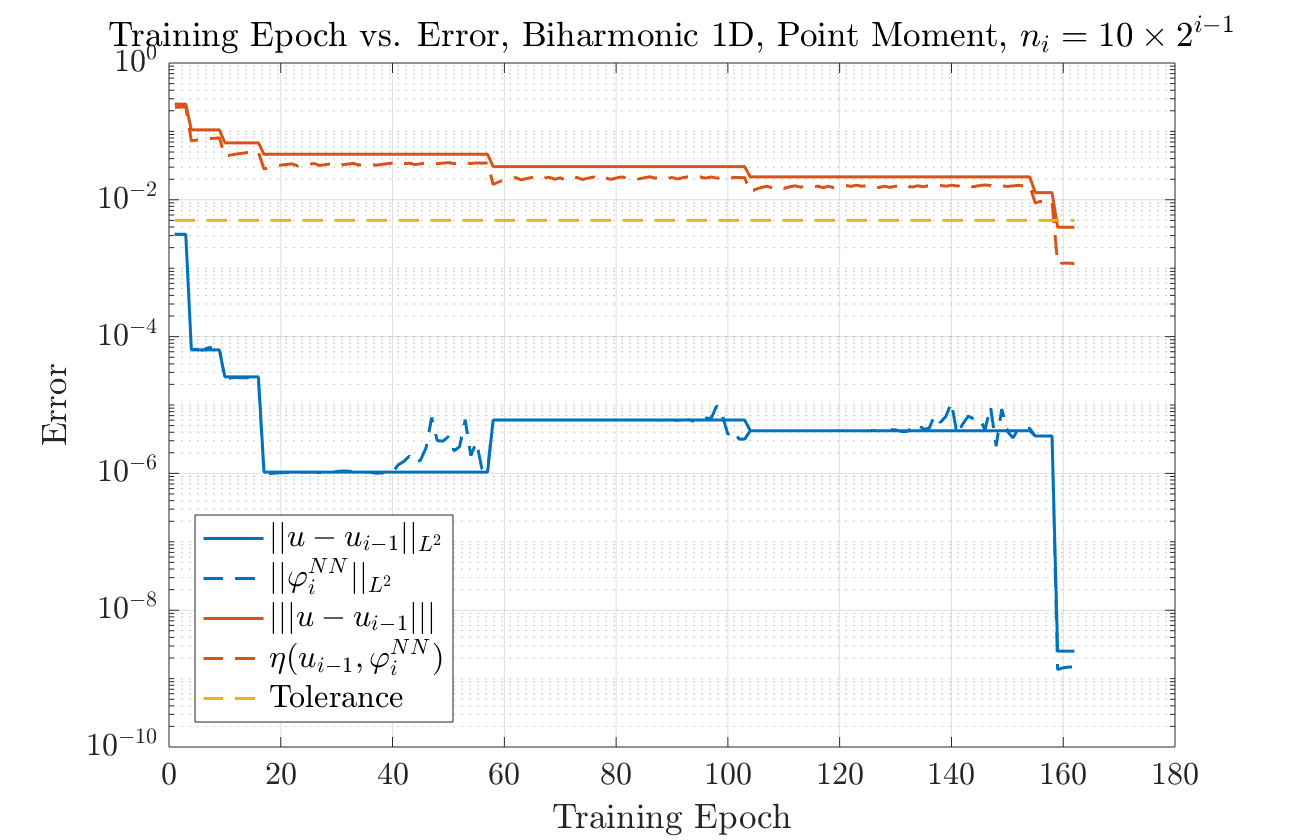}
			\caption{}
			\label{fig:biharmonic1d delta error epoch}
		\end{subfigure}
		\caption{The biharmonic equation in 1D with point moment. True and estimated error in the $L^{2}$ and energy norm at the end of each Galerkin iteration (a) and epoch (b).}
		\label{fig:biharmonic1d delta errors}
	\end{figure}
	
	The tolerance is set to $\texttt{tol} = 4\times 10^{-3}$, and $\varepsilon_{1} = \varepsilon_{2} = 10^{-5}$. The network architecture for iteration $i$ is as follows: $n_{i} = 10\times 2^{i-1}$, $\sigma_{i}(t) = \tanh(\beta_{i}t)$, $\beta_{i} = 1 + 3\cdot 2^{i-1}$. The learning rate for each basis function $\varphi_{i}^{NN}$ is $\alpha_{i} = \frac{1 \times 10^{-2}}{1.4^{i-1}}$. The hidden parameters are initialized as in Section \ref{sec:function fitting}. To evaluate the loss function and train the network, we employ a fixed Gauss-Legendre quadrature rule with 1024 nodes to approximate all inner products. A separate Gauss-Legendre quadrature rule with 1000 nodes is used for validation, i.e. to compute the exact errors in the $L^{2}$ and energy norms. 
	
\begin{figure}[t!]
		\centering
		\begin{subfigure}{.3\textwidth}
			\centering
			\includegraphics[width=1.5in]{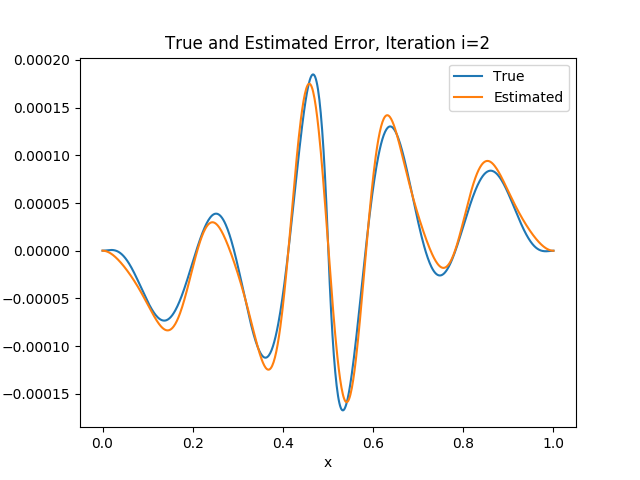}
			\caption{}
		\end{subfigure}
		\quad
		\begin{subfigure}{.3\textwidth}
			\centering
			\includegraphics[width=1.5in]{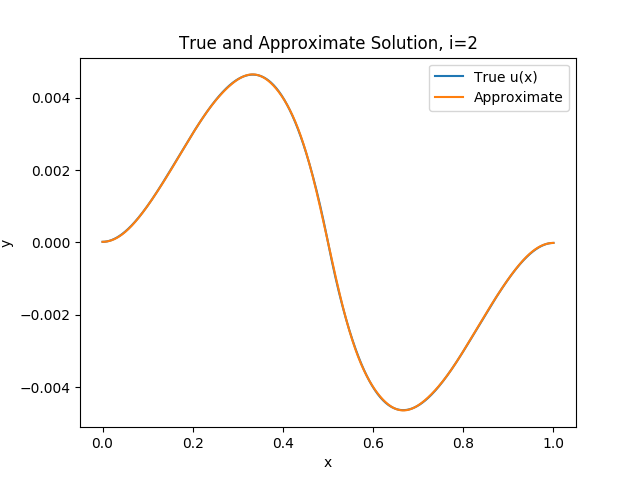}
			\caption{}
		\end{subfigure}
		\quad
		\begin{subfigure}{.3\textwidth}
			\centering
			\includegraphics[width=1.5in]{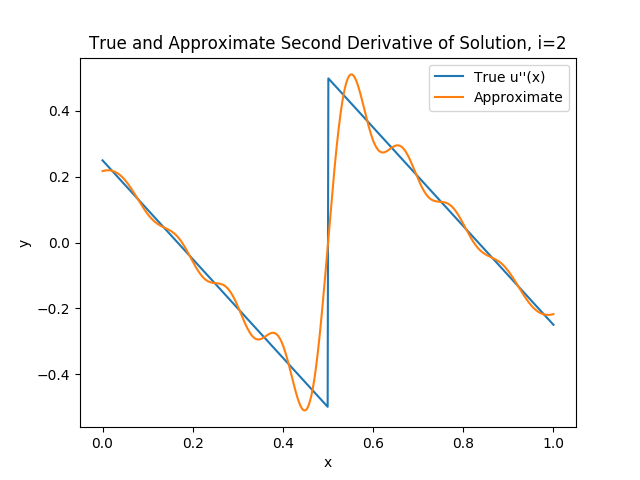}
			\caption{}
		\end{subfigure}
		\quad
		\begin{subfigure}{.3\textwidth}
			\centering
			\includegraphics[width=1.5in]{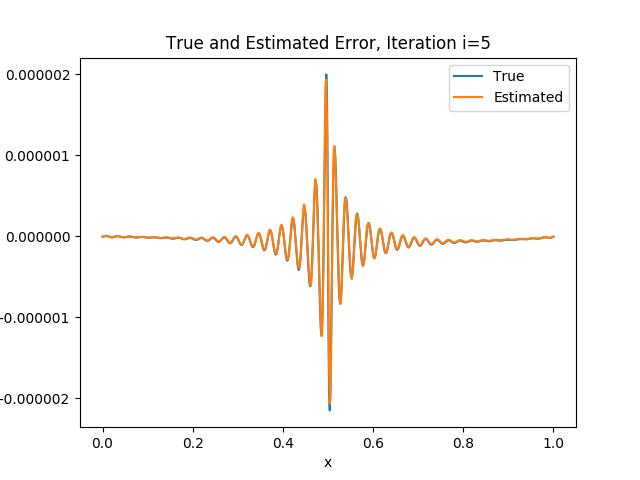}
			\caption{}
		\end{subfigure}
		\quad
		\begin{subfigure}{.3\textwidth}
			\centering
			\includegraphics[width=1.5in]{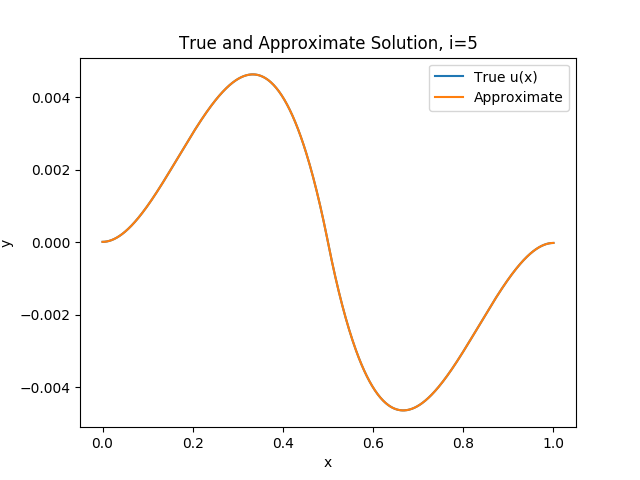}
			\caption{}
		\end{subfigure}
		\quad
		\begin{subfigure}{.3\textwidth}
			\centering
			\includegraphics[width=1.5in]{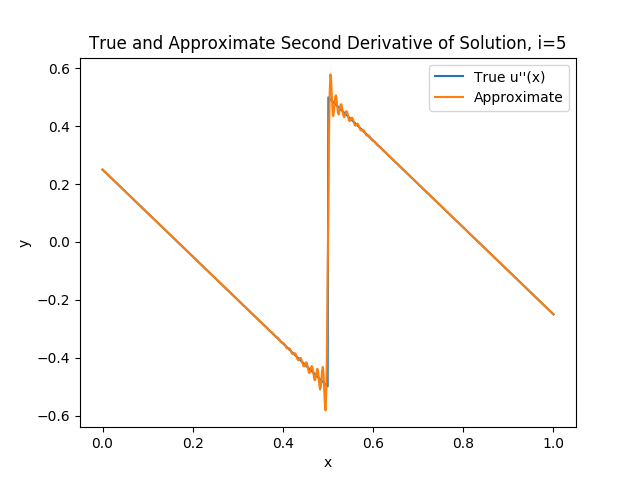}
			\caption{}
		\end{subfigure}
		\quad
		\begin{subfigure}{.3\textwidth}
			\centering
			\includegraphics[width=1.5in]{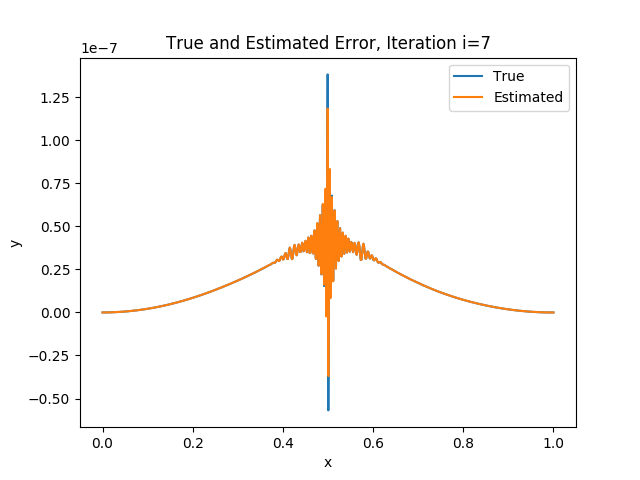}
			\caption{}
		\end{subfigure}
		\quad
		\begin{subfigure}{.3\textwidth}
			\centering
			\includegraphics[width=1.5in]{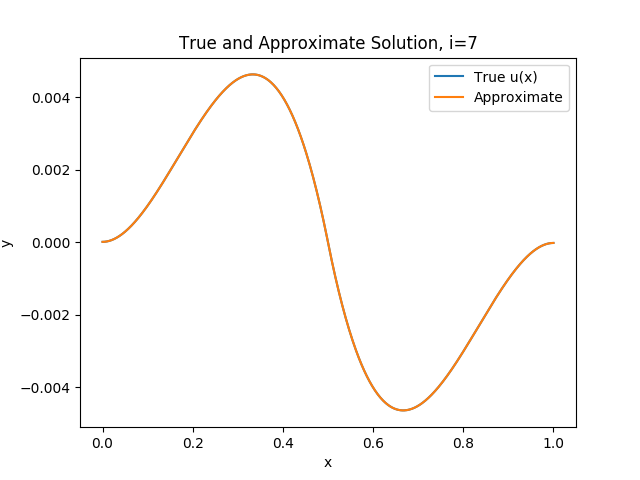}
			\caption{}
		\end{subfigure}
		\quad
		\begin{subfigure}{.3\textwidth}
			\centering
			\includegraphics[width=1.5in]{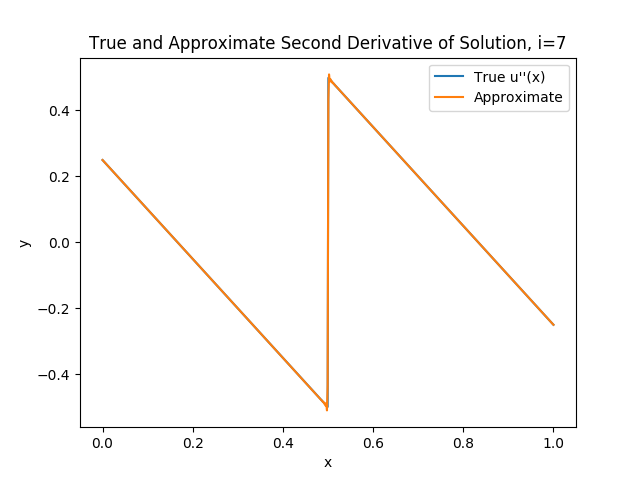}
			\caption{}
		\end{subfigure}
		\quad		
		\caption{Biharmonic equation with point moment in 1D. Exact error $u-u_{i-1}$ and approximate error $\varphi_{i}^{NN}$ as well as numerical solution $u_{i}$ and its derivative $u_{i}''$ for $i=2$ (top row), $i=5$ (middle row), and $i=7$ (bottom row).}
		\label{fig:biharmonic1d delta plots}
	\end{figure}

	Figure \ref{fig:biharmonic1d delta error galerkin} shows the true errors $||u-u_{i-1}||_{L^{2}}$ and $|||u-u_{i-1}|||$ at the end of each Galerkin iteration along with the estimated errors $||\varphi_{i}^{NN}||_{L^{2}}$ and $\eta(u_{i-1},\varphi_{i}^{NN})$ for our algorithm. Figure \ref{fig:biharmonic1d delta error epoch} shows the true and estimated errors at each training epoch, while Figure \ref{fig:biharmonic1d delta plots} shows the exact error $u-u_{i-1}$ as well as the maximizer $\varphi_{i}^{NN}$ at several stages of refinement along with both the numerical solution $u_{i}$ and its second derivative $u_{i}''$. We observe that initially, low frequency oscillations in the error are learned followed later by suppression of the high frequency error components which are particularly notable near the discontinuity in the second derivative of $u_{i}$. 

	\subsubsection{Circular Plate with Point Load}
	We consider the two-dimensional biharmonic equation with point load source term in the unit disk:
	\begin{align*}
		\begin{dcases}
			\Delta^{2}u = \delta_{\{(0,0)\}} &\text{in}\;\Omega = \{(x,y) : x^{2}+y^{2}<1\}\\
			u-\varepsilon_{1}\partial_{n}(\Delta u) = 0 &\text{on}\;\partial\Omega\\
			\Delta u + \varepsilon_{2}\partial_{n}u = 0 &\text{on}\;\partial\Omega.
		\end{dcases}
	\end{align*}
	
	\noindent The variational formulation is again posed on $X=H^{2}(\Omega)$ with bilinear operator $a(u,v) = (\Delta u,\Delta v)_{\Omega} + \varepsilon_{1}^{-1}(u,v)_{\partial\Omega} + \varepsilon_{2}(\partial_{n}u,\partial_{n}v)_{\partial\Omega}$ and data $L(v) = v(0,0)$. The exact variational solution is given by
	\begin{align*}
		u(r,\theta) &= \frac{r^{2}}{8\pi}\log(r) + c_{1}r^{2}+ c_{2}, \;\;\;c_{1} = \frac{1}{4+2\varepsilon_{2}} \cdot (-\frac{\varepsilon_{2}}{2\pi} - \frac{1}{8\pi}), \;\;\;c_{2} = -c_{1} + \frac{\varepsilon_{1}}{2\pi}.
	\end{align*}
	
	\noindent The corresponding loss function is approximated by
	\begin{align*}
		\frac{L_{1} - L_{2} - L_{3} - L_{4}} {\sqrt{ \sum_{i=1}^{n_{\Omega}} w_{i}|\Delta v(x_{i}^{\Omega};\theta)|^{2} + \varepsilon^{-1}\sum_{i=1}^{n_{\partial\Omega}} w_{i}^{\partial\Omega}v(x_{i}^{\partial\Omega};\theta)^{2} + \varepsilon_{2}\sum_{i=1}^{n_{\partial\Omega}} w_{i}^{\partial\Omega}\partial_{n}v(x_{i}^{\partial\Omega};\theta)^{2}}}
	\end{align*} 
	
	\noindent where $L_{1} = v(0;\theta)$, $L_{2} = \sum_{i=1}^{n_{\Omega}} w_{i}^{\Omega}\Delta u_{0}(x_{i}^{\Omega})\Delta v(x_{i}^{\Omega};\theta)$, $L_{3} = \varepsilon_{1}^{-1}\sum_{i=1}^{n_{\partial\Omega}} w^{\partial\Omega}_{i} u_{0}(x_{i}^{\partial\Omega})\cdot v(x_{i}^{\partial\Omega};\theta)$, and $L_{4} = \varepsilon_{2} \sum_{i=1}^{n_{\partial\Omega}} w^{\partial\Omega}_{i} \partial_{n}u_{0}(x_{i}^{\partial\Omega})\partial_{n}v(x_{i}^{\partial\Omega};\theta)$. Here, $\{w^{\Omega},x^{\Omega}\}$ is the quadrature rule in $\Omega$ while $\{w^{\partial\Omega}, x^{\partial\Omega}\}$ is the quadrature rule on $\partial\Omega$.
	
	\begin{figure}[t!]
		\centering
		\begin{subfigure}{.47\textwidth}
			\centering
			\includegraphics[width=2.0in]{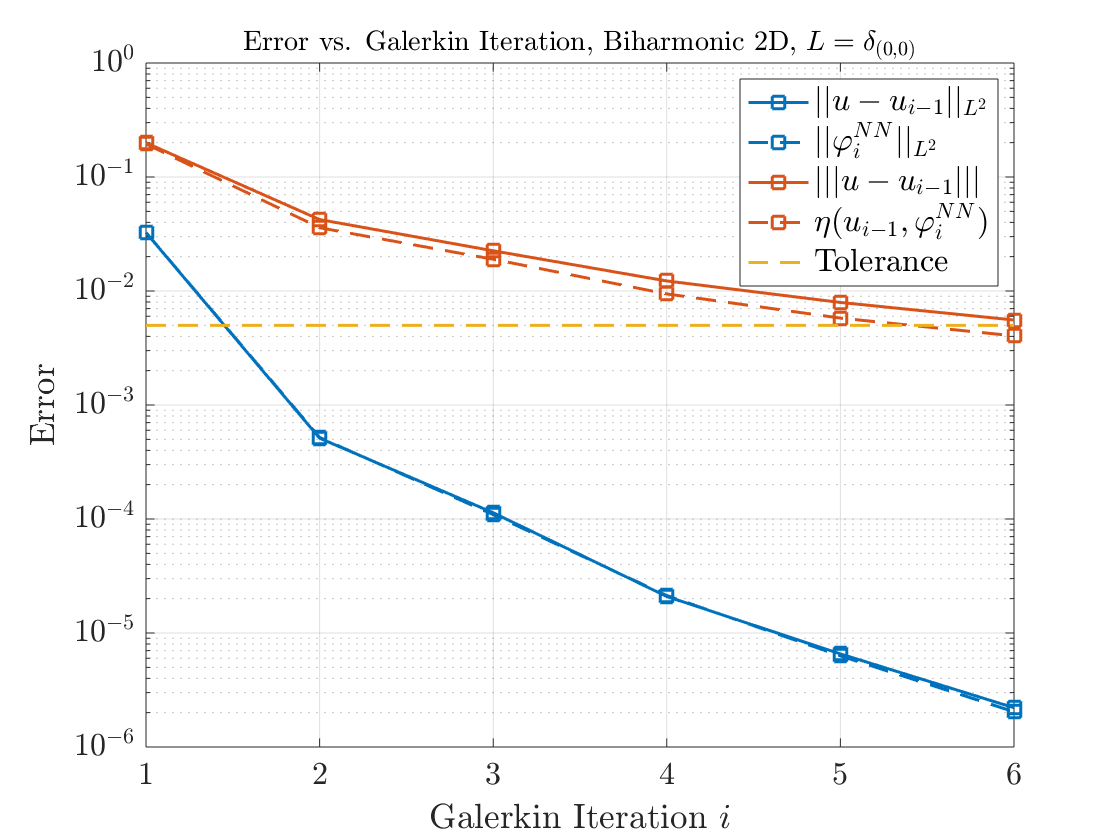}
			\caption{}
			\label{fig:biharmonic2d delta error galerkin}
		\end{subfigure}
		\quad
		\begin{subfigure}{.47\textwidth}
			\centering
			\includegraphics[width=2.0in]{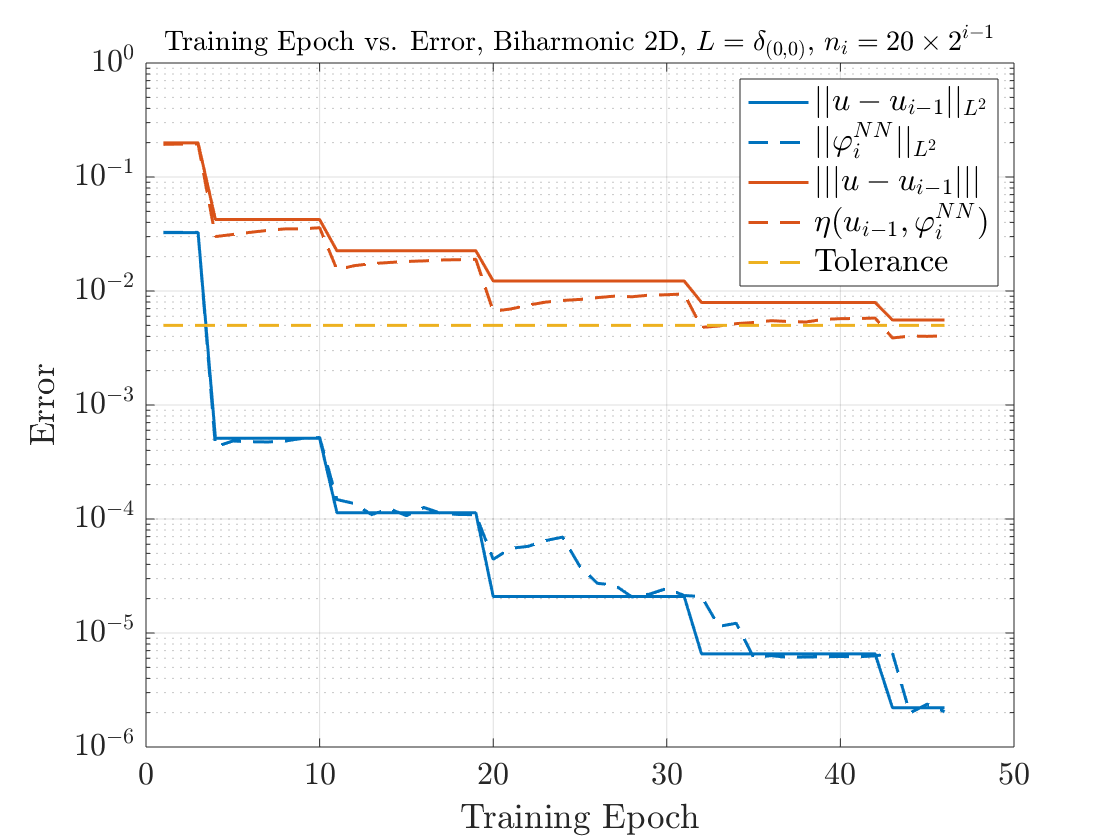}
			\caption{}
			\label{fig:biharmonic2d delta error epoch}
		\end{subfigure}
		\caption{The biharmonic equation in 2D. True and estimated error in the $L^{2}$ and energy norm at the end of each Galerkin iteration (a) and epoch (b).}
		\label{fig:biharmonic2d delta errors}
	\end{figure}
	
	The tolerance is set to $\texttt{tol} = 5\times 10^{-3}$, and $\varepsilon_{1} = \varepsilon_{2} = 10^{-5}$. The network architecture for iteration $i$ is as follows: $n_{i} = 20\times 2^{i-1}$, $\sigma_{i}(t) = \tanh(t)$. The learning rate for each basis function $\varphi_{i}$ is $\alpha_{i} = \frac{1 \times 10^{-2}}{1.1^{i-1}}$. The hidden parameters are initialized according to the box initialization in \cite{cyr}. We employ fixed tensor product Gauss-Legendre quadrature rule with $100 \times 100$ nodes in order to approximate inner products in the interior of the domain. We approximate the inner products on $\partial\Omega$ using $256$ equally-spaced nodes on $\partial\Omega$.
	
\begin{figure}[t!]
		\centering
		\begin{subfigure}{.3\textwidth}
			\centering
			\includegraphics[width=1.5in]{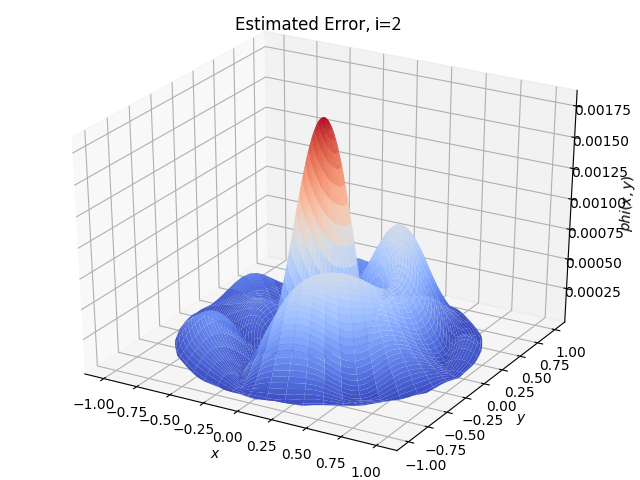}
			\caption{}
		\end{subfigure}
		\quad
		\begin{subfigure}{.3\textwidth}
			\centering
			\includegraphics[width=1.5in]{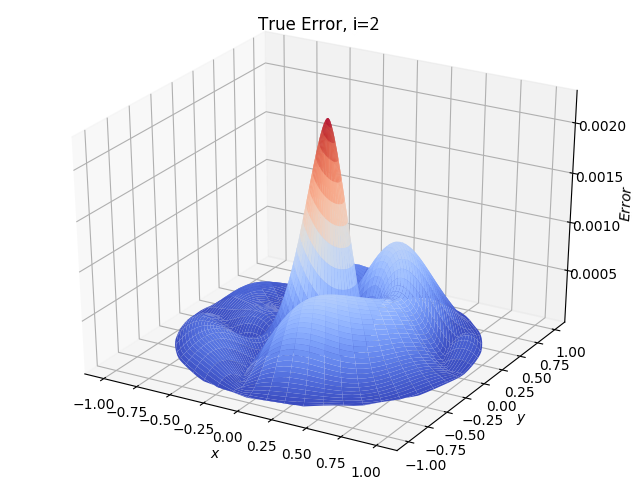}
			\caption{}
		\end{subfigure}
		\quad
		\begin{subfigure}{.3\textwidth}
			\centering
			\includegraphics[width=1.5in]{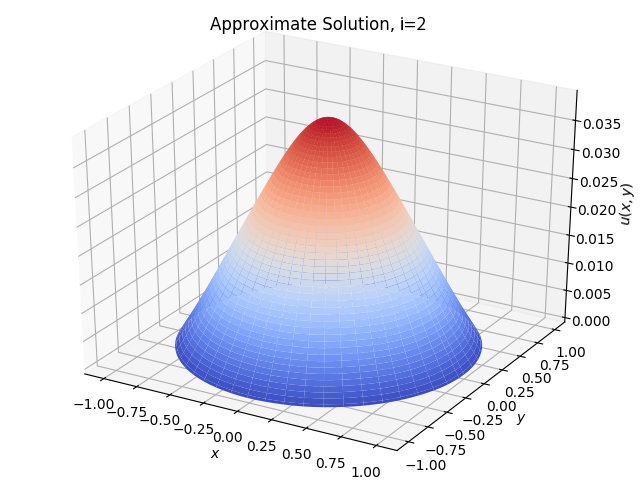}
			\caption{}
		\end{subfigure}
		\quad
		\begin{subfigure}{.3\textwidth}
			\centering
			\includegraphics[width=1.5in]{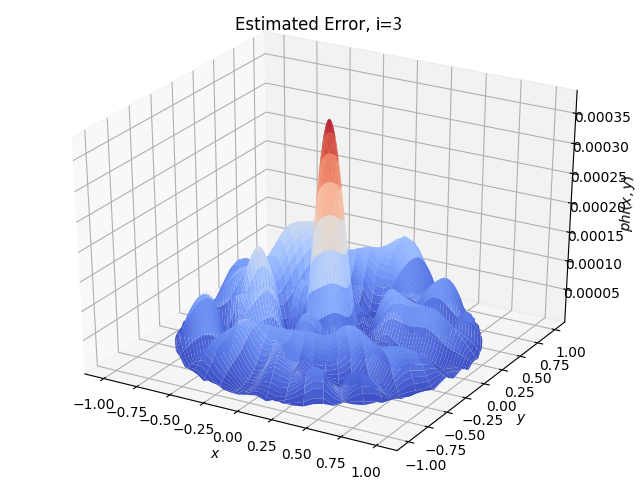}
			\caption{}
		\end{subfigure}
		\quad
		\begin{subfigure}{.3\textwidth}
			\centering
			\includegraphics[width=1.5in]{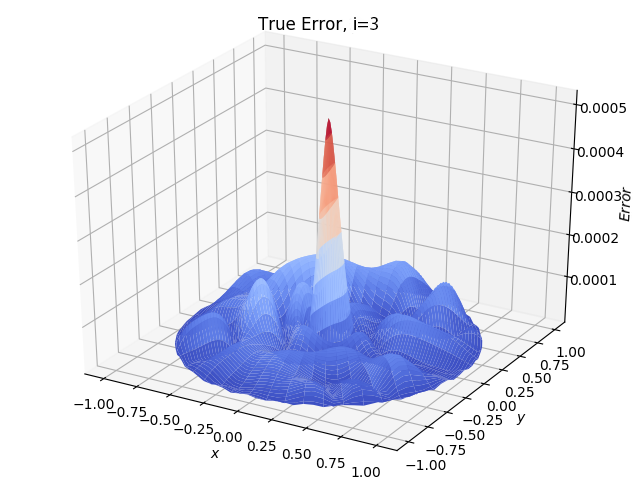}
			\caption{}
		\end{subfigure}
		\quad
		\begin{subfigure}{.3\textwidth}
			\centering
			\includegraphics[width=1.5in]{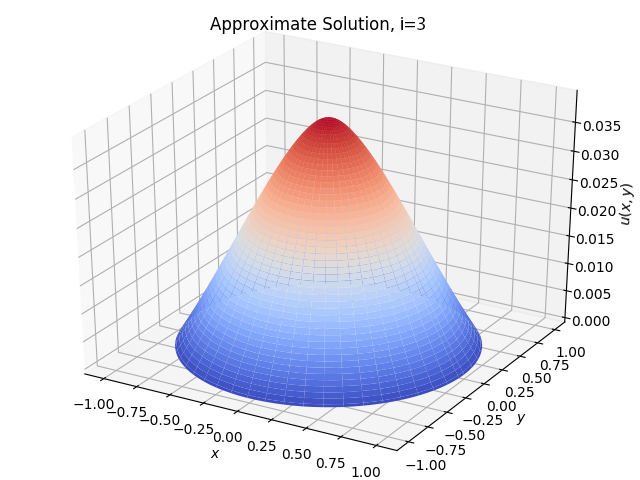}
			\caption{}
		\end{subfigure}
		\quad
		\begin{subfigure}{.3\textwidth}
			\centering
			\includegraphics[width=1.5in]{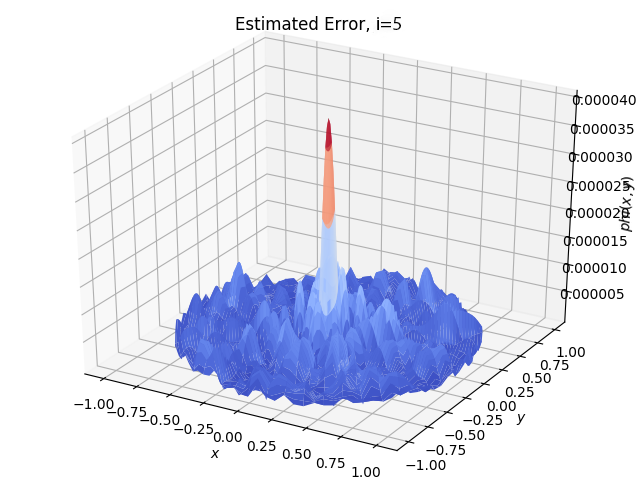}
			\caption{}
		\end{subfigure}
		\quad
		\begin{subfigure}{.3\textwidth}
			\centering
			\includegraphics[width=1.5in]{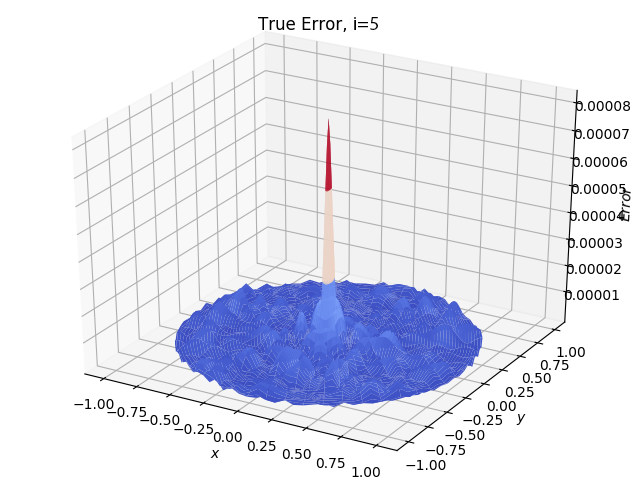}
			\caption{}
		\end{subfigure}
		\quad
		\begin{subfigure}{.3\textwidth}
			\centering
			\includegraphics[width=1.5in]{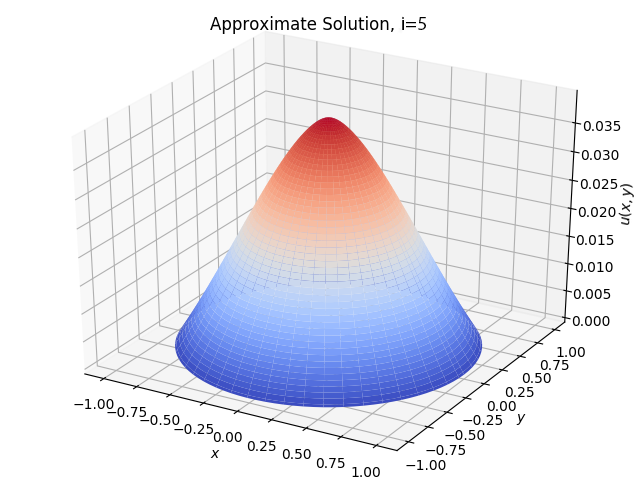}
			\caption{}
		\end{subfigure}
		\quad
		\caption{Biharmonic equation in 2D with point load source. True error $u-u_{i-1}$ (middle column) and estimated error $\varphi_{i}^{NN}$ (left column) as well as numerical solution $u_{i}$ (right column) for $i=2,3,5$.}
		\label{fig:biharmonic2d delta plots}
	\end{figure}

	Figure \ref{fig:biharmonic2d delta error galerkin} shows the true errors $||u-u_{i-1}||_{L^{2}}$ and $|||u-u_{i-1}|||$ at the end of each Galerkin iteration along with the estimated errors $||\varphi_{i}^{NN}||_{L^{2}}$ and $\eta(u_{i-1},\varphi_{i}^{NN})$ for our algorithm. Figure \ref{fig:biharmonic2d delta error epoch} shows the true and estimated errors at each training epoch, while Figure \ref{fig:biharmonic2d delta plots} shows the true error $u-u_{i}$ (middle column) as well as the maximizer $\varphi_{i}^{NN}$ (left column) and approximate solution $u_{i}$ (right column) at several stages of refinement. We observe no difficulties in handling the singular data in two dimensions.

	\section{Conclusions} \label{sec:conclusions}
	We have introduced a framework for adaptively generating basis functions for a finite-dimensional subspace which may be used in a standard Galerkin method for approximating variational problems. We have demonstrated that each basis function $\varphi_{i}^{NN}$ may be expressed as the realization of a neural network $V^{\sigma_{i}}_{n_{i},C_{i}}$ with a single hidden layer by learning the dual representation of the weak residual $r(u_{i-1}) : X \to \mathbb{R}$. Furthermore, we have shown that $\eta(u_{i-1},\varphi_{i}^{NN})$ is an equivalent measure of the error $|||u-u_{i-1}|||$ and constitutes a robust a posteriori error estimate. We emphasize that our algorithm may be applied seamlessly to the broad class of problems in this section with no major modifications. This is in contrast to e.g. the finite element method which requires a tailored approach depending on the underlying Sobolev space and spatial dimension of the variational problem.

	\section{Acknowledgments}
	This material is based upon work supported by the National Science Foundation Graduate Research Fellowship under Grant No. 1644760 as well as PhILMS grant DE-SC0019453.

	\bibliographystyle{siamplain}
	\bibliography{references}

\begin{thebibliography}{10}

\bibitem{berg}
{\sc J.~Berg and K.~Nystr{\"o}m}, {\em A unified deep artificial neural network
  approach to partial differential equations in complex geometries},
  Neurocomputing, 317 (2018), pp.~28--41.

\bibitem{brennerscott}
{\sc S.~Brenner and R.~Scott}, {\em The mathematical theory of finite element
  methods}, vol.~15, Springer Science \& Business Media, 2007.

\bibitem{cybenko}
{\sc G.~Cybenko}, {\em Approximation by superpositions of a sigmoidal
  function}, Mathematics of control, signals and systems, 2 (1989),
  pp.~303--314.

\bibitem{cyr}
{\sc E.~C. Cyr, M.~A. Gulian, R.~G. Patel, M.~Perego, and N.~A. Trask}, {\em
  Robust training and initialization of deep neural networks: An adaptive basis
  viewpoint}, arXiv preprint arXiv:1912.04862,  (2019).

\bibitem{davis}
{\sc P.~J. Davis and P.~Rabinowitz}, {\em Methods of numerical integration},
  Courier Corporation, 2007.

\bibitem{dissanayake}
{\sc M.~Dissanayake and N.~Phan-Thien}, {\em Neural-network-based
  approximations for solving partial differential equations}, communications in
  Numerical Methods in Engineering, 10 (1994), pp.~195--201.

\bibitem{ern}
{\sc A.~Ern and J.-L. Guermond}, {\em Theory and practice of finite elements},
  vol.~159, Springer Science \& Business Media, 2013.

\bibitem{folland}
{\sc G.~B. Folland}, {\em Real analysis: modern techniques and their
  applications}, vol.~40, John Wiley \& Sons, 1999.

\bibitem{gershgorin}
{\sc S.~Gerschgorin}, {\em Uber die abgrenzung der eigenwerte einer matrix},
  Izvestija Akademii Nauk SSSR, Serija Matematika, 7 (1931), pp.~749--754.

\bibitem{han}
{\sc J.~Han, A.~Jentzen, and E.~Weinan}, {\em Solving high-dimensional partial
  differential equations using deep learning}, Proceedings of the National
  Academy of Sciences, 115 (2018), pp.~8505--8510.

\bibitem{he2}
{\sc J.~He, L.~Li, J.~Xu, and C.~Zheng}, {\em Relu deep neural networks and
  linear finite elements}, Journal of Computational Mathematics,  (2019).

\bibitem{hornik}
{\sc K.~Hornik}, {\em Approximation capabilities of multilayer feedforward
  networks}, Neural networks, 4 (1991), pp.~251--257.

\bibitem{jagtap}
{\sc A.~D. Jagtap, K.~Kawaguchi, and G.~E. Karniadakis}, {\em Adaptive
  activation functions accelerate convergence in deep and physics-informed
  neural networks}, Journal of Computational Physics, 404 (2020), p.~109136.

\bibitem{kharazmi}
{\sc E.~Kharazmi, Z.~Zhang, and G.~Karniadakis}, {\em Variational
  physics-informed neural networks for solving partial differential equations},
  arXiv preprint arXiv:1912.00873,  (2019).

\bibitem{adam}
{\sc D.~P. Kingma and J.~Ba}, {\em Adam: A method for stochastic optimization},
  arXiv preprint arXiv:1412.6980,  (2014).

\bibitem{lagaris}
{\sc I.~E. Lagaris, A.~Likas, and D.~I. Fotiadis}, {\em Artificial neural
  networks for solving ordinary and partial differential equations}, IEEE
  transactions on neural networks, 9 (1998), pp.~987--1000.

\bibitem{lagaris2}
{\sc I.~E. Lagaris, A.~C. Likas, and D.~G. Papageorgiou}, {\em Neural-network
  methods for boundary value problems with irregular boundaries}, IEEE
  Transactions on Neural Networks, 11 (2000), pp.~1041--1049.

\bibitem{leshno}
{\sc M.~Leshno, V.~Y. Lin, A.~Pinkus, and S.~Schocken}, {\em Multilayer
  feedforward networks with a nonpolynomial activation function can approximate
  any function}, Neural networks, 6 (1993), pp.~861--867.

\bibitem{mishra}
{\sc S.~Mishra and R.~Molinaro}, {\em Estimates on the generalization error of
  physics informed neural networks (pinns) for approximating pdes ii: A class
  of inverse problems}, arXiv preprint arXiv:2007.01138,  (2020).

\bibitem{petersen2018topological}
{\sc P.~Petersen, M.~Raslan, and F.~Voigtlaender}, {\em Topological properties
  of the set of functions generated by neural networks of fixed size}.

\bibitem{raissi}
{\sc M.~Raissi, P.~Perdikaris, and G.~E. Karniadakis}, {\em Physics-informed
  neural networks: A deep learning framework for solving forward and inverse
  problems involving nonlinear partial differential equations}, Journal of
  Computational Physics, 378 (2019), pp.~686--707.

\bibitem{sirignano}
{\sc J.~Sirignano and K.~Spiliopoulos}, {\em Dgm: A deep learning algorithm for
  solving partial differential equations}, Journal of computational physics,
  375 (2018), pp.~1339--1364.

\bibitem{weinan}
{\sc E.~Weinan and B.~Yu}, {\em The deep ritz method: a deep learning-based
  numerical algorithm for solving variational problems}, Communications in
  Mathematics and Statistics, 6 (2018), pp.~1--12.

\bibitem{zang}
{\sc Y.~Zang, G.~Bao, X.~Ye, and H.~Zhou}, {\em Weak adversarial networks for
  high-dimensional partial differential equations}, Journal of Computational
  Physics,  (2020), p.~109409.

\end{thebibliography}
\end{document}